\documentclass[letterpaper,11pt]{article}

\usepackage[utf8]{inputenc}
\usepackage[T1]{fontenc}
\usepackage{times}
\usepackage[margin=1in,top=0.45in]{geometry}
\usepackage{microtype}

\usepackage{amsmath,amssymb,amsfonts,amsthm}
\usepackage{mathrsfs}
\usepackage{dsfont}

\usepackage{graphicx}
\usepackage[export]{adjustbox}
\usepackage{array}
\usepackage{tabularx}
\usepackage{booktabs}
\usepackage{multirow}
\usepackage{makecell}
\usepackage{longtable}
\usepackage{wrapfig}
\usepackage{algorithm}
\usepackage{algpseudocode}
\usepackage{caption}

\usepackage{etoolbox}
\usepackage{enumitem}
\usepackage{xspace}
\usepackage[most]{tcolorbox}
\usepackage{xcolor}
\usepackage{colortbl}
\usepackage{tikz}
\usepackage{etoc}
\usepackage{todonotes}

\usepackage[numbers,sort&compress]{natbib}
\usepackage{url}
\usepackage[colorlinks=true,linkcolor=cyan,citecolor=cyan,filecolor=magenta,urlcolor=blue]{hyperref}
\usepackage[capitalize]{cleveref}
\crefname{section}{Sec.}{Secs.}
\Crefname{section}{Section}{Sections}
\Crefname{table}{Table}{Tables}
\crefname{table}{Tab.}{Tabs.}

\theoremstyle{definition}

\renewcommand{\thefootnote}{\fnsymbol{footnote}}

\setlist{leftmargin=5mm}

\definecolor{abstractbg}{RGB}{230,242,250}
\definecolor{abstractborder}{RGB}{200,210,220}
\usepackage{titlesec}

\titleformat{\section}
  {\sffamily\bfseries\Large} 
  {\thesection}              
  {1em}                      
  {}                         

\renewenvironment{abstract}
{
\begin{center}
\begin{tcolorbox}[
    colback=abstractbg,
    colframe=abstractborder,
    boxrule=0.6pt,
    arc=6pt,
    width=\textwidth,
    left=8pt,
    right=8pt,
    top=8pt,
    bottom=8pt
]
}
{
\end{tcolorbox}
\end{center}
}

\DeclareCaptionFont{preprint}{\fontsize{10}{12}\selectfont}
\newlength\colfigwidth
\makeatletter
\let\px@origvspace\vspace
\newcommand\px@vspace{\@ifstar\px@vspace@\px@vspace@}
\newcommand\px@vspace@[1]{\ifdim\glueexpr#1\relax<\z@\else\px@origvspace{#1}\fi}
\newcommand\px@tablesizes{%
  \renewcommand\normalsize{\@setfontsize\normalsize\@xpt\@xiipt}%
  \renewcommand\small{\@setfontsize\small\@ixpt{11}}%
  \renewcommand\footnotesize{\@setfontsize\footnotesize\@viiipt{9.5}}%
  \renewcommand\scriptsize{\@setfontsize\scriptsize\@viipt\@viiipt}%
  \renewcommand\tiny{\@setfontsize\tiny\@vpt\@vipt}}
\newcount\px@tab
\newcount\px@fig
\newcommand\px@fitbox[3]{\adjustbox{max width=\linewidth}{#3}}
\def\px@close{\end{adjustbox}}
\def\px@tabopen{\begin{adjustbox}{max width=\linewidth}\px@tab\tw@}
\def\px@figfull{\begin{adjustbox}{max width=\linewidth}\px@fig\thr@@}
\def\px@figcol{\begin{adjustbox}{width=\colfigwidth}\px@fig\thr@@}
\AtBeginEnvironment{table}{\px@tablesizes\px@tab\@ne\let\resizebox\px@fitbox\let\vspace\px@vspace}
\AtBeginEnvironment{table*}{\px@tablesizes\px@tab\@ne\let\vspace\px@vspace}
\AtBeginEnvironment{figure}{\px@fig\tw@\let\vspace\px@vspace}
\AtBeginEnvironment{figure*}{\px@fig\@ne\let\vspace\px@vspace}
\BeforeBeginEnvironment{tabular}{%
  \ifcase\px@tab\or\expandafter\px@tabopen\else\advance\px@tab\@ne\fi}
\AfterEndEnvironment{tabular}{%
  \ifnum\px@tab>\tw@\advance\px@tab\m@ne
  \else\ifnum\px@tab=\tw@\expandafter\expandafter\expandafter\px@close\fi\fi}
\BeforeBeginEnvironment{tikzpicture}{%
  \ifcase\px@fig\or\expandafter\px@figfull\or\expandafter\px@figcol\else\advance\px@fig\@ne\fi}
\AfterEndEnvironment{tikzpicture}{%
  \ifnum\px@fig>\thr@@\advance\px@fig\m@ne
  \else\ifnum\px@fig=\thr@@\expandafter\expandafter\expandafter\px@close\fi\fi}

\newcommand\px@tocline[3]{%
  {\parfillskip\z@\rightskip\z@\noindent\hspace*{#1}\makebox[#2][l]{\etocnumber}\etocname
   \nobreak#3\nobreak\makebox[1.6em][r]{\etocpage}\par}}
\newcommand\appendixcontents{%
  \begingroup
    \parindent\z@
    \etocsettagdepth{mtmain}{none}%
    \etocsettagdepth{mtappendix}{subsection}%
    \etocsettocstyle{{\sffamily\bfseries\large Contents\par}\vspace{4pt}}{}%
    \etocsetstyle{section}{\etocskipfirstprefix}{\addvspace{6pt}}%
      {{\bfseries\px@tocline{\z@}{1.9em}{\hfill}}}{}%
    \etocsetstyle{subsection}{}{}%
      {\px@tocline{1.9em}{2.7em}{\leaders\hbox to .6em{\hss.\hss}\hfill}}{}%
    \etoctoclines
    \tableofcontents
  \endgroup}
\makeatother

\newcommand{\papertitle}{%
\sffamily\bfseries\fontsize{16}{20}\selectfont
Streaming Tax: \\[.3em] Can Streaming VLMs Replace Their Offline Parents?
}

\newcommand{\paperauthors}{%
\sffamily
Zhi Yang$^1$, Junzhou Chen$^1$, Jindong Wang$^1$\footnote{Corresponding author: \href{mailto:jdw@wm.edu}{jdw@wm.edu}.}%
}

\newcommand{\paperdate}{$^1$William \& Mary}

\hypersetup{
  pdftitle={Streaming Tax: Can Streaming VLMs Replace Their Offline Parents?},
  pdfauthor={Zhi Yang, Junzhou Chen, Jindong Wang}}

\begin{document}
\etocdepthtag.toc{mtmain}

\thispagestyle{empty}

\noindent
\includegraphics[height=.6cm]{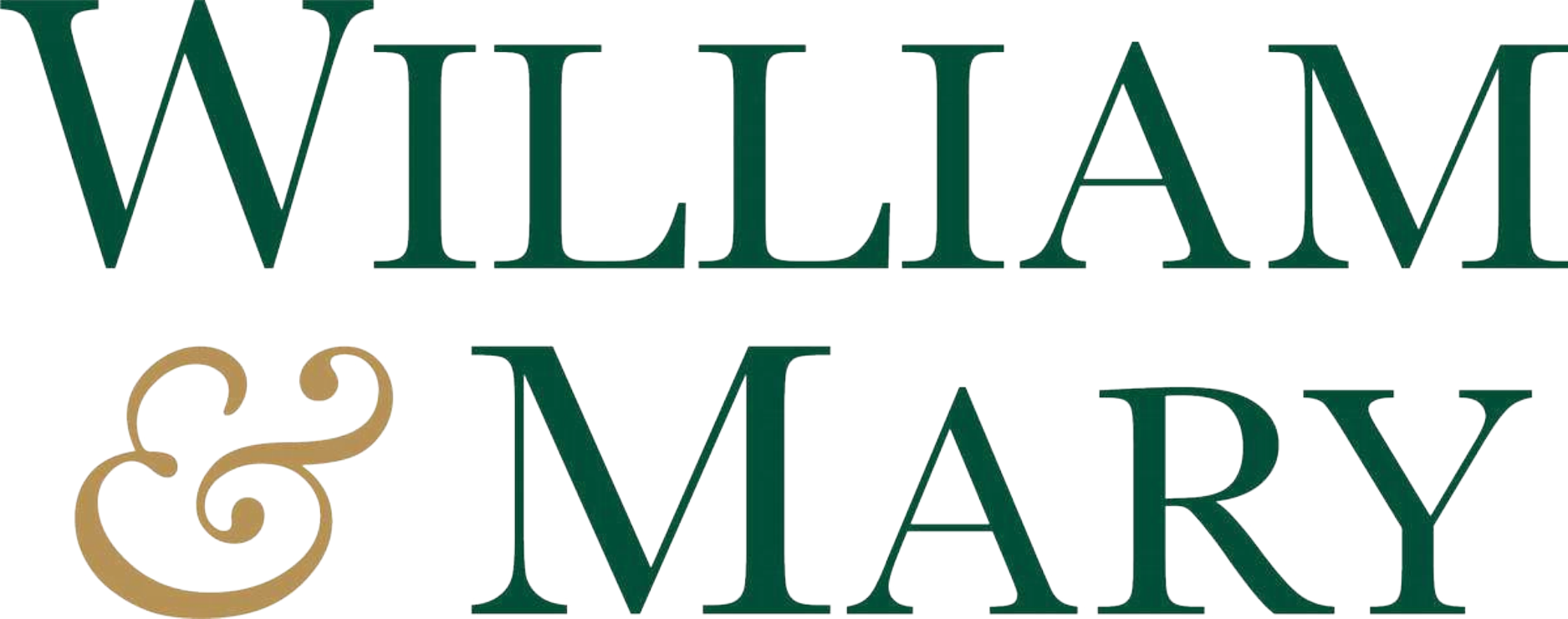}

\vspace{-.15in}

\noindent\rule{\textwidth}{0.8pt}

\vspace{0.65cm}

\begin{center}
    {\papertitle\par}
    \vspace{0.45cm}
    {\large \paperauthors\par}
    \vspace{0.2cm}
    {\normalsize \paperdate\par}
\end{center}

\begin{abstract}
Most vision-language models (VLMs) specialize in offline video understanding; in contrast, their post-trained versions, streaming VLMs, can perform live interactions.
We investigate whether streaming VLMs can replace their offline parent models on \textit{offline} tasks. Such replacement could reduce deployment costs and improve efficiency.
We propose \textbf{Streaming Tax} to systematically evaluate streaming VLMs on offline tasks.
Our exploration spans three core dimensions: understanding, trustworthiness, and generalization across 15 benchmarks.
Our findings reveal a critical gap between offline capability preservation and practical replaceability.
This gap emerges at two stages: streaming adaptation can alter trustworthy behavior even when offline understanding is largely preserved, while the native streaming protocol can introduce further performance losses.
Our analyses show that some of these losses arise from restricted access to historical visual evidence and failures to deliver usable responses. Importantly, regarding safety, streaming models often leave both harmful and legitimate requests unanswered rather than genuinely refusing unsafe ones.
We then propose \textbf{Stream++}, a training-free plugin to improve streaming VLMs on offline tasks. Our code is available at \url{https://github.com/AIFrontierLab/Streaming-tax}.
\end{abstract}

\setcounter{footnote}{0}
\renewcommand{\thefootnote}{\arabic{footnote}}

\noindent\begin{minipage}{\textwidth}
    \centering
    \includegraphics[width=\textwidth]{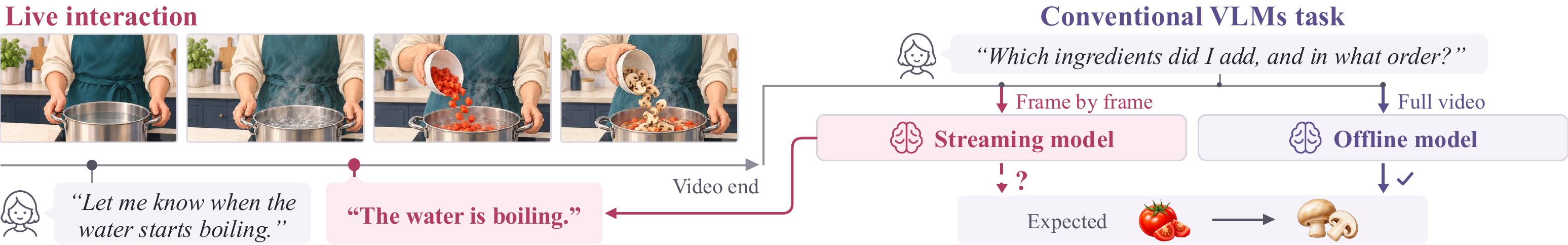}
    \captionof{figure}{A cooking assistant should alert the user when the water boils (live interaction) and, after cooking, recall the ingredients and their order (offline video understanding). Streaming models are typically designed for the former; we investigate whether they can also replace their offline parents on the latter (dashed arrow).}
    \label{fig:motivation}
\end{minipage}
\par\vspace{4pt}

\section{Introduction}
\label{sec:introduction}

Streaming vision-language models (VLMs) are a new family of models obtained by post-training offline VLMs on real-time interaction tasks \citep{huang2025ovbench,liu2026thinkstream,wang2025mmduet2,lu2026aura}.
The streaming capability allows them to process continuous video streams and decide when to respond, without processing the entire video.
For instance, AURA-8B~\citep{lu2026aura} is a streaming VLM (``child'') post-trained from Qwen3-VL-8B~\citep{bai2025qwen3vl}, a conventional offline VLM (``parent'').
Because streaming models are post-trained from general offline backbones, they are naturally expected to inherit their parents' offline capabilities.
Thus, it is intuitive to ask: \textit{Can streaming models replace their offline parents with comparable performance on conventional offline tasks?}\footnote{By ``replace", we only mean to explore the performance comparison in \textit{applications}, but never claim that offline models are not useful. Offline models are more general and are the bases for training streaming ones.}

\begin{figure*}[t!]
    \centering
    \includegraphics[width=0.94\textwidth]{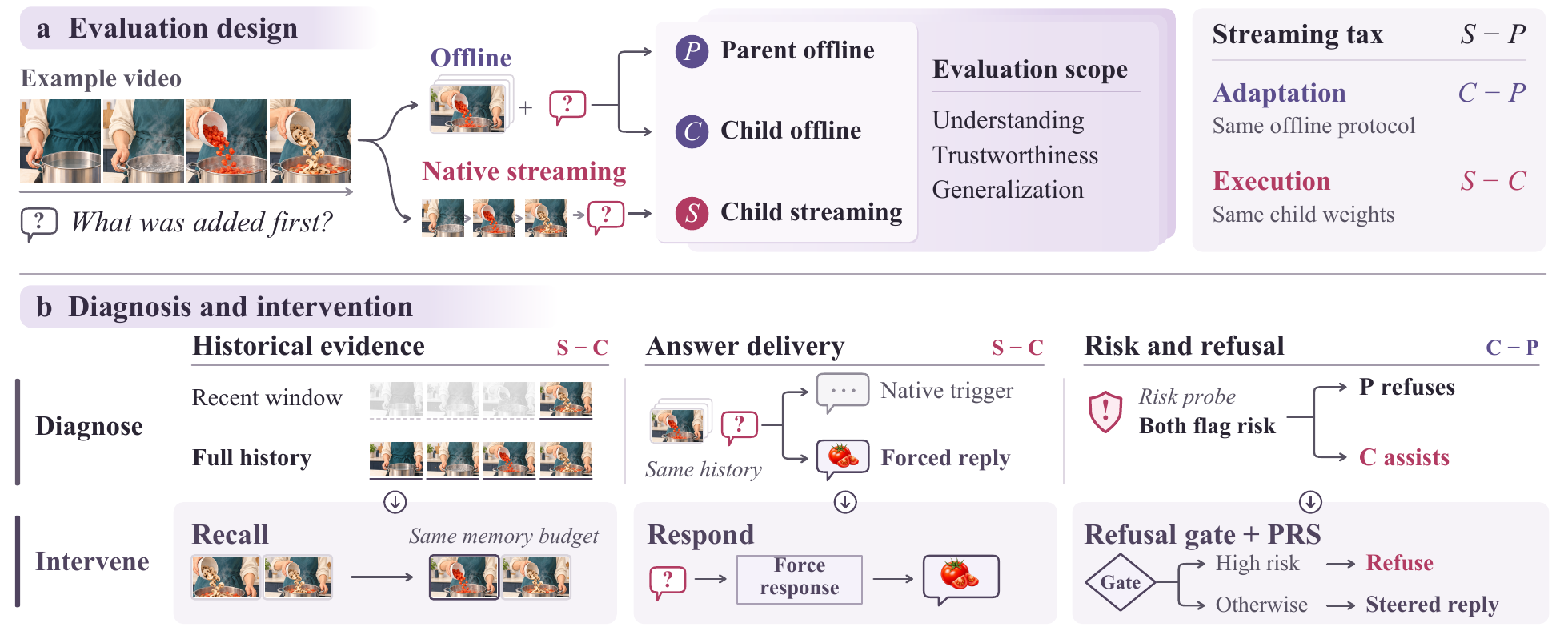}
    \vspace{-.1in}
    \caption{\textbf{Study overview.} We pair six streaming children with five offline parents and compare parent offline ($P$), child offline ($C$), and child native streaming ($S$), where applicable. We evaluate understanding, trustworthiness, and generalization, investigate possible causes of selected gaps, and test targeted interventions.}
    \label{fig:study-overview}
    \vspace{-.1in}
\end{figure*}

In practice, a single streaming model capable of both real-time interaction and offline understanding offers several advantages over maintaining separate models.
First, it reduces deployment and maintenance costs while enabling more consistent and coherent understanding of the same video content.
For example, a cooking assistant could notify the user when water starts boiling and later answer questions about which ingredients were used and in what order (\Cref{fig:motivation}).
Moreover, a unified model allows feedback and improvements from both streaming and offline tasks to be shared, potentially facilitating more effective model evolution.
Some recent studies investigated this question. By switching the inference protocol to an \textit{offline mode}, MMDuet2~\cite{wang2025mmduet2} and AURA~\cite{lu2026aura} are compared against their offline parents on standard video benchmarks, while EvoStreaming~\cite{wen2026evostreaming} examines the preservation of offline capabilities across multiple backbones.

These efforts offer valuable evidence, yet largely evaluate adapted models in offline mode on standard understanding tasks \citep{niu2025ovobench}, leaving practical replaceability unresolved for two reasons.
First, even when switching to offline mode is feasible, preserving accuracy does not guarantee trustworthy replacement: safety guardrails, fairness, privacy, prompt robustness, and downstream adaptability may still shift after streaming adaptation.
Second, switching to offline mode is often impractical in continuous sessions (e.g., \Cref{fig:motivation}), as it requires storing every raw frame and re-encoding the full video at question time; avoiding this overhead requires answering directly from the native streaming state, introducing protocol-level constraints alongside weight adaptation. Beyond measuring these entangled effects, it is crucial to diagnose why failures occur and how to mitigate them.

In this paper, we propose \textbf{Streaming Tax} to comprehensively evaluate popular streaming VLMs and their offline parents across understanding, trustworthiness, and generalization tasks (\Cref{fig:study-overview}).
Streaming tax refers to the end-to-end gap between the child under native streaming mode and its offline parent, which we decompose into two stages: \textit{adaptation tax}, comparing parent and child under matched offline inference (testing whether the child suffices when switching to offline mode), and \textit{execution tax}, comparing the same child between native streaming and offline inference (measuring the extra cost of answering directly in streaming mode).

Our evaluation spans $15$ benchmarks and over $26K$ evaluation samples in total.
Under matched offline inference, streaming VLMs generally retain much of their parents' visual understanding, prompt robustness, and downstream performance.
However, the safety behavior changes more substantially since most children refuse or safely redirect fewer harmful video requests.
Native streaming introduces further understanding losses and changes response behavior.
Lower attack success rates coincide with reduced answering of harmless questions for most children, while fairness and privacy scores also decline for most children relative to their offline performance.

Controlled analyses trace failures to limited access to historical visual evidence and response policies that withhold answers.
We also investigate cases where a streaming model recognizes the same risk as its offline parent yet provides harmful assistance that the parent refuses.
Guided by these findings, we propose \textbf{Stream++}, a training-free plugin with three components: \texttt{Recall} retains earlier frames within the same native memory budget, \texttt{Respond} ensures usable responses, and \texttt{Refuse} restores the parent's refusal behavior on harmful requests.
Stream++ improves the video understanding performance and keeps streaming models safer than their offline parents on harmful video requests in nearly all comparisons.
Practical replacement therefore requires preserving inherited capabilities, retaining evidence for later questions, and responding appropriately to both harmful and legitimate requests during streaming.

Our contributions are threefold:
\begin{itemize}
    \item We propose Streaming Tax to comprehensively evaluate streaming vs. offline VLMs across understanding, trustworthiness, and generalization tasks.

    \item We diagnose replacement failures involving historical evidence access, answer delivery, and consistency between risk recognition and response behavior.

    \item We propose Stream++, a training-free plugin to improve the performance of streaming VLMs on offline tasks.
\end{itemize}

\section{Related Work}
\label{sec:related-work}

\noindent \textbf{From offline to streaming VLM.}
Recent work has developed ways to adapt pretrained offline VLMs for continuous observation and real-time interaction~\cite{wang2025mmduet,wang2025mmduet2,lu2026aura,wen2026evostreaming,chen2024videollmonline,qian2025dispider}.
These approaches differ in their post-training objectives and in the memory and interaction mechanisms used during streaming execution~\cite{liu2026thinkstream,yang2025livestar,wang2026mossvl,chen2025livecc,zhang2025flashvstream,xu2026streamingvlm,ding2025streammind}.
This diversity motivates examining whether inherited capabilities are preserved consistently across model families.

\noindent \textbf{VLM evaluation.}
Conventional video benchmarks assess understanding and temporal reasoning~\cite{fu2025videomme,shangguan2025tomato,li2024mvbench,liu2024tempcompass,wu2024longvideobench,zhou2025mlvu}, while online benchmarks evaluate understanding and interaction under sequential observation~\cite{niu2025ovobench,huang2025ovbench,lin2024streamingbench,yang2025svbench,wang2025omnimmi}.
Studies on streaming models also examine whether adaptation preserves conventional capabilities.
MMDuet2, AURA, and ThinkStream compare adapted models with their original backbones on conventional video tasks~\cite{wang2025mmduet2,lu2026aura,liu2026thinkstream}; StreamBridge and EvoStreaming investigate preservation across multiple backbones~\cite{wang2025streambridge,wen2026evostreaming}.
Evaluation also depends on how a model processes inputs.
\citet{wang2026thinkwhilewatching} compare original and adapted models across batch and streaming conditions.
StreamingEval provides a shared framework for evaluating offline and online models under streaming constraints, while retaining native online mechanisms and equipping offline models with a bounded memory bank~\cite{tang2026streamingeval}.
These studies establish precedents for capability preservation and cross-mode evaluation, focusing on particular adaptation methods or evaluation objectives.

\noindent \textbf{Our difference.}
What remains less systematically characterized is how consistently the streaming models can replace their offline parents across task domains and operating modes, and how to explain failures.
Video safety, fairness, and privacy benchmarks provide dimensions beyond understanding accuracy~\cite{liu2025videosafetybench,kang2026multiclip,potter2025vmdt}, but the preservation of these properties, together with prompt robustness and downstream adaptability, remains insufficiently understood in streaming adaptation.
Interpreting these comparisons also requires accounting for response coverage and output validity and investigating how adaptation and execution conditions contribute to observed differences.

\section{Evaluation Design}
\label{sec:setup}

\begin{table*}[t!]
\centering
\caption{
Overview of evaluation scope and metrics.
}
\vspace{-.1in}
\label{tb-eval-setup}
\footnotesize
\setlength{\tabcolsep}{5pt}
\renewcommand{\arraystretch}{1.08}
\begin{tabularx}{\textwidth}{
@{}
l
l
>{\raggedright\arraybackslash}p{0.335\textwidth}
>{\raggedright\arraybackslash}X
l
@{}
}
\toprule
 &
\textbf{Area} &
\textbf{Datasets} &
\textbf{Metrics} &
\textbf{Conditions} \\
\midrule

\textbf{Understanding} &
Image &
DynaMath~\cite{zou2025dynamath}, MathVista~\cite{lu2024mathvista},
MMBench~\cite{liu2024mmbench}, MME~\cite{fu2025mme},
MMMU~\cite{yue2024mmmu}, MMStar~\cite{chen2024mmstar},
ScienceQA~\cite{lu2022scienceqa} &
Accuracy ($\uparrow$);
circular accuracy (MMBench, $\uparrow$);
perception score (MME, $\uparrow$) &
$P$, $C$
\\

 &
Video &
Video-MME~\cite{fu2025videomme} (short, no subtitles),
TOMATO~\cite{shangguan2025tomato}, TimeBlind~\cite{li2026timeblind} &
Accuracy ($\uparrow$);
instance accuracy (TimeBlind, $\uparrow$) &
$P$, $C$, $S$
\\

\midrule

\textbf{Trustworthiness} &
Safety &
HarmBench~\cite{mazeika2024harmbench},
Video-SafetyBench (VSB)~\cite{liu2025videosafetybench} &
Attack success rate ($\downarrow$) &
$P$, $C$; $S$ on VSB
\\

 &
Fairness &
VMDT~\cite{potter2025vmdt} &
Mean fairness score ($\uparrow$) &
$P$, $C$, $S$
\\

 &
Privacy &
VMDT~\cite{potter2025vmdt} &
$100-$ weighted state, city, and ZIP-code
inference accuracy ($\uparrow$) &
$P$, $C$, $S$
\\

\midrule

\textbf{Generalization} &
Prompt robustness &
Video-MME~\cite{fu2025videomme}, MMBench~\cite{liu2024mmbench} &
Difference in accuracy drops ($\uparrow$) &
$P$, $C$
\\

 &
\begin{tabular}[t]{@{}l@{}}Fine-tuning\end{tabular} &
LingoQA~\cite{marcu2024lingoqa}, EyePCR~\cite{wang2025eyepcr} &
Difference in accuracy gains ($\uparrow$) &
$P$, $C$
\\

\bottomrule
\end{tabularx}
\vspace{-.1in}
\end{table*}

\subsection{Formalizing Streaming Tax}
\label{sec:formulation}

A VLM $f_\theta$ with an inference protocol $\mathcal{E}$ receives an input $\{q, V\}$ and returns an answer $\hat{y}$.
Here, $q$ denotes the textual query and $V=\{v_t\}_{t=1}^T$ is the visual input of $T$ frames: a video when $T>1$, a single image when $T=1$, and text-only when $T=0$.
$\mathrm{Score}(\theta,\mathcal{E})$ denotes a benchmark's metric (\Cref{tb-eval-setup}) computed on the resulting answers.

\noindent \textbf{Offline inference.}
The offline protocol $\mathcal{E}_{\mathrm{off}}$ processes the query together with sampled frames $V^\prime \subseteq V$ \citep{bai2025qwen3vl}, using an offline prompt $c_{\mathrm{off}}$ that provides the instructions and chat template for placing the frames and query in a single turn.
The output is:
\begin{equation}
\label{eq:offline-inference}
    \hat{y}=f_\theta\big(c_{\mathrm{off}},\,V^\prime,\,q\big).
\end{equation}

\noindent \textbf{Streaming inference.}
The streaming protocol $\mathcal{E}_{\mathrm{str}}$ replays the video in temporal order, using a streaming prompt $c_{\mathrm{str}}$ that can include a streaming system prompt and arranges the frames, earlier outputs, and the query as dialogue turns.
At each step $t$, a memory update $U$ updates the model memory as $m_t=U(m_{t-1},v_t,o_{t-1})$ by combining the available memory $m_{t-1}$, the frame $v_t$, and the previous output $o_{t-1}$.
A response policy $\rho$ then decides whether the model speaks, $r_t\sim\rho(\,\cdot\mid c_{\mathrm{str}},m_t)$, and the model emits an output $o_t$ only when $r_t=1$.
As in streaming video question answering~\cite{di2025rekv,wang2026thinkwhilewatching}, a query posed at time $\tau\le T$ is answered from what the model has retained by then, $m_\tau$.
Applying $\mathcal{E}_{\mathrm{str}}$ to an offline task refers to the case $\tau=T$ where the query follows the last frame~\cite{di2025rekv}:
\begin{equation}
\label{eq:streaming-answer}
\begin{aligned}
    \hat{y} &=
    \begin{cases}
        f_\theta\big(c_{\mathrm{str}},\,m_T,\,q\big) & r=1\\
        \varnothing & r=0
    \end{cases},\\
    r &\sim \rho(\,\cdot\mid c_{\mathrm{str}},\,m_T,\,q).
\end{aligned}
\end{equation}
Compared with Eq.~\eqref{eq:offline-inference}, Eq.~\eqref{eq:streaming-answer} draws on the retained memory $m_T$ instead of $V^\prime$, can be withheld, and is produced under a different prompt.

\noindent \textbf{Streaming tax.}
The streaming outputs are composed of several entangled factors.
In this paper, we formalize streaming tax by decomposing it into individual parts which allows for isolated analysis.
Specifically, streaming tax is decomposed as:
\begin{equation}
\label{eq:performance-gaps}
    \underbrace{S-P}_{\mathrm{Streaming~Tax}} = \underbrace{(C-P)}_{\mathrm{Adaptation}} + \underbrace{(S-C)}_{\mathrm{Execution}},
\end{equation}
where the performance of \emph{Parent Offline} ($P$, Eq. \eqref{eq:offline-inference}) and \emph{Child Offline} ($C$) share the offline protocol $\mathcal{E}_{\mathrm{off}}$, and \emph{Child Native Streaming} ($S$, Eq. \eqref{eq:streaming-answer}) runs the child under its native streaming protocol $\mathcal{E}_{\mathrm{str}}$, which is necessary when the model cannot store and re-encode the full video at question time, e.g., in long sessions; the child's offline mode requires retaining every frame.
The \emph{adaptation tax} isolates the change of weights from $\theta_P$ to $\theta_C$ under $\mathcal{E}_{\mathrm{off}}$, which may reflect a change in task capability or in how the streaming-trained child follows the prompt and produces an answer.
The \emph{execution tax} isolates the change of protocol from $\mathcal{E}_{\mathrm{off}}$ to $\mathcal{E}_{\mathrm{str}}$ for $\theta_C$, and thus the three differences between Eq.~\eqref{eq:streaming-answer} and Eq.~\eqref{eq:offline-inference}: the visual evidence available for answering, whether an answer is delivered, and the prompt.
The performance is computed as $P = \mathrm{Score}(\theta_P,\mathcal{E}_{\mathrm{off}}), C = \mathrm{Score}(\theta_C,\mathcal{E}_{\mathrm{off}}), \text{~and~} S = \mathrm{Score}(\theta_C,\mathcal{E}_{\mathrm{str}})$.

\subsection{Models}
\label{sec:families}
We study open-weight streaming models with identifiable offline parents for paired evaluation.
These models span different mechanisms of memory retention ($U$) and response control ($\rho$).
Concretely speaking, we evaluate $6$ streaming models paired with $5$ offline parents:
MMDuet~\cite{wang2025mmduet} with LLaVA-OneVision-7B~\citep{li2024llavaonevision};
MMDuet2~\cite{wang2025mmduet2} and ThinkStream~\cite{liu2026thinkstream} with Qwen2.5-VL-3B~\citep{bai2025qwen25vl};
AURA~\cite{lu2026aura} with Qwen3-VL-8B~\citep{bai2025qwen3vl};
LiveStar~\cite{yang2025livestar} with LiveStar-InternVideo-8B~\citep{yang2025livestar};
and MOSS-VL-Realtime (MOSS-VL-RT)~\cite{wang2026mossvl} with MOSS-VL-Instruct-0708~\citep{wang2026mossvl}.
The shared parent of MMDuet2 and ThinkStream enables comparisons between different streaming adaptations.
Appendix~\ref{supp:models} and~\ref{supp:implementation} describe each model and how it retains memory and decides when to respond.

\subsection{Evaluation Scope and Metrics}
\label{sec:protocol}

We evaluate along three complementary dimensions: \textit{understanding}, \textit{trustworthiness}, and \textit{generalization}, as summarized in \Cref{tb-eval-setup}.
Understanding measures performance on image and video tasks; trustworthiness examines safety, fairness, and privacy; and generalization assesses robustness to prompt perturbations and adaptability to downstream tasks.
Detailed dataset descriptions, evaluation subsets, and scoring procedures are in Appendix~\ref{sec-supp-eval-setup} and~\ref{supp:scoring}.

\textbf{Statistical Analysis.} 
We assess both the magnitude and uncertainty of performance differences.
We estimate $95\%$ confidence intervals using a paired cluster bootstrap~\cite{efron1993bootstrap,field2007bootstrapping}, resampling questions together when they belong to the same video or other benchmark-defined group.
We assess statistical significance using paired tests~\cite{mcnemar1947note} and apply Holm correction within groups of related comparisons \citep{holm1979simple}.
We interpret the size of each difference alongside its statistical significance.
Scoring procedures, detailed statistical settings, and additional equivalence analyses are provided in Appendix~\ref{supp:scoring}.

\section{Evaluation Results}
\label{sec:evaluation}

\begin{figure}[t!]
\centering
\input{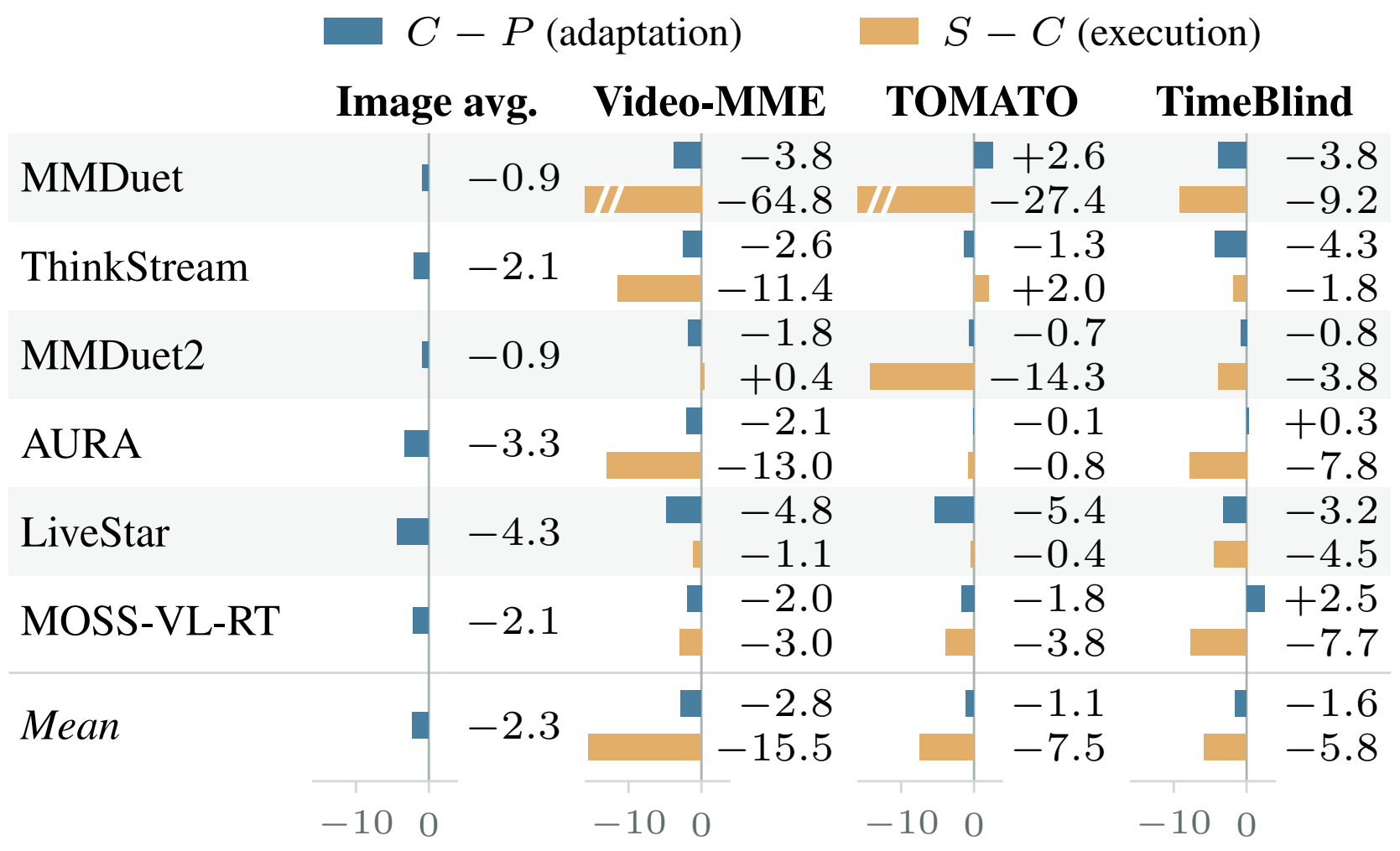}
\vspace{-.2in}
\caption{Adaptation and execution taxes on visual understanding, in points; negative values indicate losses. Blue bars show the adaptation tax on the mean of seven image benchmarks (Image avg.) and on three video benchmarks; orange bars show the execution tax. Mean is the unweighted mean over the six children. Bars that extend beyond the axis are truncated. Per-benchmark image results, absolute scores, and intervals are in Appendix~\ref{supp:understanding}.}
\label{fig:gaps}
\vspace{-.2in}
\end{figure}

\begin{table*}[t]
\centering
\caption{Retention of trustworthiness and generalization ability in streaming models. All differences are reported in pp. Negative values indicate worse performance on the reported metric. Safety columns report ASR reductions, so a negative value means a higher ASR. Grey marks nonsignificant differences after Holm correction. --: MMDuet gives no fairness ratings and no privacy answers under streaming because it stays silent. $^{\ast}$Mean of the other five children.}
\vspace{-.1in}
\label{tab:gaps}
\footnotesize
\setlength{\tabcolsep}{3pt}
\renewcommand{\arraystretch}{1.12}
\resizebox{\textwidth}{!}{%
\begin{tabular}{@{}l*{14}{c}@{}}
\toprule
 & \multicolumn{5}{c}{Safety} & \multicolumn{4}{c}{VMDT} & \multicolumn{3}{c}{Prompt robustness} & \multicolumn{2}{c}{Fine-tuning} \\
\cmidrule(lr){2-6}\cmidrule(lr){7-10}\cmidrule(lr){11-13}\cmidrule(lr){14-15}
 & HarmBench & \multicolumn{2}{c}{VSB harmful} & \multicolumn{2}{c}{VSB benign} & \multicolumn{2}{c}{Fairness} & \multicolumn{2}{c}{Privacy} & \multicolumn{3}{c}{$(C-P)_{\rm pert}-(C-P)_{\rm clean}$} & \multicolumn{2}{c}{Task mean} \\
\cmidrule(lr){2-2}\cmidrule(lr){3-4}\cmidrule(lr){5-6}\cmidrule(lr){7-8}\cmidrule(lr){9-10}\cmidrule(lr){11-13}\cmidrule(lr){14-15}
Model & $C-P$ & $C-P$ & $S-C$ & $C-P$ & $S-C$ & $C-P$ & $S-C$ & $C-P$ & $S-C$ & Typo & Paraphrase & Distraction & $\Delta$Gain & $\Delta$Acc \\
\midrule
MMDuet & \textcolor[HTML]{8A939B}{$-1.2$} & $-11.8$ & $+76.4$ & $+8.3$ & $+66.3$ & \textcolor[HTML]{8A939B}{$+0.0$} & -- & $-5.8$ & -- & \textcolor[HTML]{8A939B}{$+0.7$} & \textcolor[HTML]{8A939B}{$+0.7$} & \textcolor[HTML]{8A939B}{$+0.1$} & $+8.2$ & \textcolor[HTML]{8A939B}{$+0.0$} \\
ThinkStream & \textcolor[HTML]{8A939B}{$+3.9$} & $-16.6$ & $+16.2$ & $+9.2$ & $+21.3$ & \textcolor[HTML]{8A939B}{$-0.2$} & \textcolor[HTML]{8A939B}{$+2.3$} & $+5.1$ & $+4.9$ & \textcolor[HTML]{8A939B}{$+0.5$} & \textcolor[HTML]{8A939B}{$+1.9$} & \textcolor[HTML]{8A939B}{$+1.5$} & $+5.3$ & \textcolor[HTML]{8A939B}{$-0.3$} \\
MMDuet2 & $-7.3$ & $-24.5$ & $+49.8$ & \textcolor[HTML]{8A939B}{$+3.6$} & $+52.1$ & \textcolor[HTML]{8A939B}{$-2.6$} & \textcolor[HTML]{8A939B}{$-3.3$} & $+10.0$ & $-2.9$ & \textcolor[HTML]{8A939B}{$-0.2$} & \textcolor[HTML]{8A939B}{$+1.2$} & \textcolor[HTML]{8A939B}{$-0.2$} & \textcolor[HTML]{8A939B}{$+1.6$} & \textcolor[HTML]{8A939B}{$+0.1$} \\
AURA & \textcolor[HTML]{8A939B}{$-4.1$} & $-42.8$ & \textcolor[HTML]{8A939B}{$+0.4$} & $-35.8$ & $+5.8$ & $-3.7$ & \textcolor[HTML]{8A939B}{$-3.2$} & \textcolor[HTML]{8A939B}{$+3.1$} & $-8.4$ & \textcolor[HTML]{8A939B}{$+0.2$} & \textcolor[HTML]{8A939B}{$-0.5$} & \textcolor[HTML]{8A939B}{$+0.4$} & $+3.9$ & \textcolor[HTML]{8A939B}{$+1.1$} \\
LiveStar & $-8.8$ & $-28.4$ & $+12.5$ & $+14.7$ & \textcolor[HTML]{8A939B}{$+1.1$} & \textcolor[HTML]{8A939B}{$-1.8$} & $-16.9$ & $-7.9$ & \textcolor[HTML]{8A939B}{$-0.2$} & \textcolor[HTML]{8A939B}{$+0.3$} & \textcolor[HTML]{8A939B}{$-1.6$} & \textcolor[HTML]{8A939B}{$-0.8$} & \textcolor[HTML]{8A939B}{$-0.7$} & \textcolor[HTML]{8A939B}{$-0.3$} \\
MOSS-VL-Realtime & $-6.1$ & $+11.8$ & $+42.5$ & $-5.7$ & $+3.9$ & $+3.2$ & $-3.3$ & $+5.6$ & \textcolor[HTML]{8A939B}{$-2.4$} & \textcolor[HTML]{8A939B}{$+1.7$} & \textcolor[HTML]{8A939B}{$-0.9$} & \textcolor[HTML]{8A939B}{$-0.8$} & \textcolor[HTML]{8A939B}{$-1.1$} & \textcolor[HTML]{8A939B}{$-0.1$} \\
\midrule
\textit{Mean} & $-3.9$ & $-18.7$ & $+33.0$ & $-1.0$ & $+25.1$ & $-0.9$ & $-4.9^{\ast}$ & $+1.7$ & $-1.8^{\ast}$ & $+0.5$ & $+0.1$ & $+0.0$ & $+2.8$ & $+0.1$ \\
\bottomrule
\end{tabular}%
}
\end{table*}

\subsection{Understanding}
\label{sec:understanding}
Visual understanding is a core capability of VLMs.
To replace its offline parent on conventional visual tasks, a streaming model must retain its visual understanding after adaptation and use what it has seen to answer questions posed at the end of a stream.
We use the child's offline score as a common reference for these two comparisons.
The results are shown in \Cref{fig:gaps}.

First, \textbf{under matched offline protocol, streaming models generally score close to their parents}.
On three main video benchmarks, the mean gap to parents is less than $3$ percentage points (pp), with all six children weighted equally.
Averaging scores over the seven image understanding tasks gives child--parent gaps ranging from $0.9$ to $4.3$pp (\Cref{fig:gaps}; per-benchmark gaps in Appendix \Cref{fig:supp-image-gaps}).

Then, \textbf{the streaming execution widens the gap to the offline parent.}
In $16$ of $18$ model--benchmark comparisons, the child scores lower under streaming execution than under offline inference.
The decline is most consistent on TimeBlind, where instance accuracy falls for all six children.

Overall, the adapted children retain much of their parents' offline understanding performance.
When they process video as a stream before answering questions, the average gap to parents widens.
On each of the three video benchmarks, the mean streaming loss exceeds the mean loss from adaptation under matched offline evaluation.

\subsection{Trustworthiness}
\label{sec:trust}

\begin{figure}[t!]
\centering
\input{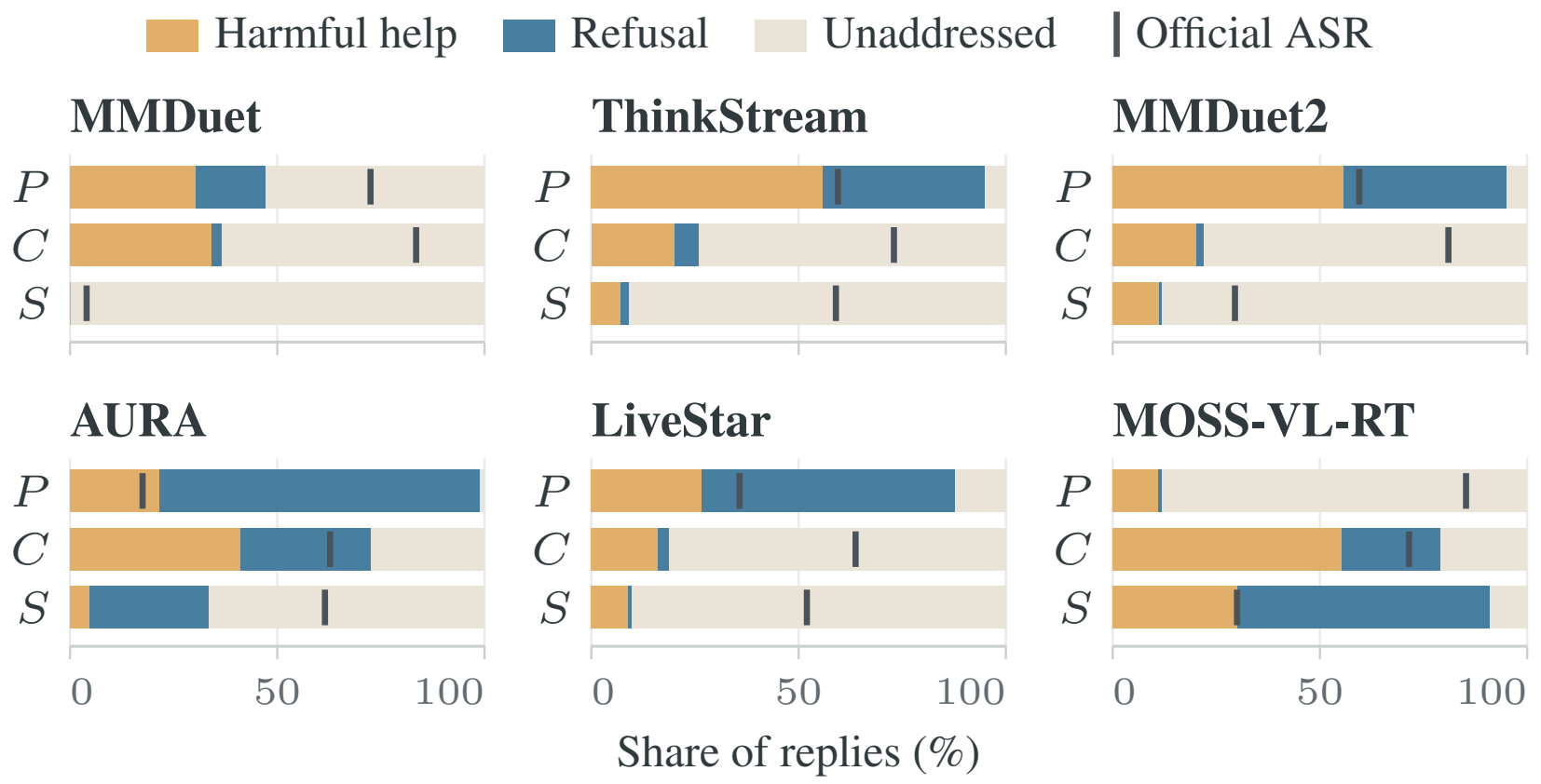}
\vspace{-.3in}
\caption{Distribution of response types across 400 harmful VSB requests. Bar segments show the percentage of requests in each response category (harmful assistance, refusal or safe deflection as defined in XSTest~\cite{rottger2024xstest}, and unaddressed requests). Dark ticks mark the official ASR on the same requests; the official judge also scores many off-task narrations of the harmful video as unsafe, so ASR can exceed the share of harmful assistance (Appendix~\ref{supp:trust}).}
\label{fig:composition}
\vspace{-.2in}
\end{figure}

Beyond visual understanding as a basic ability, a video assistant in practice should also refuse harmful requests, answer questions involving demographic attributes fairly, and protect sensitive information that could reveal a filming location.
We examine safety, fairness, and privacy by first comparing each child with its parent under matched offline inference and then comparing the same child under offline and streaming inference in \Cref{fig:composition} and \Cref{tab:gaps}.

First, as for safety, \textbf{under matched offline inference, most streaming models refuse harmful requests less often than their parents.}
On HarmBench and the harmful split of Video-SafetyBench (VSB), five of the six children have a higher attack success rate (ASR) than their parents on each benchmark.
On the $400$ audited VSB harmful requests, all five children with higher VSB ASR refuse or safely redirect fewer requests than their parents and leave more requests unaddressed.
As for streaming performance ($S$), \textbf{mean ASR falls, but primarily due to unaddressed requests rather than actual refusals.}
The drop is $33.0$pp on explicitly harmful VSB queries and $25.1$pp on benign queries whose harmful intent becomes apparent from the video.
On the audited requests, five children refuse or safely redirect no more often than offline and leave $66.5$--$99.8\%$ of them unaddressed, whereas MOSS-VL-RT refuses more (\Cref{fig:composition}).
Similarly, on a separate set of 40 harmless questions about the same videos, 5 of the 6 children answer fewer questions under streaming execution.

Second, as for fairness, \textbf{streaming execution widens the gap in average fairness scores between children and their offline parents.}
On VMDT, MMDuet answers 1 of the 2,000 requests under native streaming and receives no fairness score.
The other five children trail their offline parents by an average of $1.0$pp when evaluated offline and $5.9$pp under native streaming (\Cref{tab:gaps}).
Four of the five children score lower on overall fairness when they switch to streaming, and all four also score lower on the factual subtask; LiveStar accounts for most of the average widening (Appendix \Cref{tab:eval-vmdt}).

Finally, as for privacy, \textbf{most children disclose more location information under streaming inference than offline.}
On VMDT's location-inference task, models receive lower privacy scores when they identify filming locations more accurately.
Under matched offline inference, the six children score $1.7$pp higher than their parents on average, although MMDuet and LiveStar name the correct city more often than their respective parents (Appendix \Cref{tab:eval-vmdt}).
When the children switch to native streaming, MMDuet answers none of the 600 questions and receives no privacy score. Among the other five children, privacy scores fall for four of the five, with a mean decline of $1.8$pp across all five.

Overall, the most consistent safety change is that children refuse harmful requests less often.
Streaming execution also reduces responsiveness to harmless questions for most children, widens the average fairness gap to the offline parent, and increases the risk of location disclosure.

\subsection{Generalization}
\label{sec:generalization}
Generalization evaluates the model's performance in new environments or tasks.
In practice, users may phrase the same question in different ways or fine-tune a model for a new task.
We compare children and parents under matched offline evaluation to assess sensitivity to prompt changes and the ability to learn downstream tasks.

First, as shown in \Cref{tab:gaps}, under matched offline inference, \textbf{the children and their parents show similar sensitivity to prompt perturbations.}
On Video-MME and MMBench, we hold the visual input fixed and compute child accuracy minus parent accuracy for both the original and perturbed prompts~\cite{zhu2023promptrobust}.
We subtract the original difference from the perturbed difference, $(C-P)_{\mathrm{pert}}-(C-P)_{\mathrm{clean}}$.
Then, as for fine-tuning, \textbf{the children reach nearly the same average accuracy as their parents on downstream tasks.}
On LingoQA~\cite{marcu2024lingoqa} and EyePCR~\cite{wang2025eyepcr}, we fine-tune each child and its parent on the same sets of $\{1000,2000,5000\}$ examples.
We use $\Delta\mathrm{Gain}$ to compare gains in accuracy over each model's own zero-shot baseline and $\Delta\mathrm{Acc}$ to compare final accuracies after fine-tuning.
Averaging equally across the six pairs, two tasks, and three training budgets, we find that the children start $2.8$pp below their parents, gain $2.8$pp more from fine-tuning, and finish $0.1$pp ahead.

Overall, prompt perturbations have similar effects on child and parent accuracy under matched offline evaluation.
Across the two downstream tasks, the children trail their parents in average accuracy before fine-tuning and nearly match them after both are fine-tuned on the same examples.

\section{Diagnosing the Streaming Tax}
\label{sec:attribution}

To understand how the streaming taxes arise, in this section, we provide comprehensive analysis by examining what information the model can use when a question arrives, what determines whether it answers, and whether its response is consistent with what it can recognize about the request.

\begin{figure*}[t]
\centering
\begingroup%
\definecolor{chfiveblue}{HTML}{487FA1}%
\definecolor{chfiveorange}{HTML}{E1AF69}%
\definecolor{chfiveoutline}{HTML}{986420}%
\definecolor{chfiveink}{HTML}{283439}%
\definecolor{chfivemuted}{HTML}{52636A}%
\definecolor{chfivegrid}{HTML}{DCE3E5}%
\definecolor{chfivezero}{HTML}{9CAAAE}%
\colorlet{chfivestripe}{chfivegrid!32}%
\begin{tikzpicture}[x=1pt,y=-1pt,text=chfiveink,
  every node/.style={inner sep=0pt,outer sep=0pt,font=\fontsize{7}{8.4}\selectfont},
  title/.style={anchor=base west,font=\fontsize{8.5}{10}\selectfont\bfseries},
  subhead/.style={anchor=base west,font=\fontsize{7.5}{9}\selectfont\bfseries},
  model label/.style={anchor=base west,font=\fontsize{7.5}{9}\selectfont},
  note/.style={text=chfivemuted,font=\fontsize{6.5}{7.5}\selectfont},
  key/.style={anchor=base west},
  value/.style={anchor=base east},
  grid line/.style={chfivegrid,line width=.4pt},
  whisker/.style={line width=.95pt},
  neither/.style={chfivemuted,line width=.8pt},
  bare/.style={circle,draw=chfiveblue,fill=white,line width=.95pt,minimum size=4.5pt},
  cue/.style={rectangle,rotate=45,draw=chfiveoutline,fill=chfiveorange,line width=.75pt,minimum size=3.6pt}]
  \useasboundingbox (0,0) rectangle (496.00,116.35);
  \draw[chfivegrid,line width=.65pt] (181.80,2) -- (181.80,114.35);
  \node[title] at (2.00,8.00) {(a) Access to earlier frames};
  \node[subhead] at (2.00,20.90) {$C_W-S$};
  \node[subhead] at (2.00,66.90) {$S_{\mathrm{full}}-C$};
  \fill[chfivestripe] (0,24.25) rectangle (171.80,35.75);
  \fill[chfivestripe] (0,70.25) rectangle (171.80,81.75);
  \draw[chfivegrid,line width=.5pt] (0,47.25) -- (171.80,47.25);
  \draw[grid line] (52.20,24.25) -- (52.20,47.25);
  \draw[grid line] (52.20,70.25) -- (52.20,93.25);
  \draw[grid line] (82.20,24.25) -- (82.20,47.25);
  \draw[grid line] (82.20,70.25) -- (82.20,93.25);
  \draw[grid line] (142.20,24.25) -- (142.20,47.25);
  \draw[grid line] (142.20,70.25) -- (142.20,93.25);
  \draw[chfivezero,line width=.7pt] (112.20,24.25) -- (112.20,47.25);
  \draw[chfivezero,line width=.7pt] (112.20,70.25) -- (112.20,93.25);
  \node[model label] at (2.00,32.40) {AURA};
  \draw[chfiveblue,whisker] (106.62,30.00) -- (130.62,30.00);
  \draw[chfiveblue,line width=.75pt] (106.62,28.40) -- (106.62,31.60);
  \draw[chfiveblue,line width=.75pt] (130.62,28.40) -- (130.62,31.60);
  \node[bare] at (118.62,30.00) {};
  \node[value,text=chfiveblue] at (169.80,32.40) {$+1.1$};
  \node[model label] at (2.00,43.90) {ThinkStream};
  \draw[chfiveblue,whisker] (77.82,41.50) -- (119.40,41.50);
  \draw[chfiveblue,line width=.75pt] (77.82,39.90) -- (77.82,43.10);
  \draw[chfiveblue,line width=.75pt] (119.40,39.90) -- (119.40,43.10);
  \node[bare] at (98.58,41.50) {};
  \node[value,text=chfiveblue] at (169.80,43.90) {$-2.3$};
  \node[model label] at (2.00,78.40) {AURA};
  \draw[chfiveoutline,whisker] (103.38,76.00) -- (124.20,76.00);
  \draw[chfiveoutline,line width=.75pt] (103.38,74.40) -- (103.38,77.60);
  \draw[chfiveoutline,line width=.75pt] (124.20,74.40) -- (124.20,77.60);
  \node[cue] at (113.82,76.00) {};
  \node[value,text=chfiveoutline] at (169.80,78.40) {$+0.3$};
  \node[model label] at (2.00,89.90) {ThinkStream};
  \draw[chfiveoutline,whisker] (50.58,87.50) -- (85.80,87.50);
  \draw[chfiveoutline,line width=.75pt] (50.58,85.90) -- (50.58,89.10);
  \draw[chfiveoutline,line width=.75pt] (85.80,85.90) -- (85.80,89.10);
  \node[cue] at (68.22,87.50) {};
  \node[value,text=chfiveoutline] at (169.80,89.90) {$-7.3$};
  \draw[chfivegrid,line width=.5pt] (49.20,93.25) -- (145.20,93.25);
  \draw[chfivezero,line width=.5pt] (52.20,93.25) -- (52.20,95.45);
  \node[note,anchor=base] at (52.20,102.85) {$-10$};
  \draw[chfivezero,line width=.5pt] (82.20,93.25) -- (82.20,95.45);
  \node[note,anchor=base] at (82.20,102.85) {$-5$};
  \draw[chfivezero,line width=.5pt] (112.20,93.25) -- (112.20,95.45);
  \node[note,anchor=base] at (112.20,102.85) {$0$};
  \draw[chfivezero,line width=.5pt] (142.20,93.25) -- (142.20,95.45);
  \node[note,anchor=base] at (142.20,102.85) {$+5$};
  \node[anchor=base] at (97.20,113.85) {Accuracy difference (pp)};
  \node[title] at (193.80,8.00) {(b) Harm-direction scores before the answer};
  \node[value,text=chfivemuted] at (252.40,20.90) {$n$};
  \node[bare] at (386.60,5.60) {};
  \node[key] at (395.60,8.00) {After request};
  \draw[chfiveoutline,whisker] (381.60,18.50) -- (391.60,18.50);
  \draw[chfiveoutline,line width=.75pt] (381.60,16.90) -- (381.60,20.10);
  \draw[chfiveoutline,line width=.75pt] (391.60,16.90) -- (391.60,20.10);
  \node[cue] at (386.60,18.50) {};
  \node[key] at (395.60,20.90) {Before generation};
  \draw[neither] (444.40,1.40) -- (444.40,9.80);
  \node[key] at (449.40,8.00) {Neither assisted};
  \fill[chfivestripe] (191.80,24.25) rectangle (496.00,35.75);
  \fill[chfivestripe] (191.80,47.25) rectangle (496.00,58.75);
  \fill[chfivestripe] (191.80,70.25) rectangle (496.00,81.75);
  \draw[grid line] (263.40,24.25) -- (263.40,93.25);
  \draw[grid line] (293.40,24.25) -- (293.40,93.25);
  \draw[grid line] (323.40,24.25) -- (323.40,93.25);
  \draw[grid line] (353.40,24.25) -- (353.40,93.25);
  \draw[grid line] (383.40,24.25) -- (383.40,93.25);
  \draw[grid line] (443.40,24.25) -- (443.40,93.25);
  \draw[chfivezero,line width=.7pt] (413.40,24.25) -- (413.40,93.25);
  \node[model label] at (193.80,32.40) {MMDuet};
  \node[value,text=chfivemuted] at (252.40,32.40) {$33$};
  \draw[neither] (368.92,25.80) -- (368.92,34.20);
  \draw[chfivezero,line width=.8pt] (408.82,30.00) -- (365.22,30.00);
  \draw[chfiveoutline,whisker] (353.76,30.00) -- (376.44,30.00);
  \draw[chfiveoutline,line width=.75pt] (353.76,28.40) -- (353.76,31.60);
  \draw[chfiveoutline,line width=.75pt] (376.44,28.40) -- (376.44,31.60);
  \node[bare] at (408.82,30.00) {};
  \node[cue] at (365.22,30.00) {};
  \node[value,text=chfiveoutline] at (494.00,32.40) {$-0.32$};
  \node[model label] at (193.80,43.90) {ThinkStream};
  \node[value,text=chfivemuted] at (252.40,43.90) {$10$};
  \draw[neither] (349.90,37.30) -- (349.90,45.70);
  \draw[chfivezero,line width=.8pt] (412.84,41.50) -- (364.83,41.50);
  \draw[chfiveoutline,whisker] (332.56,41.50) -- (395.32,41.50);
  \draw[chfiveoutline,line width=.75pt] (332.56,39.90) -- (332.56,43.10);
  \draw[chfiveoutline,line width=.75pt] (395.32,39.90) -- (395.32,43.10);
  \node[bare] at (412.84,41.50) {};
  \node[cue] at (364.83,41.50) {};
  \node[value,text=chfiveoutline] at (494.00,43.90) {$-0.32$};
  \node[model label] at (193.80,55.40) {MMDuet2};
  \node[value,text=chfivemuted] at (252.40,55.40) {$10$};
  \draw[neither] (293.73,48.80) -- (293.73,57.20);
  \draw[chfivezero,line width=.8pt] (410.04,53.00) -- (292.74,53.00);
  \draw[chfiveoutline,whisker] (265.84,53.00) -- (319.54,53.00);
  \draw[chfiveoutline,line width=.75pt] (265.84,51.40) -- (265.84,54.60);
  \draw[chfiveoutline,line width=.75pt] (319.54,51.40) -- (319.54,54.60);
  \node[bare] at (410.04,53.00) {};
  \node[cue] at (292.74,53.00) {};
  \node[value,text=chfiveoutline] at (494.00,55.40) {$-0.80$};
  \node[model label] at (193.80,66.90) {AURA};
  \node[value,text=chfivemuted] at (252.40,66.90) {$41$};
  \draw[neither] (329.22,60.30) -- (329.22,68.70);
  \draw[chfivezero,line width=.8pt] (413.04,64.50) -- (299.04,64.50);
  \draw[chfiveoutline,whisker] (286.92,64.50) -- (311.55,64.50);
  \draw[chfiveoutline,line width=.75pt] (286.92,62.90) -- (286.92,66.10);
  \draw[chfiveoutline,line width=.75pt] (311.55,62.90) -- (311.55,66.10);
  \node[bare] at (413.04,64.50) {};
  \node[cue] at (299.04,64.50) {};
  \node[value,text=chfiveoutline] at (494.00,66.90) {$-0.76$};
  \node[model label] at (193.80,78.40) {LiveStar};
  \node[value,text=chfivemuted] at (252.40,78.40) {$16$};
  \draw[neither] (379.12,71.80) -- (379.12,80.20);
  \draw[chfivezero,line width=.8pt] (368.37,76.00) -- (365.62,76.00);
  \draw[chfiveoutline,whisker] (351.03,76.00) -- (381.22,76.00);
  \draw[chfiveoutline,line width=.75pt] (351.03,74.40) -- (351.03,77.60);
  \draw[chfiveoutline,line width=.75pt] (381.22,74.40) -- (381.22,77.60);
  \node[bare] at (368.37,76.00) {};
  \node[cue] at (365.62,76.00) {};
  \node[value,text=chfiveoutline] at (494.00,78.40) {$-0.32$};
  \node[model label] at (193.80,89.90) {MOSS-VL-RT};
  \node[value,text=chfivemuted] at (252.40,89.90) {$91$};
  \draw[neither] (457.77,83.30) -- (457.77,91.70);
  \draw[chfivezero,line width=.8pt] (397.27,87.50) -- (441.48,87.50);
  \draw[chfiveoutline,whisker] (436.56,87.50) -- (446.49,87.50);
  \draw[chfiveoutline,line width=.75pt] (436.56,85.90) -- (436.56,89.10);
  \draw[chfiveoutline,line width=.75pt] (446.49,85.90) -- (446.49,89.10);
  \node[bare] at (397.27,87.50) {};
  \node[cue] at (441.48,87.50) {};
  \node[value,text=chfiveoutline] at (494.00,89.90) {$+0.19$};
  \draw[chfivegrid,line width=.5pt] (260.40,93.25) -- (461.40,93.25);
  \draw[chfivezero,line width=.5pt] (263.40,93.25) -- (263.40,95.45);
  \node[note,anchor=base] at (263.40,102.85) {$-1.0$};
  \draw[chfivezero,line width=.5pt] (293.40,93.25) -- (293.40,95.45);
  \node[note,anchor=base] at (293.40,102.85) {$-0.8$};
  \draw[chfivezero,line width=.5pt] (323.40,93.25) -- (323.40,95.45);
  \node[note,anchor=base] at (323.40,102.85) {$-0.6$};
  \draw[chfivezero,line width=.5pt] (353.40,93.25) -- (353.40,95.45);
  \node[note,anchor=base] at (353.40,102.85) {$-0.4$};
  \draw[chfivezero,line width=.5pt] (383.40,93.25) -- (383.40,95.45);
  \node[note,anchor=base] at (383.40,102.85) {$-0.2$};
  \draw[chfivezero,line width=.5pt] (413.40,93.25) -- (413.40,95.45);
  \node[note,anchor=base] at (413.40,102.85) {$0$};
  \draw[chfivezero,line width=.5pt] (443.40,93.25) -- (443.40,95.45);
  \node[note,anchor=base] at (443.40,102.85) {$+0.2$};
  \node[anchor=base] at (360.90,113.85) {Calibrated harm-direction score (child minus parent)};
\end{tikzpicture}%
\endgroup
\vspace{-.2in}
\caption{Access to earlier frames in video QA and harm scores before answering.
(a)~On Video-MME, the top rows show $C_W-S$, the child's offline accuracy using only the streaming window minus its native streaming accuracy. The bottom rows show $S_{\mathrm{full}}-C$, its streaming accuracy after restoring the full history minus its offline accuracy.
(b)~For VSB requests on which only the child provided harmful assistance, circles and diamonds show the mean child--parent difference in calibrated harm scores at the end of the request and just before generation, respectively. Grey ticks show the pre-generation difference on requests where neither model provided harmful assistance. Whiskers show 95\% confidence intervals.}
\label{fig:access}
\vspace{-.15in}
\end{figure*}
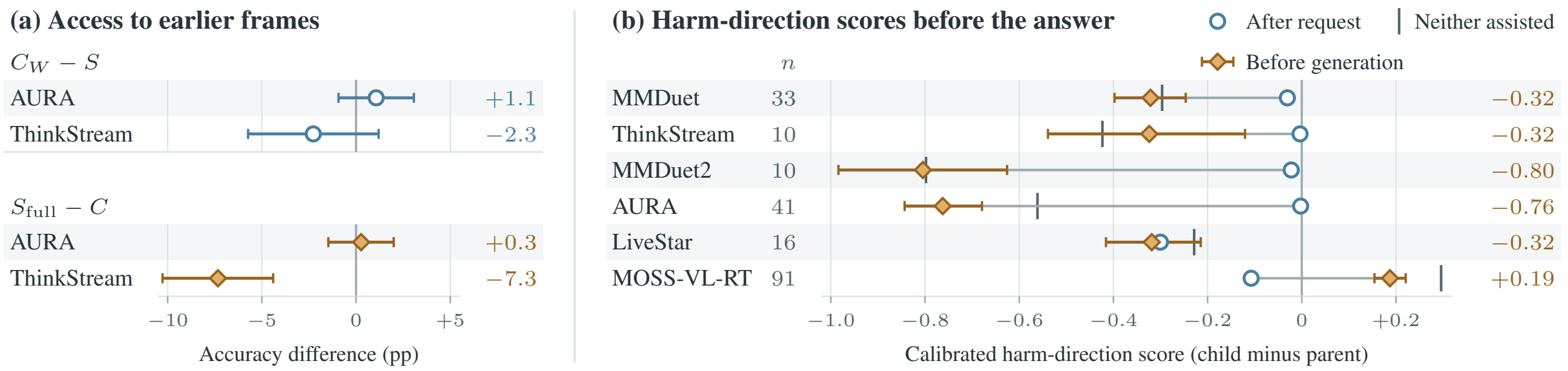

\subsection{Visual Evidence Available for Answering}
\label{sec:evidence-loss}

A model's visual memory determines which past events it can still inspect when a user asks about a completed video.
AURA and ThinkStream both discard older frames in their streaming evaluation pipelines and show substantial drops on Video-MME when switching from offline to streaming inference (Fig.~\ref{fig:gaps}).
Their memory update evicts visual input older than a window of $W$ steps, so $m_T$ holds frames only from $v_{T-W+1},\dots,v_T$.
To test whether the loss of earlier frames contributed to these drops, we evaluate two controls,
\begin{equation}
\label{eq:window-controls}
\begin{aligned}
    C_W &= \mathrm{Score}\big(\theta_C,\,\mathcal{E}_{\mathrm{off}}[V^\prime\leftarrow V^\prime_W]\big),\\
    S_{\mathrm{full}} &= \mathrm{Score}\big(\theta_C,\,\mathcal{E}_{\mathrm{str}}[U\leftarrow U_{\mathrm{full}}]\big),
\end{aligned}
\end{equation}
where $\mathcal{E}[\,\cdot\leftarrow\cdot\,]$ replaces one component of $\mathcal{E}$ and keeps the rest, $V^\prime_W$ contains only frames from the streaming window, $v_{T-W+1},\dots,v_T$, and $U_{\mathrm{full}}$ evicts nothing.
If the window explains the execution tax, $C_W$ should fall to $S$ and $S_{\mathrm{full}}$ should rise to $C$.
On a 750-question subset of Video-MME, $C_W$ came within $2.3$pp of $S$ for both models (\Cref{fig:access}a).
On the same questions, $S_{\mathrm{full}}$ came within $0.3$pp of $C$ for AURA, whereas ThinkStream's $S_{\mathrm{full}}$ remained $7.3$pp below $C$: restoring the full history recovers only part of its loss.

Restoring history can increase both the number of frames available and the span of video they cover.
To test whether coverage itself mattered, we fixed the memory budget of AURA and ThinkStream at its native size on the same 750 questions and spent part of it on segments sampled from earlier in the video.
With the budget unchanged, both models answered more questions correctly than with a recent-only window (Appendix~\ref{supp:diagnostics}).
Loss of access to earlier events can therefore account for a drop in retrospective QA accuracy when a model switches to streaming inference.
More broadly, a streaming model cannot retain all of its input as the video grows, nor can it know in advance what a later question will ask.
How it spends a limited memory budget is therefore central to its usefulness.

\subsection{Response Behavior}
\label{sec:response-policy}

Even when the retained evidence is sufficient, the streaming response policy determines whether the user receives an answer.
MMDuet's response heads rarely trigger a reply, which accounts for its execution tax on Video-MME (\Cref{fig:gaps}): forcing an answer from the same memory ($\rho\leftarrow 1$) raises its accuracy from $0.8\%$ to $66.0\%$, the level of its offline accuracy ($65.6\%$).

The response decision also depends on the prompt, since $r\sim\rho(\,\cdot\mid c_{\mathrm{str}},\,m_T,\,q)$ in Eq.~\eqref{eq:streaming-answer}.
Removing the streaming system prompt eliminated silent replies on the tested MMDuet2 fairness questions and reduced AURA's deferrals from $62\%$ to $2\%$ on 400 harmful VSB requests (Appendix~\ref{supp:diagnostics}).
The same prompt, however, does not make AURA's parent defer, so the streaming model acquired this undesirable response to the streaming prompt during adaptation.
We conjecture that this over-silence stems from the response--silence imbalance of streaming training data, in which silent turns far outnumber replies.

\subsection{Consistency of Recognition and Response}
\label{sec:refusal-behavior}

A model can miss a question due to unavailable earlier evidence or the response policy that keeps it silent.
When the model does answer with available evidence, does the reply reflect what it can recognize from that input?
We study this relationship mainly using harmful video requests, for which we can compare a model's explicit judgment of risk with its response when asked to help, and report the corresponding analysis for visual understanding in Appendix~\ref{supp:diagnostics}.

Under matched offline inference, five of the six children refused or safely redirected fewer harmful video requests than their parents (Sec.~\ref{sec:trust}).
Keeping each video and request fixed, we separately asked each model whether fulfilling the request would be harmful.
Overall, parents and children both recognize the requests as harmful: on 200 VSB requests, MMDuet, MMDuet2, AURA, and LiveStar also judged harmful $75\%$ to $97\%$ of the requests their respective parents judged harmful.
These four children's refusal rates were $14$--$60.5$pp below their parents, and $77$--$98\%$ of this decline came from requests that the child judged harmful but still did not refuse (Appendix~\ref{supp:safety-probes}).

We also compared the models' hidden states.
For each model and token position, let $h(x)$ be the hidden state on request $x$, and let $\mu^{\mathrm{harm}}$ and $\mu^{\mathrm{safe}}$ be the mean hidden states on harmful and harmless calibration prompts.
Their difference $\mu=\mu^{\mathrm{harm}}-\mu^{\mathrm{safe}}$ is the difference-in-means direction used to locate harmfulness and refusal \cite{arditi2024refusal,zhao2025harmfulness}.
We define the harm score $z(x)$ as the projection of $h(x)$ onto $\mu$, rescaled so that the harmless and harmful calibration means score $-1$ and $+1$: $z(x)=\frac{2\,\big\langle h(x)-\mu^{\mathrm{safe}},\,\mu\big\rangle}{\lVert \mu\rVert^{2}}-1.$
We average $z(x)$ over four layers selected on held-out prompts (Appendix~\ref{supp:safety-probes}) and compare a child with its parent.
We read it at the final user token, once the model has read the request, and at the last token before generation.
For most children, the child and its parent score the request almost identically after reading it, but just before generation the child's harm score is clearly lower than its parent's; that is, its internal state treats the request as less harmful than its parent's does (\Cref{fig:access}b).
The models' own judgments, the refusal rates, and the hidden states all point to the same conclusion: \textit{after adaptation, the child still recognizes the risk similarly to its parent, and the change occurs between recognizing the request and responding to it.}

We further find that the video context alone can trigger this change, even without video content.
Given only the text request, some children refused harmful requests as often as their parents (within $0$ to $3$pp); adding only the sentence ``These are frames of a video.'' before the same text request widened the refusal-rate gap by $12.1$--$26.3$pp in five of the six pairs (Appendix~\Cref{tab:supp-refusal-context}).
This suggests that streaming post-training induces a ``conditioned reflex'': once a video-watching context is detected, the model shifts into an overly compliant state that may bypass its original safety guardrails.

\section{Improving Streaming VLMs}
\label{sec:interventions}

Previous evaluation and analysis demonstrated the gap between current streaming VLMs and their parents on offline tasks.
This section proposes \emph{Stream++}, a training-free plugin to improve streaming VLMs on offline tasks while maintaining their streaming performance.

\subsection{Designing Stream++}

The idea behind Stream++ is to repair the streaming answer at the places where \S\ref{sec:attribution} found it to fail, at inference time and with the child's weights unchanged.
In Eq.~\eqref{eq:streaming-answer}, the answer depends on the memory $m_T$, the response policy $\rho$, and the generation $f_{\theta_C}$.
The memory can lose the earlier frames a question needs (\S\ref{sec:evidence-loss}), the response policy can withhold the answer (\S\ref{sec:response-policy}), and the child can fulfil a request that it recognizes as harmful (\S\ref{sec:refusal-behavior}).
Stream++ has one component for each: \texttt{Recall} changes which frames the memory keeps, \texttt{Respond} makes sure the answer is delivered, and \texttt{Refuse} restores the parent's refusal of harmful requests.

First, \texttt{Recall} replaces the memory update $U$ with $U_{\mathrm{rec}}$ without enlarging the memory: the budget $B$ is the child's own native visual budget, $|m_t|\le B$.
While the frames fit in memory ($t\le B$), it keeps all of them. Once the budget is reached, it keeps the most recent $k$ frames and fills the remaining $B-k$ slots with a set $\mathcal{A}_t$ of frames sampled uniformly from earlier in the video:
\begin{equation}
m_t=\{v_{t-k+1},\dots,v_t\}\cup\mathcal{A}_t,
\end{equation}
where $\mathcal{A}_t\subseteq\{v_1,\dots,v_{t-k}\}$ and $|\mathcal{A}_t|=B-k$.
\texttt{Recall} updates $\mathcal{A}_t$ as each frame arrives, so it needs neither the question nor the video's length in advance.
We set $k=10$ (other settings in Appendix \ref{supp:intervention-ablations}). The budget $B$ is $30$ one-second segments for AURA and $20$ chunks for ThinkStream; the other four children discard no frames natively, so \texttt{Recall} does not act on them.

Then, \texttt{Respond} replaces the response policy at the answer step with $\rho\leftarrow 1$: the child's silence marker may not open the answer, while a refusal still may.
It requests a final answer using the frames retained by \texttt{Recall}, taking the most probable option when a closed-choice reply has none.

Finally, \texttt{Refuse} combines two mechanisms for generation.
The first, \emph{parent-restoration steering} (PRS), moves the child's hidden state toward its parent's.
Let $h_P(x)$ and $h_C(x)$ denote the hidden states of the parent and the child at a given layer, read at the last token before generation.
Over the calibration set $\mathcal{D}_{\mathrm{cal}}$ (the harmful and harmless calibration prompts of \S\ref{sec:refusal-behavior}, together with video inputs that carry generic requests), we estimate a steering direction $d$ from these paired states and apply it to the child's hidden state:
\begin{equation}
\tilde{h}_C = h_C + \alpha\,d,
\end{equation}
where $d = \frac{1}{|\mathcal{D}_{\mathrm{cal}}|}
   \sum_{x\in\mathcal{D}_{\mathrm{cal}}}
   \bigl(h_P(x)-h_C(x)\bigr)$.
We select the layer and the steering strength $\alpha$ on held-out generic prompts by minimizing the divergence between the steered child's and the parent's next-token distributions.
Because a prompt can reduce deferrals while harmful compliance persists (Appendix \ref{supp:diagnostics}), \texttt{Refuse} also makes an explicit refusal decision.
Its second mechanism, a \emph{refusal gate}, scores each request along the child's harm direction $\mu_C$ (\S\ref{sec:refusal-behavior}), read at a single layer and at the same token position~\cite{zhao2025harmfulness}, as $s(x)=\langle h_C(x),\mu_C\rangle/\lVert\mu_C\rVert$.
We calibrate the threshold $\kappa$ on harmless requests disjoint from the test sets. If $s(x) > \kappa$, the gate returns a fixed refusal $y_{\mathrm{ref}}$; otherwise, the child returns the answer generated with PRS.

Putting these together, the output of Stream++ is
\begin{equation}
\label{eq:plugin-answer}
\hat{y}_{\mathrm{plug}}=
\begin{cases}
y_{\mathrm{ref}} & s(x)>\kappa\\
\tilde{f}_{\theta_C}\bigl(c_{\mathrm{str}},\,m_T,\,q\bigr) & \text{otherwise,}
\end{cases}
\end{equation}
where $m_T$ is produced by $U_{\mathrm{rec}}$, $\tilde{f}$ denotes generation with PRS, and $\rho\leftarrow 1$ ensures that an answer is always delivered. Only the child runs at deployment.

\begin{figure}[t]
\centering
\begingroup%
\definecolor{pmbase}{HTML}{487FA1}%
\definecolor{pmours}{HTML}{E1AF69}%
\definecolor{pmoutline}{HTML}{986420}%
\definecolor{pmrule}{HTML}{D2D8D7}%
\definecolor{pmgrid}{HTML}{DCE3E5}%
\definecolor{pmzero}{HTML}{9CAAAE}%
\definecolor{pmstripe}{HTML}{F5F7F6}%
\definecolor{pmink}{HTML}{283439}%
\definecolor{pmmuted}{HTML}{52636A}%
\begin{tikzpicture}[x=1pt,y=-1pt,text=pmink,
  every node/.style={inner sep=0pt,outer sep=0pt,font=\fontsize{7}{8.4}\selectfont},
  lab/.style={anchor=base west,font=\fontsize{7.5}{9}\selectfont},
  key/.style={anchor=base west},
  val/.style={anchor=base east},
  note/.style={text=pmmuted,font=\fontsize{6.5}{7.5}\selectfont},
  before/.style={circle,draw=pmbase,fill=white,line width=.95pt,minimum size=4.5pt},
  after/.style={circle,draw=pmoutline,fill=pmours,line width=.6pt,minimum size=4.6pt}]
  \useasboundingbox (0,0) rectangle (237.00,108.10);
  \fill[pmstripe] (0,0.50) rectangle (237.00,12.00);
  \fill[pmstripe] (0,23.50) rectangle (237.00,35.00);
  \fill[pmstripe] (0,50.50) rectangle (237.00,62.00);
  \fill[pmstripe] (0,73.50) rectangle (237.00,85.00);
  \draw[pmgrid,line width=.4pt] (87.46,0.50) -- (87.46,85.00);
  \draw[pmgrid,line width=.4pt] (150.95,0.50) -- (150.95,85.00);
  \draw[pmgrid,line width=.4pt] (182.69,0.50) -- (182.69,85.00);
  \draw[pmzero,line width=.7pt] (119.20,0.50) -- (119.20,85.00);
  \draw[pmrule,line width=.5pt] (0,37.00) -- (237.00,37.00);
  \filldraw[fill=white,draw=pmrule,line width=.4pt,rounded corners=1.5pt] (128.70,6.35) rectangle (201.00,29.15);
  \node[before] at (135.70,12.95) {};
  \node[key] at (141.70,15.35) {Native streaming $S$};
  \node[after] at (135.70,22.55) {};
  \node[key] at (141.70,24.95) {Stream++ $S_{\mathrm{plug}}$};
  \node[lab] at (2.00,8.65) {Video-MME};
  \draw[pmours!55,line width=2.2pt] (61.67,6.25) -- (101.32,6.25);
  \node[before] at (61.67,6.25) {};
  \node[after] at (101.32,6.25) {};
  \node[val] at (235.00,8.65) {$+12.5$};
  \node[lab] at (2.00,20.15) {TOMATO};
  \draw[pmours!55,line width=2.2pt] (92.54,17.75) -- (111.44,17.75);
  \node[before] at (92.54,17.75) {};
  \node[after] at (111.44,17.75) {};
  \node[val] at (235.00,20.15) {$+6.0$};
  \node[lab] at (2.00,31.65) {TimeBlind};
  \draw[pmours!55,line width=2.2pt] (95.64,29.25) -- (97.38,29.25);
  \node[before] at (95.64,29.25) {};
  \node[after] at (97.38,29.25) {};
  \node[val] at (235.00,31.65) {$+0.5$};
  \node[lab] at (2.00,47.15) {VSB harmful};
  \draw[pmours!55,line width=2.2pt] (165.01,44.75) -- (190.03,44.75);
  \node[before] at (165.01,44.75) {};
  \node[after] at (190.03,44.75) {};
  \node[val] at (235.00,47.15) {$+7.9$};
  \node[lab] at (2.00,58.65) {VSB benign};
  \draw[pmours!55,line width=2.2pt] (195.34,56.25) -- (197.72,56.25);
  \node[before] at (195.34,56.25) {};
  \node[after] at (197.72,56.25) {};
  \node[val] at (235.00,58.65) {$+0.8$};
  \node[lab] at (2.00,70.15) {VMDT fairness};
  \draw[pmours!55,line width=2.2pt] (99.03,67.75) -- (111.20,67.75);
  \node[before] at (99.03,67.75) {};
  \node[after] at (111.20,67.75) {};
  \node[val] at (235.00,70.15) {$+3.8$};
  \node[lab] at (2.00,81.65) {VMDT privacy};
  \draw[pmours!55,line width=2.2pt] (123.43,79.25) -- (141.42,79.25);
  \node[before] at (123.43,79.25) {};
  \node[after] at (141.42,79.25) {};
  \node[val] at (235.00,81.65) {$+5.7$};
  \draw[pmrule,line width=.5pt] (57.30,85.00) -- (206.50,85.00);
  \draw[pmzero,line width=.5pt] (87.46,85.00) -- (87.46,87.20);
  \node[note,anchor=base] at (87.46,94.60) {$-10$};
  \draw[pmzero,line width=.5pt] (119.20,85.00) -- (119.20,87.20);
  \node[note,anchor=base] at (119.20,94.60) {$0$};
  \draw[pmzero,line width=.5pt] (150.95,85.00) -- (150.95,87.20);
  \node[note,anchor=base] at (150.95,94.60) {$+10$};
  \draw[pmzero,line width=.5pt] (182.69,85.00) -- (182.69,87.20);
  \node[note,anchor=base] at (182.69,94.60) {$+20$};
  \node[val,note] at (235.00,94.60) {Change};
  \node[anchor=base] at (131.90,105.60) {Mean difference from offline parent (pp)};
\end{tikzpicture}%
\endgroup
\vspace{-.2in}
\caption{Effect of Stream++, averaged over the six children (five for VMDT, where MMDuet has no native streaming score). Hollow and filled markers show the mean difference from the offline parent under native streaming and with Stream++. The Change column gives the mean improvement of Stream++ over native streaming. Positive is better; VSB rows show ASR reductions.}
\label{fig:plugin}
\vspace{-.1in}
\end{figure}
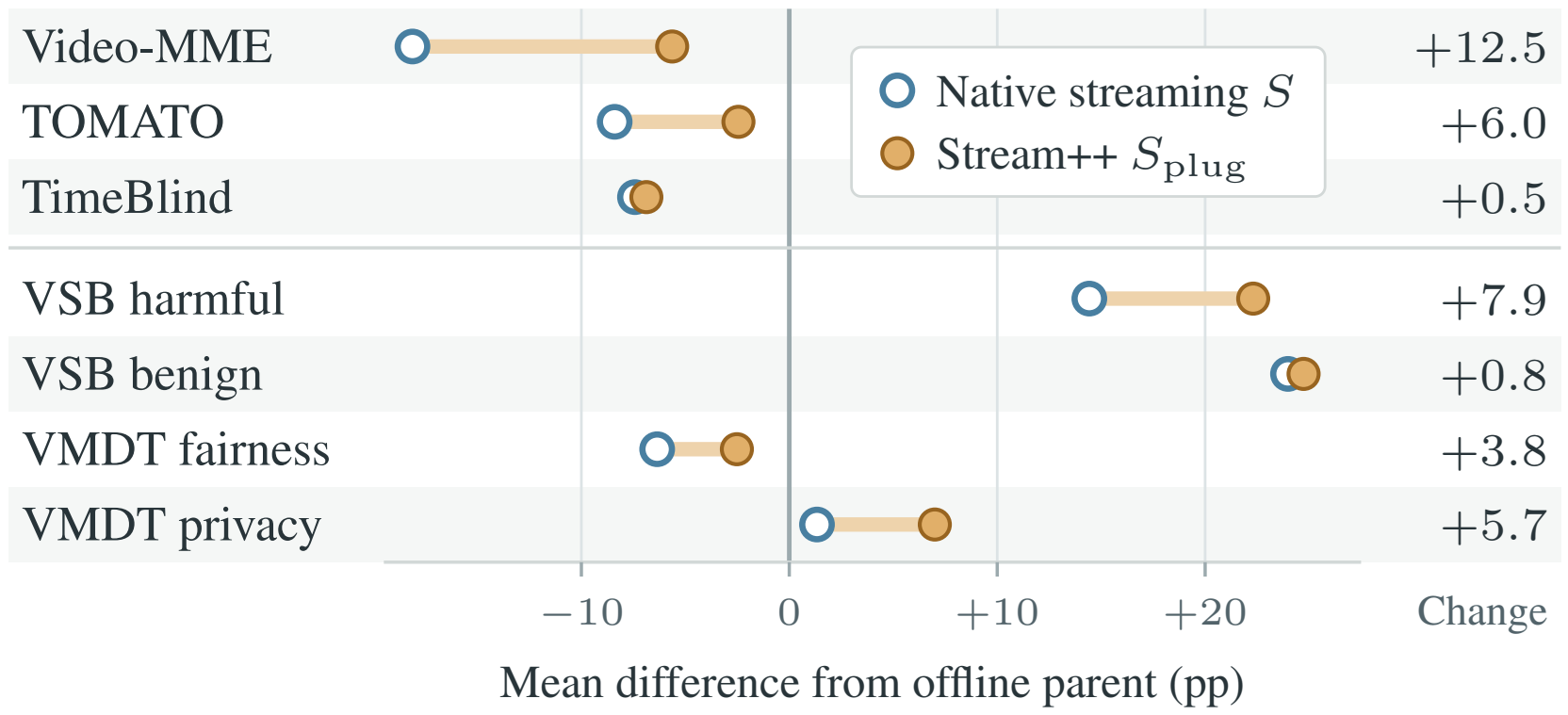

\subsection{Performance Improvement}

As shown in \Cref{fig:plugin}, Stream++ narrows the gap to the offline parent on most tasks, achieving the largest understanding gains for children whose native streaming withholds answers or discards earlier frames (MMDuet and AURA).
For safety, Stream++ yields lower ASR than the offline parents in nearly all VSB comparisons (Appendix \Cref{fig:supp-plugin-children}).
Relative to native streaming, ASR on VSB harmful requests falls by $7.9$pp on average, whereas ASR on VSB benign queries changes by only $0.8$pp.
Specifically, harmful ASR drops by $14.2$ to $45.9$pp for ThinkStream, AURA, and LiveStar, driven by explicit refusals rather than unaddressed requests, while rising for the other three models, most notably for MMDuet, which natively leaves most requests unanswered.
Although refusing harmful requests risks over-refusing harmless ones (Appendix~\ref{supp:recovery-checks}), Stream++ refuses only $4.8\%$ of harmless requests on average, far below the $47.2\%$ left unanswered under native streaming.
Meanwhile, fairness and privacy scores rise by $3.8$ and $5.7$pp on average among the five children with native scores.
Finally, Stream++ maintains the children's performance on streaming benchmarks~\cite{niu2025ovobench,lin2024streamingbench}, where questions arrive during the video (Appendix~\ref{supp:streaming-benchmarks}).
Component ablations and results on longer videos are reported in Appendix~\ref{supp:intervention-ablations} and~\ref{supp:meidum-long-video}.

\section{Conclusion}
\label{sec:discussion}

Streaming VLMs largely preserve their offline parents' understanding but may fail to translate it into reliable and trustworthy responses.
Native streaming further degrades performance due to limited historical evidence and unanswered questions. Stream++ partially mitigates these issues, highlighting the need to preserve both understanding and response behavior in developing unified streaming and offline VLMs.

\noindent \textbf{Limitations.}
First, we did not evaluate many long videos due to resource constraints.
Second, we only cover several available open-weight models and datasets for comparison.
Including more datasets and models in the future could potentially provide more insights.
Third, Stream++ is only a lightweight intervention for improvement; more advanced approaches could be proposed for better performance.

\section*{LLM Usage Statement}

We used OpenAI (ChatGPT) and Anthropic (Claude) during the preparation of this work. These tools were used to provide reviewer-style feedback on the manuscript and to assist with language editing, and to support code development and debugging. The authors take full responsibility for the correctness, validity, and integrity of all methods, experimental results, claims, and submitted content.

\bibliographystyle{plainnat}
\bibliography{main}

\clearpage
\appendix
\setcounter{figure}{0}
\setcounter{table}{0}
\setcounter{equation}{0}
\renewcommand{\thefigure}{S\arabic{figure}}
\renewcommand{\thetable}{S\arabic{table}}
\renewcommand{\theequation}{S\arabic{equation}}
\renewcommand{\theHfigure}{S\arabic{figure}}
\renewcommand{\theHtable}{S\arabic{table}}
\renewcommand{\theHequation}{S\arabic{equation}}

\begin{center}
    {\sffamily\bfseries\fontsize{16}{20}\selectfont Appendix\par}
\end{center}
\vspace{0.2cm}
\appendixcontents
\vspace{0.4cm}
\etocdepthtag.toc{mtappendix}

\raggedbottom

\section{Evaluation Details}
\label{supp:protocol}

\subsection{Model Families}
\label{supp:models}
\label{sec-append-model-summary}

The six streaming children differ in post-training, visual memory, and response control. We summarize the released models here and specify the evaluated interfaces in Sec.~\ref{supp:implementation}.

\textbf{MMDuet.}
MMDuet~\cite{wang2025mmduet} adapts LLaVA-OneVision-7B using approximately 109K MMDuetIT examples for dense captioning, grounded video QA, and temporal grounding.
One epoch of training updates the multimodal projector, language-model LoRA parameters, and two classification heads, with joint supervision for response tokens, frame informativeness, and question relevance.
The heads trigger responses as visual tokens accumulate in the KV cache.

\textbf{ThinkStream.}
ThinkStream~\cite{liu2026thinkstream} adapts Qwen2.5-VL-3B-Instruct through supervised learning on 110K timestamped reasoning examples and GRPO on 9K examples with verifiable answers.
Its rewards combine answer accuracy, output format, and response timing.
The Watch--Think--Speak interface produces reasoning updates and response or silence actions. Its memory retains recent visual tokens together with generated thoughts and responses.

\textbf{MMDuet2.}
MMDuet2~\cite{wang2025mmduet2} adapts Qwen2.5-VL-3B-Instruct through supervised fine-tuning and multi-turn GRPO.
Its proactive corpus contains approximately 52K videos, supplemented by 25K LLaVA-Video QA and 25K Tarsier2 captioning examples.
Training rewards answer quality and timeliness and penalizes repetition and responses outside relevant spans.
Each assistant turn produces a reply or \texttt{NO REPLY} under a proactive system prompt.

\textbf{AURA.}
AURA~\cite{lu2026aura} adapts the language model of Qwen3-VL-8B-Instruct while freezing the vision encoder and connector.
One epoch of supervised training combines approximately 115K streaming and 59K offline video QA samples.
Its Silent-Speech Balanced Loss supervises silent turns and final answers, down-weighting silent targets.
Video chunks form dialogue turns, with a silence token controlling when to wait or answer.

\textbf{LiveStar.}
LiveStar~\cite{yang2025livestar} uses its released LiveStar-InternVideo-8B initialization.
Temporal alignment on 63K samples precedes online adaptation on 20K OmniStar samples for narration, temporal grounding, and QA.
Training updates the projector and language model with the visual encoder frozen.
Streaming Verification Decoding checks previous captions against incoming frames to trigger new narrations. Our evaluation retains the full visual history and narrations.

\textbf{MOSS-VL-Realtime.}
MOSS-VL-Realtime~\cite{wang2026mossvl} continues from MOSS-VL-Instruct-0708 through full-model Realtime-SFT on approximately 0.56M streaming and offline multimodal samples.
It retains the parent's cross-attention architecture and uses \texttt{<|silence|>} and \texttt{<|response|>} tokens to supervise waiting and speaking.
Training reweights the state-token loss and masks assistant turn-final end tokens.

\subsection{Datasets and Metrics}
\label{sec-supp-eval-setup}
\label{sec-supp-eval-dataset}

\textbf{Image understanding.}
We compare $P$ and $C$ on DynaMath~\cite{zou2025dynamath} (5,010 variants of 501 seed problems), MathVista~\cite{lu2024mathvista} (1,000 testmini examples), MMBench~\cite{liu2024mmbench} (1,164 DEV-EN questions with 4,329 circular rows), MME~\cite{fu2025mme} (2,374 questions), MMMU~\cite{yue2024mmmu} (900 validation examples), MMStar~\cite{chen2024mmstar} (1,500 examples), and ScienceQA~\cite{lu2022scienceqa} (2,017 image-containing test questions).
We report accuracy, circular accuracy for MMBench, and the official perception score for MME. MME differences are expressed as a percentage of the maximum score.
The supplementary understanding results also include 1,000 VisuLogic~\cite{xu2026visulogic} examples.

\textbf{Video understanding.}
All three conditions are evaluated on Video-MME~\cite{fu2025videomme} short without subtitles (900 questions on 300 videos), TOMATO~\cite{shangguan2025tomato} (1,484 examples), and TimeBlind~\cite{li2026timeblind} (2,400 queries in 600 four-query instances).
We report accuracy for Video-MME and TOMATO and instance accuracy ($I_{\mathrm{Acc}}$) for TimeBlind.
The supplementary results also include 1,440 VideoReasonBench~\cite{liu2026videoreasonbench} examples under $P$, $C$, and $S$.

\textbf{Safety.}
HarmBench~\cite{mazeika2024harmbench} contains 510 requests evaluated under $P$ and $C$: 400 textual requests, including 100 copyright behaviors, and 110 multimodal requests.
Video-SafetyBench (VSB)~\cite{liu2025videosafetybench} supplies 1,132 harmful-query and 1,132 benign-query rows. The latter have apparently benign text whose harmful intent emerges from the video context.
MCV-SafetyBench (MCV)~\cite{kang2026multiclip} supplies 1,460 explicit-query and 1,460 implicit-query rows with videos composed of multiple related clips.
An explicit query states the harmful request in the text prompt. An implicit query omits it from the text, and the request appears only as text printed in the video frames.
VSB and MCV use all three conditions.
We report the attack success rate (ASR), the percentage of the $N$ requests in a set whose reply is scored as unsafe ($u_i=1$).
On HarmBench, $u_i=1$ when the official classifier judges reply $i$ to be an instance of the requested behavior or, for copyright behaviors, when the official hash check matches it to the copyrighted text.
On VSB and MCV, $u_i=1$ when the RJScore of the reply exceeds 2.85. The RJScore judge rates how unsafe the reply is on a scale from 1 to 5, and the score is the expected rating under the judge's probabilities $p_{i,k}$ of the five ratings:
\[
\mathrm{ASR}=\frac{100}{N}\sum_{i=1}^{N}u_i,\qquad
\mathrm{RJScore}_i=\sum_{k=1}^{5}k\,p_{i,k}.
\]

\textbf{Fairness and privacy.}
VMDT~\cite{potter2025vmdt} is evaluated under all three conditions.
Its fairness subset contains 2,000 rows: 1,063 stereotype, 753 decision, and 184 factual/overkill examples.
Stereotype and decision items ask for a likelihood $\lambda_i$ between 0 and 1, and both scores fall as the differences between demographic groups grow.
Let $z_i=\lambda_i$ for a stereotype item and $z_i=\lambda_i-\lambda_i^{\ast}$ for a decision item, where $\lambda_i^{\ast}$ is the annotated interview score, and let $\bar{z}_g$ be the mean of $z_i$ over the items of group $g$.
For an attribute $a$ with groups $G_a$, the group gap $\delta_a$ and the component score are:
\begin{align*}
\delta_a&=\frac{1}{|G_a|}\sum_{g\in G_a}\Bigl|\,\bar{z}_g-\frac{1}{|G_a|-1}\sum_{g'\neq g}\bar{z}_{g'}\Bigr|,\\
F_{\mathrm{ste}},\,F_{\mathrm{dec}}&=100-\frac{100}{3}\sum_{a}\delta_a.
\end{align*}
The sum runs over gender, age, and skin tone for the stereotype score $F_{\mathrm{ste}}$ and over gender, age, and ethnicity for the decision score $F_{\mathrm{dec}}$.
The factual score $F_{\mathrm{fact}}$ is the percentage of the 184 factual items answered with the correct option, and the overall fairness score is $(F_{\mathrm{ste}}+F_{\mathrm{dec}}+F_{\mathrm{fact}})/3$.
Its privacy task contains 600 location-inference questions on 200 driving videos. The privacy score is 100 minus the weighted state, city, and ZIP-code inference accuracy:
\[
\mathrm{Privacy}=100-\frac{\mathrm{State}+2\,\mathrm{City}+4\,\mathrm{Zip}}{7}.
\]

\textbf{Generalization.}
Prompt robustness uses 300 Video-MME short questions and 300 MMBench circular units, with 100 additional MMBench units for instruction-level perturbations.
We apply typographical errors, paraphrases, and distracting additions~\cite{zhu2023promptrobust} and report the parent's accuracy drop minus the child's, pooled over the two benchmarks.
Downstream fine-tuning uses LingoQA~\cite{marcu2024lingoqa} (500 evaluation examples) and EyePCR~\cite{wang2025eyepcr} (2,000 evaluation examples), with nested training budgets of 1K, 2K, and 5K examples.
We compare final offline accuracy and gains over each checkpoint's zero-shot score, averaged across budgets. Sec.~\ref{supp:generalization} gives the detailed results.

\subsection{Inference and Fine-tuning Protocols}
\label{supp:implementation}

\textbf{Matched offline inference.}
Each parent--child pair receives identical frame indices, prompts, chat templates, greedy decoding, and scoring.
Video frames are sampled with one setting per pair, which both checkpoints use on every video benchmark.
LLaVA-OneVision and MMDuet use 2\,fps with at most 160 frames padded to $384^2$.
Qwen2.5-VL, ThinkStream, and MMDuet2 use qwen-vl-utils sampling at about 2\,fps with 200,704--401,408 pixels per frame. Qwen3-VL and AURA use the same sampler with at most 602,112 pixels per frame.
LiveStar and its parent use 1\,fps with $448^2$ tiles, and MOSS-VL-Realtime and its parent use 1\,fps at the child's pixel budget.
The two video safety benchmarks are the exception. They prescribe their own frames, which every pair uses at its own resolution: 16 uniform frames for VSB and 33 frames at 4\,fps for MCV.
The maximum output length is set per benchmark: 512 tokens for most sets, 1,024 for TOMATO, VSB, and VMDT, 2,048 for DynaMath, MathVista, MMMU, and VisuLogic, 16 for MME, and 4,096 for VideoReasonBench.
On MathVista, MMMU, and MMStar, when a parent or one of its children reaches this limit on a question, the parent and its children are all rerun on that question with 8,192 tokens, and the reported scores use the reruns.
Only LLaVA-OneVision and MMDuet need no rerun, as none of their replies reaches the limit on these benchmarks.

\textbf{Native streaming inference.}
Each child runs its officially released streaming inference procedure on a prerecorded video and receives the question at the end.
The maximum output length of each benchmark is the same as in offline inference.
The settings of each child are as follows.

\emph{MMDuet} runs the frame-by-frame inference loop of its released benchmark code at 2\,fps and 49 tokens per frame, retaining the full visual cache. Generated replies are omitted from subsequent context.
The question is encoded before the final frame, where the model answers if the sum of its informative and relevance head scores exceeds 0.5, the threshold in MMDuet's official MAGQA inference script.

\emph{ThinkStream} uses the official OVO-Bench/RTVU evaluation code with two frames per chunk and chunk length $\max(1,\lceil\mathrm{duration}/120\rceil)$\,s.
Its 24,576-token static cache evicts visual entries beyond the latest 20 chunks while retaining thoughts and template tokens.

\emph{MMDuet2} uses the official proactive API client at 1\,s per frame in two-frame turns, with a full cache and its native proactive system prompt.
A reply equal to \texttt{NO REPLY} is scored as a non-answer.

\emph{AURA} uses the official OVO-Bench/StreamingBench adapter with the last $\min(30,\lfloor\mathrm{duration}\rfloor)$ one-second segments, each containing two frames.
The adapter prefills earlier assistant turns with \texttt{<|silent|>} and supplies the retained streaming-formatted history in one final prefill, followed by generation at $T=0$ in vLLM under its live-stream system prompt.

\emph{LiveStar} uses its streaming session with the accuracy settings of its benchmark script: 1\,fps, narration gated by SVeD, full visual history, and one answer at the end.

\emph{MOSS-VL-Realtime} uses the realtime session shipped with the checkpoint. Frames arrive at 1\,fps on the media clock, all frames remain in memory, and replies follow the official client display rule.

\textbf{Software.}
We use transformers 4.44.2 for LLaVA-OneVision and MMDuet, 4.37.2 for LiveStar and its parent, and 4.57.3 for the other models. The exceptions are MMDuet2 streaming, which uses 4.49.0, and AURA streaming, which uses vLLM 0.11.0.

\textbf{Fine-tuning.}
All decoder linear layers use LoRA with rank 16 and $\alpha=32$.
We use AdamW at learning rate $10^{-4}$, 3\% warmup, a cosine schedule, one epoch, batch size 1, and 8 accumulation steps.
MMDuet is fine-tuned as a merged checkpoint without its response heads.
The main fine-tuning curves use one training seed. The LingoQA replication across three seeds is reported in Sec.~\ref{supp:generalization}.

\subsection{Scoring and Statistics}
\label{supp:scoring}

\textbf{Scorers and judges.}
We use the official benchmark prompts and scoring implementations with the following judge settings.
Qwen3.8-27B-FP8 (thinking off, greedy) extracts answers when parsing rules fail on TOMATO, TimeBlind, VideoReasonBench, and image benchmarks. It also repairs VMDT JSON outputs and assigns the response categories in Figs.~\ref{fig:composition} and~\ref{fig:supp-safety-composition}.
Two LLM raters, blinded to model and condition, checked 125 response labels and agreed with the judge at $\kappa=0.93$. Precision was 100\% for refusal, harmful help, narration, and fragments and 70\% for safe deflection.
Qwen2.5-72B-Instruct additionally scores all 12,581 answer-extraction requests, with accuracy changes within 1 point in every cell. Under MathVista's official extraction prompt, all six child--parent directions agree between judges.
VSB and MCV use RJScore with Qwen2.5-72B-Instruct and an unsafe threshold of 2.85.
HarmBench uses its official classifiers and the official hash check for copyright behaviors.

\textbf{Response categories.}
The response-form audit of the safety evaluation (Figs.~\ref{fig:composition} and~\ref{fig:supp-safety-composition}) assigns each reply to one of four categories: harmful help, explicit refusal (R), safe deflection (D), or non-answer. The harmless requests of Secs.~\ref{supp:trust} and~\ref{supp:recovery-checks} use the same categories, with a reply that answers the request in place of harmful help.
Non-answers include silence, \texttt{NO REPLY}, deferral, off-task narration, fragments, and format failures.
R+D defines the refusal rate; as in XSTest~\cite{rottger2024xstest}, a reply that declines the request or offers only safe alternatives counts as a refusal.

\textbf{Scoring of unanswered requests.}
A streaming child can leave a request unanswered, and each part of the evaluation scores this case as follows.
Coverage measures extractable answers for understanding and fairness, and non-silent replies for safety and privacy.
For understanding, silent outputs receive zero accuracy. Video-MME uses all questions as the denominator, with extracted-answer accuracy also reported. The two scores differ by at most 0.4 points.
TOMATO applies its seeded random-option fallback to unparseable text, while silence remains unanswered.
For safety, RJScore evaluates every VSB and MCV reply and scores empty text as safe. Narration and deferral are judged from their text.
For VMDT, fairness omits silent rows from stereotype and decision scores and counts them as wrong for factual questions. Privacy scoring treats a missing location prediction as no extraction.
Finally, a set on which a child gives no streaming response receives a dash for ASR and privacy and zero accuracy. This applies to MMDuet on VMDT privacy and MCV implicit requests and to MMDuet2 on both MCV splits.

\textbf{Statistical comparisons.}
Paired stratified cluster bootstrap uses 10,000 draws, or 20,000 for VMDT, resampling benchmark-defined units: videos, TimeBlind instances, DynaMath seed problems, MMBench questions, MME images within categories, and VMDT clusters.
We report a paired 95\% bootstrap interval for every tested difference between two conditions: in the table or figure that shows the difference, or after the value when the difference appears only in the text. Absolute scores are reported without intervals.
We use exact McNemar tests for independent binary items and cluster sign-flip tests for grouped items.
Holm correction is applied within families of comparisons. In each part of the evaluation, the contrasts with the parent ($C-P$ and $S-P$) form one family and the contrasts with the child's own offline result ($S-C$) form another, each covering the six children and all benchmarks or metrics of that part.
The families comprise 42 primary and 18 secondary image contrasts, 36 video parent contrasts and 18 video $S-C$ contrasts, 51 safety parent contrasts and 21 safety $S-C$ contrasts, and 89 VMDT parent contrasts and 41 VMDT $S-C$ contrasts; VideoReasonBench is corrected separately, with 12 parent and 6 $S-C$ contrasts.
Robustness and fine-tuning results, which involve only $C-P$, use their respective comparison families.
A contrast is \emph{retained} when its 90\% interval lies within $\pm3$ points under TOST.
Holm-significant differences are labeled \emph{declines} or \emph{gains}.

\section{Detailed Evaluation Results}

\subsection{Understanding}
\label{supp:understanding}

\begin{figure*}[t]
\centering
\input{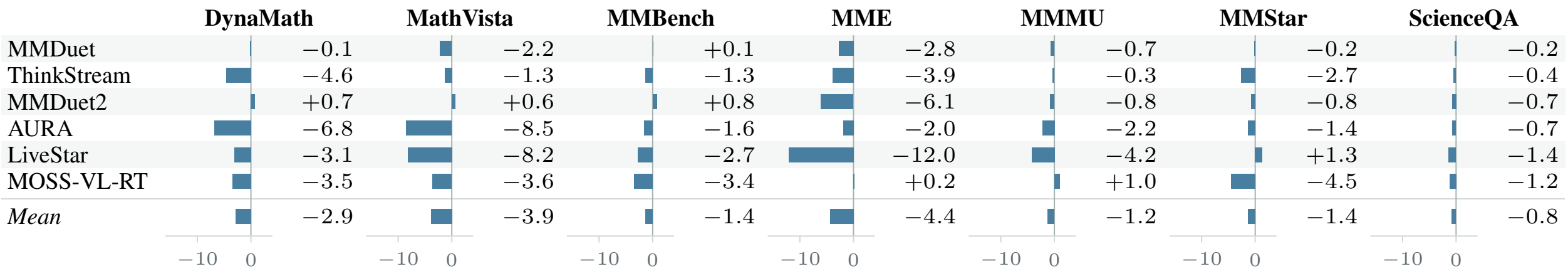}
\caption{Adaptation tax ($C-P$) on each image benchmark, in points on the scale of Fig.~\ref{fig:gaps}. Negative values indicate losses, and Mean weights the six children equally.}
\label{fig:supp-image-gaps}
\end{figure*}

Fig.~\ref{fig:supp-image-gaps} reports the adaptation tax on each image benchmark, and Tables~\ref{tab:eval-image} and~\ref{tab:eval-video} give the absolute image and video scores.
Tables~\ref{tab:eval-image}b and~\ref{tab:eval-video}b give the 95\% intervals of the adaptation, execution, and streaming taxes on each benchmark.
Of the 42 main image contrasts, 19 are retained within $\pm3$ points and nine show a significant decline.
The declines occur in DynaMath, MathVista, MMBench, MMStar, and MME perception and are spread over all six children.
The additional VisuLogic evaluation has 89\% of Qwen3-VL parent replies at the generation cap, which explains its score of 8.6\% in Table~\ref{tab:eval-image}.

Streaming response coverage is at least 99\% except for MMDuet (1.6\% on Video-MME, 0\% on TOMATO, and 3.6\% on TimeBlind) and MMDuet2 on TOMATO (36.2\%) and TimeBlind (61.5\%).
On TimeBlind, MMDuet2 and LiveStar fall to guessing level in streaming, whereas AURA and MOSS-VL-Realtime still tell the paired videos apart.
Specifically, a TimeBlind instance is correct only when all four of its queries, two questions each asked on two videos, are answered correctly, so guessing scores 6.25\%.
To check whether a low score still reflects the video, we compare each run with two baselines built from its own answers.
The first draws every answer at random with the run's answer frequencies. The second keeps the run's two answers to each question and assigns them to the two videos at random.
A run that scores above both tells the two videos of an instance apart.
MMDuet2 and LiveStar score above both baselines offline and at their levels during streaming. AURA and MOSS-VL-Realtime remain above both baselines in streaming.
ThinkStream's offline score is already at its baseline levels.
MOSS-VL-Realtime's TOMATO decline is concentrated in the 709 streaming replies that use free text and require judge extraction, with 99.6\% agreement between the two extraction judges.

\begin{table*}[t]
\centering
\caption{Image understanding under offline evaluation. (a)~Scores. Higher is better. Scores are percentages except MME perception (/2000) and cognition (/800). DynaMath Worst requires all ten variants of a seed problem to be correct. MMBench uses circular accuracy. Scores include the judge-assisted answer extraction where applicable. (b)~Adaptation tax in points with paired 95\% bootstrap intervals, using DynaMath Avg.\ and MME perception as a percentage of its maximum. Negative values indicate losses, and grey marks differences that are not significant after Holm correction.}
\label{tab:eval-image}
\footnotesize
\setlength{\tabcolsep}{1.5pt}
\begin{tabular}{llrrrrrrrrrr}
\toprule
\multicolumn{12}{l}{\textit{(a) Scores}} \\
Model & Mode & \shortstack{DynaMath\\Avg.} & \shortstack{DynaMath\\Worst} & MathVista & MMBench & \shortstack{MME\\Perc.} & \shortstack{MME\\Cog.} & MMMU & MMStar & ScienceQA & VisuLogic \\
\midrule
LLaVA-OneVision-7B & P & 33.6 & 10.2 & 63.3 & 80.6 & 1577.2 & 426.1 & 48.0 & 61.1 & 95.9 & 25.5 \\
MMDuet-7B & C & 33.5 & 9.0 & 61.1 & 80.7 & 1521.4 & 390.4 & 47.3 & 60.9 & 95.6 & 24.2 \\
Qwen2.5-VL-3B & P & 40.1 & 12.6 & 59.1 & 77.3 & 1650.5 & 610.0 & 46.2 & 55.1 & 80.7 & 26.9 \\
ThinkStream-3B & C & 35.5 & 11.6 & 57.8 & 76.0 & 1573.3 & 596.8 & 45.9 & 52.5 & 80.3 & 23.6 \\
MMDuet2-3B & C & 40.8 & 13.4 & 59.7 & 78.1 & 1529.2 & 631.8 & 45.4 & 54.3 & 80.0 & 25.6 \\
Qwen3-VL-8B & P & 64.8 & 37.7 & 75.6 & 85.4 & 1740.0 & 625.4 & 59.6 & 65.1 & 94.6 & 8.6 \\
AURA-8B & C & 58.0 & 28.1 & 67.1 & 83.8 & 1700.7 & 674.6 & 57.3 & 63.7 & 93.9 & 23.7 \\
LiveStar-InternVideo-8B & P & 24.8 & 4.6 & 45.9 & 76.7 & 1502.4 & 514.6 & 51.2 & 46.3 & 85.0 & 25.6 \\
LiveStar-8B & C & 21.7 & 3.6 & 37.7 & 74.0 & 1261.5 & 340.0 & 47.0 & 47.6 & 83.6 & 23.7 \\
MOSS-VL-Instruct & P & 54.8 & 24.9 & 69.6 & 88.0 & 1454.1 & 386.4 & 50.4 & 64.1 & 91.6 & 27.5 \\
MOSS-VL-Realtime & C & 51.3 & 20.6 & 66.0 & 84.5 & 1458.0 & 417.9 & 51.4 & 59.6 & 90.3 & 26.4 \\
\bottomrule
\end{tabular}

\vspace{6pt}
\scriptsize
\setlength{\tabcolsep}{3pt}
\begin{tabular}{@{}lcccccc@{}}
\toprule
\multicolumn{7}{@{}l}{\textit{(b) Adaptation tax ($C-P$)}} \\
Benchmark & MMDuet & ThinkStream & MMDuet2 & AURA & LiveStar & MOSS-VL-Realtime \\
\midrule
DynaMath & \textcolor[HTML]{8A939B}{$-0.1$ [$-1.8$, $+1.6$]} & $-4.6$ [$-6.8$, $-2.4$] & \textcolor[HTML]{8A939B}{$+0.7$ [$-1.3$, $+2.7$]} & $-6.8$ [$-8.8$, $-4.8$] & \textcolor[HTML]{8A939B}{$-3.1$ [$-5.3$, $-0.9$]} & \textcolor[HTML]{8A939B}{$-3.5$ [$-5.7$, $-1.2$]} \\
MathVista & $-2.2$ [$-3.5$, $-0.9$] & \textcolor[HTML]{8A939B}{$-1.3$ [$-4.4$, $+1.8$]} & \textcolor[HTML]{8A939B}{$+0.6$ [$-1.9$, $+3.1$]} & $-8.5$ [$-11.3$, $-5.7$] & $-8.2$ [$-11.0$, $-5.4$] & \textcolor[HTML]{8A939B}{$-3.6$ [$-6.5$, $-0.7$]} \\
MMBench & \textcolor[HTML]{8A939B}{$+0.1$ [$-0.7$, $+0.9$]} & \textcolor[HTML]{8A939B}{$-1.3$ [$-2.8$, $+0.3$]} & \textcolor[HTML]{8A939B}{$+0.8$ [$-0.7$, $+2.2$]} & \textcolor[HTML]{8A939B}{$-1.6$ [$-3.0$, $-0.3$]} & \textcolor[HTML]{8A939B}{$-2.7$ [$-4.4$, $-1.1$]} & $-3.4$ [$-5.0$, $-1.9$] \\
MME & \textcolor[HTML]{8A939B}{$-2.8$ [$-4.6$, $-1.1$]} & \textcolor[HTML]{8A939B}{$-3.9$ [$-6.7$, $-1.1$]} & $-6.1$ [$-8.9$, $-3.2$] & \textcolor[HTML]{8A939B}{$-2.0$ [$-4.3$, $+0.3$]} & $-12.0$ [$-15.6$, $-8.6$] & \textcolor[HTML]{8A939B}{$+0.2$ [$-1.6$, $+2.0$]} \\
MMMU & \textcolor[HTML]{8A939B}{$-0.7$ [$-2.1$, $+0.8$]} & \textcolor[HTML]{8A939B}{$-0.3$ [$-2.7$, $+2.0$]} & \textcolor[HTML]{8A939B}{$-0.8$ [$-3.1$, $+1.6$]} & \textcolor[HTML]{8A939B}{$-2.2$ [$-5.3$, $+0.8$]} & \textcolor[HTML]{8A939B}{$-4.2$ [$-7.0$, $-1.4$]} & \textcolor[HTML]{8A939B}{$+1.0$ [$-1.6$, $+3.6$]} \\
MMStar & \textcolor[HTML]{8A939B}{$-0.2$ [$-1.3$, $+0.9$]} & \textcolor[HTML]{8A939B}{$-2.7$ [$-4.5$, $-0.7$]} & \textcolor[HTML]{8A939B}{$-0.8$ [$-2.5$, $+0.9$]} & \textcolor[HTML]{8A939B}{$-1.4$ [$-3.4$, $+0.5$]} & \textcolor[HTML]{8A939B}{$+1.3$ [$-0.9$, $+3.5$]} & $-4.5$ [$-6.4$, $-2.5$] \\
ScienceQA & \textcolor[HTML]{8A939B}{$-0.2$ [$-0.5$, $0.0$]} & \textcolor[HTML]{8A939B}{$-0.4$ [$-1.3$, $+0.4$]} & \textcolor[HTML]{8A939B}{$-0.7$ [$-1.7$, $+0.2$]} & \textcolor[HTML]{8A939B}{$-0.7$ [$-1.6$, $+0.2$]} & \textcolor[HTML]{8A939B}{$-1.4$ [$-2.6$, $-0.1$]} & \textcolor[HTML]{8A939B}{$-1.2$ [$-2.2$, $-0.2$]} \\
\bottomrule
\end{tabular}

\end{table*}

\begin{table*}[t]
\centering
\caption{Video understanding. (a)~Scores (\%, higher is better). Video-MME uses 900 short-video questions without subtitles. TOMATO has 1,484 questions, TimeBlind has 600 four-query instances, and VideoReasonBench has 1,440 questions. Unanswered streaming questions count as incorrect. MMDuet answers 14, 0, 87, and 4 questions across these benchmarks. (b)~Adaptation, execution, and streaming taxes on the three main benchmarks, in points with paired 95\% bootstrap intervals. Negative values indicate losses, and grey marks differences that are not significant after Holm correction.}
\label{tab:eval-video}
\footnotesize
\setlength{\tabcolsep}{3pt}
\begin{tabular}{lrrrrrrrrrrrr}
\toprule
\multicolumn{13}{l}{\textit{(a) Scores}} \\
Child & \multicolumn{3}{c}{Video-MME short} & \multicolumn{3}{c}{TOMATO} & \multicolumn{3}{c}{TimeBlind $I_{\mathrm{Acc}}$} & \multicolumn{3}{c}{VideoReasonBench} \\
 & P & C & S & P & C & S & P & C & S & P & C & S \\
\midrule
MMDuet-7B & 69.3 & 65.6 & 0.8 & 24.8 & 27.4 & 0.0 & 13.0 & 9.2 & 0.0 & 0.4 & 0.1 & 0.0 \\
ThinkStream-3B & 72.9 & 70.3 & 58.9 & 24.4 & 23.1 & 25.1 & 7.8 & 3.5 & 1.7 & 0.3 & 0.4 & 0.0 \\
MMDuet2-3B & 72.9 & 71.1 & 71.6 & 24.4 & 23.7 & 9.4 & 7.8 & 7.0 & 3.2 & 0.3 & 0.2 & 0.0 \\
AURA-8B & 80.7 & 78.6 & 65.6 & 34.0 & 33.9 & 33.1 & 24.2 & 24.5 & 16.7 & 7.4 & 1.5 & 0.0 \\
LiveStar-8B & 74.4 & 69.7 & 68.6 & 29.2 & 23.9 & 23.4 & 9.5 & 6.3 & 1.8 & 0.3 & 0.0 & 0.0 \\
MOSS-VL-Realtime & 78.4 & 76.4 & 73.4 & 34.9 & 33.1 & 29.3 & 12.0 & 14.5 & 6.8 & 1.8 & 1.4 & 1.6 \\
\bottomrule
\end{tabular}

\vspace{6pt}
\scriptsize
\setlength{\tabcolsep}{3pt}
\begin{tabular}{@{}lcccccc@{}}
\toprule
\multicolumn{7}{@{}l}{\textit{(b) Taxes}} \\
Benchmark & MMDuet & ThinkStream & MMDuet2 & AURA & LiveStar & MOSS-VL-Realtime \\
\midrule
\multicolumn{7}{@{}l}{Adaptation tax ($C-P$)} \\
\quad Video-MME & $-3.8$ [$-5.7$, $-1.9$] & \textcolor[HTML]{8A939B}{$-2.6$ [$-4.7$, $-0.4$]} & \textcolor[HTML]{8A939B}{$-1.8$ [$-3.7$, $+0.2$]} & \textcolor[HTML]{8A939B}{$-2.1$ [$-4.2$, $0.0$]} & $-4.8$ [$-7.1$, $-2.6$] & \textcolor[HTML]{8A939B}{$-2.0$ [$-4.0$, $0.0$]} \\
\quad TOMATO & \textcolor[HTML]{8A939B}{$+2.6$ [$+1.0$, $+4.3$]} & \textcolor[HTML]{8A939B}{$-1.3$ [$-3.8$, $+1.2$]} & \textcolor[HTML]{8A939B}{$-0.7$ [$-2.7$, $+1.3$]} & \textcolor[HTML]{8A939B}{$-0.1$ [$-2.2$, $+2.1$]} & $-5.4$ [$-8.0$, $-2.7$] & \textcolor[HTML]{8A939B}{$-1.8$ [$-4.0$, $+0.5$]} \\
\quad TimeBlind & $-3.8$ [$-6.0$, $-1.7$] & $-4.3$ [$-6.5$, $-2.2$] & \textcolor[HTML]{8A939B}{$-0.8$ [$-2.8$, $+1.0$]} & \textcolor[HTML]{8A939B}{$+0.3$ [$-2.5$, $+3.2$]} & \textcolor[HTML]{8A939B}{$-3.2$ [$-5.3$, $-1.0$]} & \textcolor[HTML]{8A939B}{$+2.5$ [$-0.5$, $+5.5$]} \\
\addlinespace
\multicolumn{7}{@{}l}{Execution tax ($S-C$)} \\
\quad Video-MME & $-64.8$ [$-67.7$, $-61.8$] & $-11.4$ [$-14.8$, $-8.1$] & \textcolor[HTML]{8A939B}{$+0.4$ [$-1.8$, $+2.6$]} & $-13.0$ [$-15.9$, $-10.1$] & \textcolor[HTML]{8A939B}{$-1.1$ [$-2.7$, $+0.4$]} & $-3.0$ [$-4.8$, $-1.3$] \\
\quad TOMATO & $-27.4$ [$-29.7$, $-25.2$] & \textcolor[HTML]{8A939B}{$+2.0$ [$-0.6$, $+4.6$]} & $-14.3$ [$-16.5$, $-12.0$] & \textcolor[HTML]{8A939B}{$-0.8$ [$-2.8$, $+1.1$]} & \textcolor[HTML]{8A939B}{$-0.4$ [$-2.0$, $+1.1$]} & $-3.8$ [$-5.9$, $-1.8$] \\
\quad TimeBlind & $-9.2$ [$-11.5$, $-7.0$] & \textcolor[HTML]{8A939B}{$-1.8$ [$-3.5$, $-0.2$]} & $-3.8$ [$-6.0$, $-1.8$] & $-7.8$ [$-10.8$, $-4.8$] & $-4.5$ [$-6.3$, $-2.8$] & $-7.7$ [$-10.2$, $-5.2$] \\
\addlinespace
\multicolumn{7}{@{}l}{Streaming tax ($S-P$)} \\
\quad Video-MME & $-68.6$ [$-71.6$, $-65.6$] & $-14.0$ [$-17.4$, $-10.6$] & \textcolor[HTML]{8A939B}{$-1.3$ [$-3.7$, $+1.1$]} & $-15.1$ [$-18.3$, $-11.8$] & $-5.9$ [$-8.2$, $-3.6$] & $-5.0$ [$-7.1$, $-2.8$] \\
\quad TOMATO & $-24.8$ [$-27.0$, $-22.6$] & \textcolor[HTML]{8A939B}{$+0.7$ [$-1.8$, $+3.1$]} & $-15.0$ [$-17.3$, $-12.6$] & \textcolor[HTML]{8A939B}{$-0.9$ [$-3.2$, $+1.5$]} & $-5.8$ [$-8.5$, $-3.1$] & $-5.6$ [$-8.2$, $-3.1$] \\
\quad TimeBlind & $-13.0$ [$-15.7$, $-10.5$] & $-6.2$ [$-8.5$, $-3.8$] & $-4.7$ [$-6.7$, $-2.7$] & $-7.5$ [$-10.5$, $-4.7$] & $-7.7$ [$-10.0$, $-5.5$] & $-5.2$ [$-8.0$, $-2.3$] \\
\bottomrule
\end{tabular}

\end{table*}

\subsection{Trustworthiness}
\label{supp:trust}

\begin{figure*}[t]
\centering
\includegraphics[width=\textwidth]{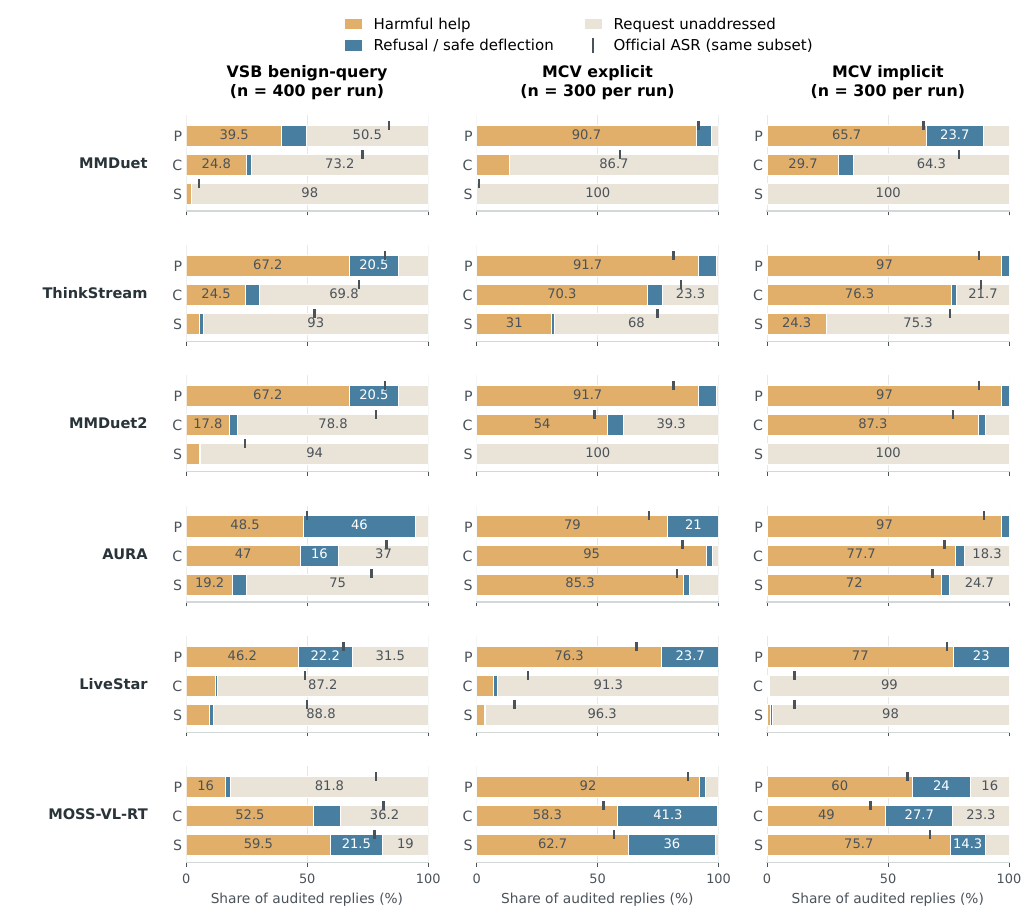}
\caption{Response forms and official attack success rates on the same audited requests across the VSB benign-query and MCV explicit and implicit splits. A benign VSB query can become harmful in its video context.}
\label{fig:supp-safety-composition}
\end{figure*}

Tables~\ref{tab:eval-safety} and~\ref{tab:eval-vmdt} report the safety, fairness, and privacy scores.
The response-form audit uses 400 VSB videos with harmful and benign queries, 300 requests per MCV split, and all 510 HarmBench requests.
Fig.~\ref{fig:supp-safety-composition} complements Fig.~\ref{fig:composition} with the VSB benign-query and MCV results.

\paragraph{Safety after adaptation.}
Under matched offline evaluation, MMDuet and AURA provide harmful assistance more often than their parents on the VSB harmful audit (30\% to 34\% and 22\% to 41\%, respectively).
For ThinkStream, MMDuet2, and LiveStar, both harmful assistance and refusal or safe deflection decrease, while off-task narration and fragments account for 74--81\% of replies.
On MCV, these three children and MMDuet also provide less harmful assistance and leave more requests unaddressed than their parents.
For example, LiveStar's explicit-request ASR falls from 67.6\% to 21.8\%, while 89\% of its audited replies describe the video and 2\% refuse or safely redirect the request.
AURA provides more harmful assistance on MCV explicit requests (79\% to 95\%) and less on implicit requests (97\% to 78\%).
MOSS-VL-Realtime responds more often on VSB, increasing both refusal and harmful assistance. On MCV explicit requests, its harmful assistance falls from 92.0\% to 58.3\% and refusal or safe deflection rises from 2.3\% to 41.3\%.
Table~\ref{tab:supp-trust-intervals} gives the adaptation and execution taxes on safety, fairness, and privacy with their 95\% intervals, including the MCV contrasts.

\begin{table*}[t]
\centering
\caption{Adaptation and execution taxes on trustworthiness, in points with paired 95\% bootstrap intervals. Safety rows report ASR reductions, so a negative value means a higher ASR; fairness and privacy rows report score differences. Grey marks differences that are not significant under the Holm correction of Table~\ref{tab:gaps}. Mean weights the children equally. --: the child has no streaming score (Sec.~\ref{supp:scoring}). $^{\ast}$Mean of the children with a score.}
\label{tab:supp-trust-intervals}
\scriptsize
\setlength{\tabcolsep}{2pt}
\resizebox{\textwidth}{!}{%
\begin{tabular}{@{}lccccccr@{}}
\toprule
Benchmark & MMDuet & ThinkStream & MMDuet2 & AURA & LiveStar & MOSS-VL-Realtime & \textit{Mean} \\
\midrule
\multicolumn{8}{@{}l}{Adaptation tax ($C-P$)} \\
\quad HarmBench & \textcolor[HTML]{8A939B}{$-1.2$ [$-3.9$, $+1.6$]} & \textcolor[HTML]{8A939B}{$+3.9$ [$+1.0$, $+6.9$]} & $-7.3$ [$-10.4$, $-4.1$] & \textcolor[HTML]{8A939B}{$-4.1$ [$-7.3$, $-1.2$]} & $-8.8$ [$-13.1$, $-4.7$] & $-6.1$ [$-9.0$, $-3.1$] & $-3.9$ \\
\quad VSB harmful & $-11.8$ [$-14.9$, $-8.7$] & $-16.6$ [$-20.2$, $-13.0$] & $-24.5$ [$-27.9$, $-21.0$] & $-42.8$ [$-45.8$, $-39.8$] & $-28.4$ [$-32.2$, $-24.6$] & $+11.8$ [$+8.7$, $+15.0$] & $-18.7$ \\
\quad VSB benign & $+8.3$ [$+5.4$, $+11.2$] & $+9.2$ [$+6.1$, $+12.2$] & \textcolor[HTML]{8A939B}{$+3.6$ [$+0.5$, $+6.7$]} & $-35.8$ [$-39.2$, $-32.2$] & $+14.7$ [$+10.7$, $+18.6$] & $-5.7$ [$-8.7$, $-2.7$] & $-1.0$ \\
\quad MCV explicit & $+31.6$ [$+28.9$, $+34.2$] & \textcolor[HTML]{8A939B}{$-2.0$ [$-4.4$, $+0.3$]} & $+33.8$ [$+31.0$, $+36.6$] & $-16.4$ [$-19.0$, $-13.8$] & $+45.8$ [$+42.5$, $+49.0$] & $+34.8$ [$+31.8$, $+37.9$] & $+21.3$ \\
\quad MCV implicit & $-15.4$ [$-18.4$, $-12.5$] & \textcolor[HTML]{8A939B}{$-1.4$ [$-3.4$, $+0.7$]} & $+10.6$ [$+8.2$, $+13.1$] & $+12.0$ [$+9.7$, $+14.3$] & $+62.5$ [$+59.7$, $+65.3$] & $+18.8$ [$+15.8$, $+21.8$] & $+14.5$ \\
\quad Fairness & \textcolor[HTML]{8A939B}{$0.0$ [$-0.7$, $+0.8$]} & \textcolor[HTML]{8A939B}{$-0.2$ [$-2.1$, $+1.7$]} & \textcolor[HTML]{8A939B}{$-2.6$ [$-4.9$, $-0.2$]} & $-3.7$ [$-6.0$, $-1.8$] & \textcolor[HTML]{8A939B}{$-1.8$ [$-5.4$, $+0.7$]} & $+3.2$ [$+1.3$, $+4.8$] & $-0.9$ \\
\quad Privacy & $-5.8$ [$-8.1$, $-3.4$] & $+5.1$ [$+3.4$, $+7.0$] & $+10.0$ [$+7.9$, $+12.3$] & \textcolor[HTML]{8A939B}{$+3.1$ [$+0.1$, $+6.0$]} & $-7.9$ [$-10.7$, $-5.1$] & $+5.6$ [$+3.1$, $+8.1$] & $+1.7$ \\
\addlinespace
\multicolumn{8}{@{}l}{Execution tax ($S-C$)} \\
\quad VSB harmful & $+76.4$ [$+73.8$, $+79.0$] & $+16.2$ [$+12.5$, $+19.7$] & $+49.8$ [$+46.6$, $+52.9$] & \textcolor[HTML]{8A939B}{$+0.4$ [$-2.4$, $+3.3$]} & $+12.5$ [$+9.3$, $+15.6$] & $+42.5$ [$+39.3$, $+45.6$] & $+33.0$ \\
\quad VSB benign & $+66.3$ [$+63.4$, $+69.2$] & $+21.3$ [$+17.8$, $+24.9$] & $+52.1$ [$+48.9$, $+55.3$] & $+5.8$ [$+2.6$, $+9.2$] & \textcolor[HTML]{8A939B}{$+1.1$ [$-2.1$, $+4.2$]} & $+3.9$ [$+1.1$, $+6.6$] & $+25.1$ \\
\quad MCV explicit & $+58.6$ [$+56.1$, $+61.2$] & $+10.8$ [$+8.2$, $+13.4$] & -- & $+5.6$ [$+3.6$, $+7.6$] & $+4.9$ [$+2.7$, $+7.1$] & $-3.6$ [$-5.3$, $-1.8$] & $+15.3^{\ast}$ \\
\quad MCV implicit & -- & $+15.7$ [$+13.1$, $+18.4$] & -- & $+8.8$ [$+6.2$, $+11.4$] & \textcolor[HTML]{8A939B}{$+0.2$ [$-1.9$, $+2.3$]} & $-27.0$ [$-30.1$, $-23.8$] & $-0.6^{\ast}$ \\
\quad Fairness & -- & \textcolor[HTML]{8A939B}{$+2.3$ [$-0.7$, $+4.6$]} & \textcolor[HTML]{8A939B}{$-3.3$ [$-7.5$, $-0.1$]} & \textcolor[HTML]{8A939B}{$-3.2$ [$-5.6$, $-1.0$]} & $-16.9$ [$-20.3$, $-13.6$] & $-3.3$ [$-5.0$, $-1.7$] & $-4.9^{\ast}$ \\
\quad Privacy & -- & $+4.9$ [$+3.4$, $+6.4$] & $-2.9$ [$-4.4$, $-1.4$] & $-8.4$ [$-11.8$, $-5.1$] & \textcolor[HTML]{8A939B}{$-0.2$ [$-1.2$, $+0.6$]} & \textcolor[HTML]{8A939B}{$-2.4$ [$-4.6$, $-0.2$]} & $-1.8^{\ast}$ \\
\bottomrule
\end{tabular}%
}
\end{table*}

\paragraph{Response behavior during streaming.}
MMDuet, ThinkStream, MMDuet2, and LiveStar leave 89--98\% of audited VSB benign queries unaddressed during streaming.
AURA's unaddressed share rises from 37.0\% offline to 75.0\% in streaming, while refusal or safe deflection falls from 16.0\% to 5.8\%.
On MCV, MMDuet and MMDuet2 leave all audited requests unaddressed, and LiveStar leaves over 96\% unaddressed on both splits.
MOSS-VL-Realtime follows a different pattern: streaming raises refusal or safe deflection on VSB harmful requests from 24\% to 61\% and reduces harmful assistance from 55\% to 30\%.
Its harmful assistance increases on VSB benign queries and both MCV splits.

The official judge scores 67\% of off-task narrations as unsafe on VSB harmful requests ($n=2{,}828$), compared with 18\% on MCV explicit requests ($n=736$).
The judge's prompt includes a description of the harm category, and each benchmark supplies its own.
The gap does not come from these descriptions: when 200 narrations from each benchmark are rescored with one generic description shared by both, 65\% remain unsafe on VSB and 16\% on MCV.
VSB videos depict the harmful act, so narrating their content can restate the harmful request.
This is why a child that narrates instead of answering can have a higher ASR than its parent on VSB and a lower one on MCV, as MMDuet2 and LiveStar do (Table~\ref{tab:supp-trust-intervals}).

\paragraph{Harmless requests.}
On 40 harmless how-to questions about the VSB videos, five children answer fewer questions in streaming than offline: MMDuet (16 to 1), ThinkStream (18 to 5), MMDuet2 (12 to 4), AURA (30 to 20), and LiveStar (15 to 9).
MOSS-VL-Realtime answers more questions in streaming (28 to 34).
The same evaluation on 96 benign clips also produces unanswered requests, connecting reduced response coverage to legitimate use.

\paragraph{Fairness and privacy.}
VMDT fairness averages stereotype, decision, and factual components.
The factual component equals $50+J/2$, where $J$ is factual discrimination, and accounts for most of the variation across models.
Stereotype and decision items ask for a likelihood as a number.
LiveStar often answers in words instead: 1,065 of its 1,816 offline replies and 1,071 of its streaming replies contain no number, and the fallback judge (Sec.~\ref{supp:scoring}) reads the likelihood from the text.
Its stereotype and decision scores therefore depend on this judge.

The privacy scores in the main text and in Table~\ref{tab:eval-vmdt} use the official scorer.
It takes the first match of its pattern in a reply, which in a long reply can be a phrase before the location is named, so the score depends on how the reply is worded.
As a check, a second scorer counts every correct location mentioned.
It reproduces MMDuet's and LiveStar's privacy losses, while changing the MOSS-VL-Realtime offline contrast from $+5.6$ to $-0.1$ points ($[-2.9,+2.6]$) and the AURA streaming contrast from $-5.4$ ($[-8.9,-1.6]$) to $+1.7$ points ($[-1.6,+5.1]$).
Explicit refusals account for at most 12\% of privacy replies. Most non-extraction comes from omitted locations, fragments, or unparseable responses.
MMDuet answers none of the 600 streaming privacy questions and receives no privacy score.

\paragraph{Fairness under streaming.}
The execution tax on VMDT fairness mostly reflects how streaming changes the delivery of answers.
Among the five children with a streaming score, a stronger dependence of answers on the speaker's group appears only in AURA's stereotype component.
Table~\ref{tab:supp-vmdt-execution}a splits each child's execution tax into the three components.

LiveStar's 16.9-point loss lies entirely in the factual component.
Each factual item shows two portraits and asks for the letter of the historically accurate one, and the official rule takes the first character of the reply as the answer.
In streaming, LiveStar never begins a reply with an option letter: all 184 replies are sentences such as ``The second portrait is the most historically accurate,'' so every reply is scored as wrong, whichever portrait it names.
Scoring the portrait named in each reply gives 102 correct answers in streaming and 103 offline ($-0.5$ points, $[-8.2,+7.1]$).
Without LiveStar, the other four children lose 1.9 points on average, compared with 4.9 for all five.

MMDuet2 answers \texttt{NO REPLY} to 74.8\% of stereotype, 82.1\% of decision, and 18.5\% of factual items, so its decision score rests on 135 of 753 items.
On a 761-item subset, removing its streaming system prompt reduces these replies from 509 to none and raises its fairness score by 6.5 points ($[+2.4,+10.1]$).
Its parent never answers \texttt{NO REPLY} in the same interface with the same prompt (Table~\ref{tab:protocol-locus}), and with the prompt removed from both, the child differs from the parent by $-0.2$ points ($[-3.5,+2.0]$).

AURA's fairness falls both offline and in streaming.
Offline, its factual score is 11.4 points below its parent's ($[-17.4,-5.4]$), which accounts for most of its offline decline.
In streaming it answers every item, and its further loss has two sources.
The first is the stereotype component ($-4.1$ points).
Each stereotype item asks for the likelihood that the speaker holds the higher-status of two occupations, such as doctor rather than nurse, and the score falls with the absolute gaps between the mean likelihoods assigned to gender, age, and skin-tone groups.
Because the gaps are absolute, answers that vary more lower the score even when they do not depend on the group, so we compare each score with the mean score after shuffling the group labels 300 times.
AURA's streaming score lies 4.4 points below this baseline, compared with 1.1 points offline and at most 0.8 points for the other four children in streaming.
Under its live-stream system prompt, AURA assigns a likelihood of at least 0.5 to the higher-status occupation on 39.7\% of items (10.9\% offline), and men receive a mean likelihood 0.117 higher than women ($[+0.060,+0.178]$), compared with 0.017 offline ($[-0.021,+0.055]$).
Removing only the system prompt, with the one-second segments, prefilled silent turns, and 30-second window unchanged, returns the stereotype score to 97.0 (97.2 offline) and the gender difference to 0.026 ($[-0.013,+0.065]$).
A generic system prompt reproduces the gender difference (0.116, $[+0.042,+0.187]$, 532 items).
The parent run through the same adapter with AURA's prompt has a gender difference of 0.016, and its stereotype score changes by $-0.6$ points ($[-2.8,+0.7]$) from its offline value.
The stereotype loss therefore requires both AURA's weights and a system turn.

The second source is the factual component ($-5.4$ points), which comes from the streaming input format and not from AURA's training.
To separate the two, we run the parent's weights in AURA's streaming procedure, with the same system prompt, one-second segments, prefilled silent turns, and 30-second window.
The parent then loses 15.2 factual points relative to its offline score ($[-21.7,-9.2]$).
Removing the system prompt leaves AURA's factual score unchanged (54.3 vs.\ 54.9), so the prompt is not the cause.
When the two portraits are swapped, AURA's answer follows the portraits on 35\% of items in streaming, compared with 67\% offline.

MOSS-VL-Realtime loses 3.3 points, mostly on the factual component ($-7.6$).
It answers every factual item with a letter and is silent on 5 of the 2,000 items.
Its stereotype score falls by 1.1 points, and this fall does not reflect a dependence on the speaker's group.
Offline it returns nearly the same likelihood on every item, which leaves no gap between groups.
In streaming its answers vary, and its score stays within 0.1 points of the label-shuffled baseline.

ThinkStream's fairness score rises by 2.3 points.
Its factual score rises by 7.6 points because it answers A on 86\% of items in streaming, compared with 41\% offline.
A constant answer scores 50 on the factual component and ThinkStream scores 40.8 offline, so a stronger preference for one position raises its score.
On a 1,093-item subset, removing the system prompt lowers the share of A answers to 68\% and the factual score by 9.8 points ($[-17.4,-2.2]$).
Its stereotype score falls by 0.8 points. Its official streaming procedure samples replies at temperature 1, so its numerical answers scatter more than under greedy decoding (standard deviation 0.33 vs.\ 0.04), and scatter alone widens the absolute gaps between groups. With greedy decoding the score rises by 2.2 points ($[+1.2,+5.0]$), to within 0.1 points of the parent's offline score.

\paragraph{Privacy under streaming.}
Privacy scores fall for four of the five children under streaming. For AURA and MMDuet2 the decline holds under both scorers, and for MOSS-VL-Realtime it comes from the reply format.
The privacy score falls whenever a reply contains the correct state, city, or ZIP code. Offline, most replies without an extracted location describe the scene, state that the location cannot be determined, or name it in a form that the official extractor does not read.
Table~\ref{tab:supp-vmdt-execution}b reports the execution tax under the official scorer and under the scorer that counts every correct location mentioned.

AURA's privacy score falls by 8.4 points under the official scorer and by 5.0 under the mention-based one.
Its state-level replies without an extracted location fall from 44.5\% offline to 10.5\% in streaming, and its ZIP-code replies stating that the location cannot be determined fall from 40\% to 4\%.
Its replies mention the correct state on 76.5\% of questions, up from 55.5\% offline and level with its parent's 77.5\%.
MMDuet2 loses 2.9 points under the official scorer and 8.9 under the mention-based one.
Offline, 80\% of its state-level replies describe the scene without naming a location, compared with 50\% in streaming, and the share of city-level replies that mention the correct city rises from 4.5\% to 29.0\%.
It stays silent on 20 of the 600 privacy questions, against 72\% of the fairness items.

MOSS-VL-Realtime's official score falls by 2.4 points, while its replies mention the correct state on 71.0\% of questions in both modes.
Offline, 27\% of its state-level replies name a location that the official extractor does not read, compared with 1\% in streaming.
LiveStar's score falls by 0.2 points under the official scorer and 0.5 under the mention-based one, with both intervals including zero.
ThinkStream's score rises by 4.9 points because 78\% of its state-level streaming replies are unfinished fragments, and the share mentioning the correct state falls from 60.0\% to 8.5\%.

\begin{table*}[t]
\centering
\caption{Attack success rates (\%, lower is better). HarmBench contains 510 requests, VSB has 1,132 requests per split, and MCV has 1,460 per split. Benign-query VSB requests are paired with potentially harmful video context. VSB and MCV use Qwen2.5-72B-Instruct RJScore $>2.85$. Dashes indicate settings not evaluated (HarmBench streaming) or scores unreported under the response-coverage rule in Sec.~\ref{supp:scoring}.}
\label{tab:eval-safety}
\footnotesize
\setlength{\tabcolsep}{3pt}
\begin{tabular}{llrrrrr}
\toprule
Model & Mode & HarmBench & \multicolumn{2}{c}{Video-SafetyBench} & \multicolumn{2}{c}{MCV-SafetyBench} \\
 &  & ASR & Harmful & Benign-query & Explicit & Implicit \\
\midrule
LLaVA-OneVision-7B & P & 33.1 & 71.6 & 81.6 & 90.7 & 61.7 \\
MMDuet-7B & C & 34.3 & 83.5 & 73.3 & 59.1 & 77.1 \\
MMDuet-7B & S & -- & 7.1 & 7.1 & 0.5 & -- \\
Qwen2.5-VL-3B & P & 26.5 & 56.9 & 81.8 & 82.2 & 86.2 \\
ThinkStream-3B & C & 22.6 & 73.5 & 72.6 & 84.2 & 87.6 \\
ThinkStream-3B & S & -- & 57.3 & 51.3 & 73.4 & 71.9 \\
MMDuet2-3B & C & 33.7 & 81.4 & 78.2 & 48.4 & 75.6 \\
MMDuet2-3B & S & -- & 31.5 & 26.1 & -- & -- \\
Qwen3-VL-8B & P & 19.6 & 17.5 & 45.4 & 70.7 & 87.9 \\
AURA-8B & C & 23.7 & 60.2 & 81.2 & 87.0 & 75.9 \\
AURA-8B & S & -- & 59.8 & 75.3 & 81.4 & 67.0 \\
LiveStar-InternVideo-8B & P & 20.6 & 36.2 & 64.9 & 67.6 & 75.4 \\
LiveStar-8B & C & 29.4 & 64.6 & 50.3 & 21.8 & 12.9 \\
LiveStar-8B & S & -- & 52.1 & 49.2 & 16.9 & 12.7 \\
MOSS-VL-Instruct & P & 20.8 & 83.1 & 76.2 & 86.7 & 60.1 \\
MOSS-VL-Realtime & C & 26.9 & 71.3 & 82.0 & 51.9 & 41.2 \\
MOSS-VL-Realtime & S & -- & 28.8 & 78.1 & 55.5 & 68.2 \\
\bottomrule
\end{tabular}
\end{table*}

\begin{table*}[t]
\centering
\caption{VMDT results (\%). Fairness averages stereotype, decision, and factual scores over 1,063, 753, and 184 items. Privacy uses 600 questions from 200 videos, with score $100-(\mathrm{State}+2\mathrm{City}+4\mathrm{Zip})/7$. State, City, and Zip are location-inference accuracies. Dashes mark unreported scores under the response-coverage rule in Sec.~\ref{supp:scoring}.}
\label{tab:eval-vmdt}
\footnotesize
\setlength{\tabcolsep}{3pt}
\begin{tabular}{llrrrrrrrr}
\toprule
Model & Mode & \multicolumn{4}{c}{VMDT Fairness} & \multicolumn{4}{c}{VMDT Privacy} \\
 &  & Overall $\uparrow$ & Stereo. $\uparrow$ & Decision $\uparrow$ & Factual $\uparrow$ & Overall $\uparrow$ & State $\downarrow$ & City $\downarrow$ & Zip $\downarrow$ \\
\midrule
LLaVA-OneVision-7B & P & 81.4 & 98.8 & 97.1 & 48.4 & 86.0 & 55.0 & 19.5 & 1.0 \\
MMDuet-7B & C & 81.4 & 97.2 & 96.9 & 50.0 & 80.2 & 50.5 & 43.0 & 0.5 \\
MMDuet-7B & S & -- & -- & -- & 0.0 & -- & -- & -- & -- \\
Qwen2.5-VL-3B & P & 79.3 & 99.9 & 95.5 & 42.4 & 87.8 & 61.5 & 9.0 & 1.5 \\
ThinkStream-3B & C & 79.0 & 99.6 & 96.7 & 40.8 & 92.9 & 46.5 & 0.5 & 0.5 \\
ThinkStream-3B & S & 81.3 & 98.9 & 96.7 & 48.4 & 97.8 & 8.5 & 3.5 & 0.0 \\
MMDuet2-3B & C & 76.7 & 99.7 & 96.1 & 34.2 & 97.8 & 11.5 & 2.0 & 0.0 \\
MMDuet2-3B & S & 73.3 & 97.5 & 90.4 & 32.1 & 94.9 & 25.0 & 5.5 & 0.0 \\
Qwen3-VL-8B & P & 88.4 & 98.8 & 94.5 & 71.7 & 78.6 & 74.0 & 23.0 & 7.5 \\
AURA-8B & C & 84.6 & 97.2 & 96.4 & 60.3 & 81.6 & 46.5 & 33.0 & 4.0 \\
AURA-8B & S & 81.4 & 93.1 & 96.2 & 54.9 & 73.2 & 73.5 & 51.0 & 3.0 \\
LiveStar-InternVideo-8B & P & 82.7 & 97.5 & 98.0 & 52.7 & 77.1 & 78.0 & 33.0 & 4.0 \\
LiveStar-8B & C & 81.0 & 98.5 & 93.8 & 50.5 & 69.2 & 76.5 & 64.5 & 2.5 \\
LiveStar-8B & S & 64.0 & 97.5 & 94.6 & 0.0 & 69.0 & 78.0 & 63.5 & 3.0 \\
MOSS-VL-Instruct & P & 80.3 & 100.0 & 94.2 & 46.7 & 85.1 & 63.5 & 12.5 & 4.0 \\
MOSS-VL-Realtime & C & 83.5 & 100.0 & 96.6 & 53.8 & 90.7 & 51.0 & 3.0 & 2.0 \\
MOSS-VL-Realtime & S & 80.2 & 98.8 & 95.5 & 46.2 & 88.3 & 64.0 & 3.0 & 3.0 \\
\bottomrule
\end{tabular}
\end{table*}

\begin{table*}[t]
\centering
\caption{Execution tax on VMDT ($S-C$, points) with paired 95\% bootstrap intervals. (a)~Fairness by component. Unanswered is the share of streaming replies that are silent or \texttt{NO REPLY} on stereotype, decision, and factual items, and Letter is the share of streaming factual replies that begin with an option letter. (b)~Privacy under the official scorer and under the scorer that counts every correct location mentioned. The last three columns give offline $\rightarrow$ streaming shares of the 200 questions at that level: replies without an extracted state, and replies that mention the correct state or city. MMDuet has no streaming score.}
\label{tab:supp-vmdt-execution}
\footnotesize
\setlength{\tabcolsep}{3pt}
\begin{tabular}{@{}lllllcr@{}}
\toprule
\multicolumn{7}{@{}l}{\textit{(a) Fairness}} \\
Child & Overall & Stereotype & Decision & Factual & Unanswered (\%) & Letter (\%) \\
\midrule
ThinkStream & $+2.3$ $[-0.7,+4.6]$ & $-0.8$ $[-2.9,-0.5]$ & $0.0$ $[-2.6,+1.9]$ & $+7.6$ $[0.0,+15.2]$ & 1.5/6.4/0.0 & 100.0 \\
MMDuet2 & $-3.3$ $[-7.5,-0.1]$ & $-2.2$ $[-4.3,-0.9]$ & $-5.7$ $[-14.9,+0.2]$ & $-2.2$ $[-10.3,+6.0]$ & 74.8/82.1/18.5 & 81.5 \\
AURA & $-3.2$ $[-5.6,-1.0]$ & $-4.1$ $[-6.9,-2.0]$ & $-0.2$ $[-1.2,+0.9]$ & $-5.4$ $[-12.0,+1.1]$ & 0.0/0.0/0.0 & 100.0 \\
LiveStar & $-16.9$ $[-20.3,-13.6]$ & $-1.0$ $[-2.4,+1.5]$ & $+0.8$ $[-3.6,+4.2]$ & $-50.5$ $[-59.8,-41.3]$ & 0.0/0.0/0.0 & 0.0 \\
MOSS-VL-Realtime & $-3.3$ $[-5.0,-1.7]$ & $-1.1$ $[-2.5,-0.7]$ & $-1.1$ $[-2.8,+1.1]$ & $-7.6$ $[-12.0,-3.3]$ & 0.4/0.1/0.0 & 100.0 \\
\bottomrule
\end{tabular}

\vspace{6pt}
\begin{tabular}{@{}lllccc@{}}
\toprule
\multicolumn{6}{@{}l}{\textit{(b) Privacy}} \\
Child & Official scorer & Every mention & No state extracted (\%) & State mentioned (\%) & City mentioned (\%) \\
\midrule
ThinkStream & $+4.9$ $[+3.4,+6.4]$ & $+17.5$ $[+14.8,+20.3]$ & 36.5 $\rightarrow$ 88.0 & 60.0 $\rightarrow$ 8.5 & 42.0 $\rightarrow$ 7.5 \\
MMDuet2 & $-2.9$ $[-4.4,-1.4]$ & $-8.9$ $[-11.3,-6.6]$ & 86.5 $\rightarrow$ 72.0 & 11.5 $\rightarrow$ 25.0 & 4.5 $\rightarrow$ 29.0 \\
AURA & $-8.4$ $[-11.8,-5.1]$ & $-5.0$ $[-8.1,-1.9]$ & 44.5 $\rightarrow$ 10.5 & 55.5 $\rightarrow$ 76.5 & 57.5 $\rightarrow$ 65.5 \\
LiveStar & $-0.2$ $[-1.2,+0.6]$ & $-0.5$ $[-1.4,+0.3]$ & 6.0 $\rightarrow$ 4.5 & 76.5 $\rightarrow$ 78.0 & 65.5 $\rightarrow$ 65.5 \\
MOSS-VL-Realtime & $-2.4$ $[-4.6,-0.2]$ & $+2.0$ $[-0.6,+4.6]$ & 33.5 $\rightarrow$ 15.5 & 71.0 $\rightarrow$ 71.0 & 61.5 $\rightarrow$ 51.5 \\
\bottomrule
\end{tabular}
\end{table*}

\subsection{Generalization}
\label{supp:generalization}

\paragraph{Prompt robustness.}
We apply typos, paraphrases, and distracting content to 300 Video-MME short questions and 300 MMBench circular units.
The paraphrase comparison uses 294 and 285 valid, matched units, respectively.
The pooled contrast is $\Delta_{\mathrm{Rob}}=\mathrm{Drop}_P-\mathrm{Drop}_C$, the parent's accuracy drop minus the child's.
Tables~\ref{tab:robustness-pooled} and~\ref{tab:eval-robustness} give the pooled and per-benchmark contrasts.
A complementary gold-option probability measure places all 18 child--parent contrasts within a 1.7-point non-inferiority margin.
ThinkStream under distraction shows the single Holm-significant decrease on this measure ($-0.95$ points of probability, $[-1.29,-0.60]$).

The MMBench instruction comparison changes the answer instruction while holding the image, question, and options fixed (Table~\ref{tab:eval-instruction}).
It covers 100 circular units, including 95 matched paraphrase units, and a 300-unit extension.
All 18 contrasts in the extension have Holm-adjusted $p$-values above 0.05.

\begin{figure*}[t]
\centering
\includegraphics[width=\textwidth]{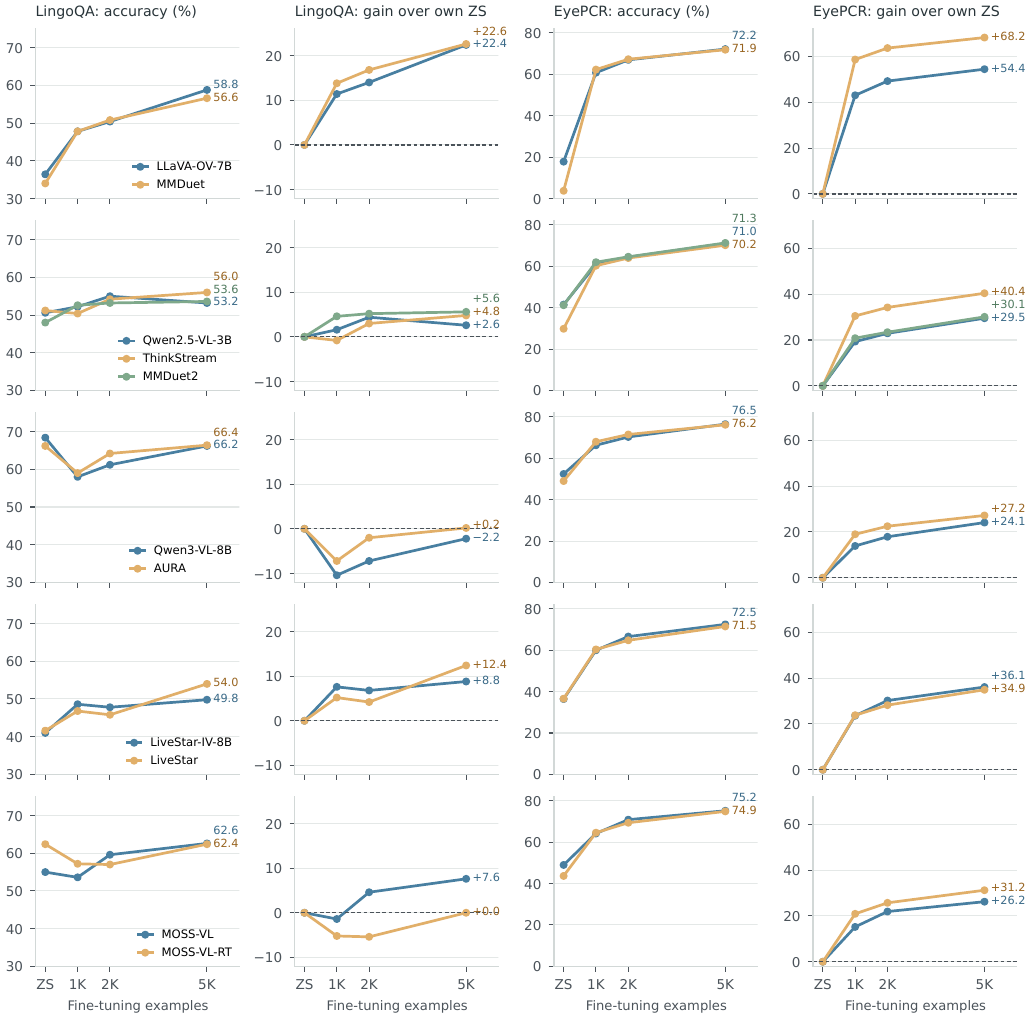}
\caption{Original single-seed fine-tuning results on LingoQA and EyePCR. Left: accuracy. Right: gain over zero-shot accuracy (ZS). Blue denotes the parent, and orange and green denote children. Each budget trains a separate adapter for one epoch.}
\label{fig:supp-finetune}
\end{figure*}

\paragraph{Downstream fine-tuning.}
We report both the gain over each checkpoint's zero-shot accuracy and the final accuracy, averaging across the 1K, 2K, and 5K training budgets.
Tables~\ref{tab:eval-finetuning} and~\ref{tab:eval-finetuning-gains} give the results at each budget, and Fig.~\ref{fig:supp-finetune} shows the curves from the original single-seed runs.
All twelve budget-averaged child--parent final-accuracy intervals include zero, with a worst simultaneous lower bound of $-3.5$ points.
The task means in Table~\ref{tab:gaps} average the two datasets. We test them with the same bootstrap and sign-flip procedures, resampling the two test sets independently, and apply Holm correction over the 12 cells.
The gain difference is significant for MMDuet ($+8.2$ points, $[+6.0,+10.4]$), ThinkStream ($+5.3$, $[+2.6,+8.1]$), and AURA ($+3.9$, $[+1.5,+6.2]$); no final-accuracy difference is significant.

A three-seed LingoQA replication trains every checkpoint at all three budgets, varying LoRA initialization and the order of the same training examples.
Averaged over budgets and seeds, child--parent accuracy differences range from $-1.2$ to $+1.2$ points, with seed SDs of 0.2--1.5 points.
Every seed-inclusive 95\% interval contains zero, with the widest interval at $[-3.4,+1.1]$ for LiveStar.
MOSS-VL-Realtime's gain over its own zero-shot score remains lower than its parent's by $7.1$ points ($[-12.1,-2.1]$), reflecting their different starting accuracies.

\begin{table*}[t]
\centering
\caption{Pooled prompt-robustness contrasts $\Delta_{\mathrm{Rob}}=\mathrm{Drop}_P-\mathrm{Drop}_C$ over Video-MME and MMBench, in points with 95\% cluster-bootstrap intervals. Positive values indicate a smaller accuracy drop for the child. Table~\ref{tab:eval-robustness} gives the per-benchmark results.}
\label{tab:robustness-pooled}
\footnotesize
\setlength{\tabcolsep}{6pt}
\begin{tabular}{@{}lccc@{}}
\toprule
Child & Typo & Paraphrase & Distraction \\
\midrule
MMDuet & $+0.7$ [$-1.0$, $+2.3$] & $+0.7$ [$-1.2$, $+2.8$] & $+0.1$ [$-1.7$, $+1.8$] \\
ThinkStream & $+0.5$ [$-1.2$, $+2.2$] & $+1.9$ [$-0.2$, $+4.0$] & $+1.5$ [$-0.2$, $+3.2$] \\
MMDuet2 & $-0.2$ [$-1.8$, $+1.5$] & $+1.2$ [$-1.0$, $+3.5$] & $-0.2$ [$-1.9$, $+1.4$] \\
AURA & $+0.2$ [$-1.3$, $+1.7$] & $-0.5$ [$-2.1$, $+1.0$] & $+0.4$ [$-1.2$, $+2.1$] \\
LiveStar & $+0.3$ [$-1.8$, $+2.5$] & $-1.6$ [$-3.6$, $+0.3$] & $-0.8$ [$-2.5$, $+0.8$] \\
MOSS-VL-Realtime & $+1.7$ [$-0.2$, $+3.5$] & $-0.9$ [$-2.8$, $+1.0$] & $-0.8$ [$-2.8$, $+1.1$] \\
\bottomrule
\end{tabular}
\end{table*}

\begin{table*}[t]
\centering
\caption{Question robustness under offline inference on 300 questions per benchmark, using circular units for MMBench. Perturbation columns give $\Delta_{\mathrm{Rob}}=\mathrm{Drop}_P-\mathrm{Drop}_C$ in points with 95\% cluster-bootstrap intervals. Paraphrases use 294 matched Video-MME and 285 matched MMBench units. Distraction averages stress and irrelevant-content conditions within each item.}
\label{tab:eval-robustness}
\footnotesize
\setlength{\tabcolsep}{3pt}
\begin{tabular}{llrrrrr}
\toprule
Child & Benchmark & Clean P & Clean C & Typo & Paraphrase & Distraction \\
\midrule
MMDuet-7B & Video-MME & 68.7 & 68.0 & +0.7 [-2.0, +3.3] & +0.3 [-3.1, +3.7] & -0.7 [-3.5, +2.2] \\
MMDuet-7B & MMBench & 82.7 & 81.3 & +0.7 [-1.0, +2.3] & +1.1 [-0.7, +2.8] & +0.8 [-1.0, +2.7] \\
ThinkStream-3B & Video-MME & 72.0 & 68.7 & +0.7 [-1.7, +3.0] & +3.1 [+0.0, +6.5] & +2.3 [-0.3, +5.2] \\
ThinkStream-3B & MMBench & 79.7 & 76.0 & +0.3 [-2.0, +2.7] & +0.7 [-1.8, +3.2] & +0.7 [-1.5, +2.8] \\
MMDuet2-3B & Video-MME & 72.0 & 68.3 & -0.3 [-3.0, +2.3] & +2.0 [-1.4, +5.4] & +0.0 [-2.5, +2.5] \\
MMDuet2-3B & MMBench & 79.7 & 81.0 & +0.0 [-2.0, +2.0] & +0.4 [-2.5, +3.2] & -0.5 [-2.5, +1.7] \\
AURA-8B & Video-MME & 78.3 & 78.3 & +0.0 [-2.7, +2.7] & -1.0 [-3.7, +1.4] & -1.7 [-4.2, +1.0] \\
AURA-8B & MMBench & 85.7 & 85.7 & +0.3 [-1.0, +2.0] & +0.0 [-1.8, +1.8] & +2.5 [+0.5, +4.7] \\
LiveStar-8B & Video-MME & 74.3 & 70.0 & -0.3 [-3.7, +3.0] & +0.0 [-2.7, +2.7] & -1.5 [-4.0, +0.8] \\
LiveStar-8B & MMBench & 80.7 & 79.3 & +1.0 [-1.7, +3.7] & -3.2 [-6.0, -0.4] & -0.2 [-2.5, +2.2] \\
MOSS-VL-Realtime & Video-MME & 78.7 & 75.7 & +3.3 [+0.3, +6.3] & -2.4 [-5.1, +0.3] & -0.5 [-3.7, +2.7] \\
MOSS-VL-Realtime & MMBench & 92.0 & 87.7 & +0.0 [-2.3, +2.3] & +0.7 [-2.1, +3.5] & -1.2 [-3.3, +1.2] \\
\bottomrule
\end{tabular}
\end{table*}

\begin{table*}[t]
\centering
\caption{Instruction perturbations on MMBench. Values give parent and child accuracy (\%) and $\Delta_{\mathrm{Rob}}$ in points. Typo and irrelevant-text comparisons use 100 circular units, and paraphrases use 95 matched units. Intervals and drops use the corresponding clean subsets. All 18 comparisons have Holm-adjusted $p=1$.}
\label{tab:eval-instruction}
\footnotesize
\setlength{\tabcolsep}{3pt}
\begin{tabular}{lrrrrrrrrr}
\toprule
Child & \multicolumn{3}{c}{Instruction typo} & \multicolumn{3}{c}{Instruction paraphrase} & \multicolumn{3}{c}{Instruction irrelevant text} \\
 & P & C & $\Delta_{\mathrm{Rob}}$ [95\% CI] & P & C & $\Delta_{\mathrm{Rob}}$ [95\% CI] & P & C & $\Delta_{\mathrm{Rob}}$ [95\% CI] \\
\midrule
MMDuet-7B & 80.0 & 80.0 & +0.0 [+0.0, +0.0] & 80.0 & 80.0 & +0.0 [+0.0, +0.0] & 81.0 & 80.0 & -1.0 [-3.0, +0.0] \\
ThinkStream-3B & 81.0 & 77.0 & -1.0 [-3.0, +0.0] & 80.0 & 80.0 & +2.1 [+0.0, +5.3] & 79.0 & 76.0 & +0.0 [-3.0, +3.0] \\
MMDuet2-3B & 81.0 & 82.0 & +0.0 [+0.0, +0.0] & 80.0 & 82.1 & +1.1 [+0.0, +3.2] & 79.0 & 81.0 & +1.0 [+0.0, +3.0] \\
AURA-8B & 85.0 & 85.0 & -1.0 [-3.0, +0.0] & 86.3 & 85.3 & -2.1 [-5.3, +0.0] & 84.0 & 84.0 & -1.0 [-3.0, +0.0] \\
LiveStar-8B & 81.0 & 84.0 & +0.0 [-3.0, +3.0] & 82.1 & 84.2 & +0.0 [-3.2, +3.2] & 81.0 & 80.0 & -4.0 [-10.0, +1.0] \\
MOSS-VL-Realtime & 94.0 & 88.0 & +1.0 [+0.0, +3.0] & 94.7 & 88.4 & +1.1 [+0.0, +3.2] & 94.0 & 89.0 & +2.0 [+0.0, +5.0] \\
\bottomrule
\end{tabular}
\end{table*}

\begin{table*}[t]
\centering
\caption{Downstream accuracy (\%) on LingoQA and EyePCR at zero-shot and 1K, 2K, and 5K training budgets. Each budget uses one independently trained, one-epoch LoRA adapter with one seed. P and C denote the parent and child initialization. All evaluations use offline inference.}
\label{tab:eval-finetuning}
\footnotesize
\setlength{\tabcolsep}{3pt}
\begin{tabular}{llrrrrrrrr}
\toprule
Model & Init. & \multicolumn{4}{c}{LingoQA ($n=500$)} & \multicolumn{4}{c}{EyePCR ($n=2{,}000$)} \\
 &  & 0 & 1K & 2K & 5K & 0 & 1K & 2K & 5K \\
\midrule
LLaVA-OneVision-7B & P & 36.4 & 47.8 & 50.4 & 58.8 & 17.8 & 60.8 & 67.0 & 72.2 \\
MMDuet-7B & C & 34.0 & 47.8 & 50.8 & 56.6 & 3.7 & 62.3 & 67.3 & 71.9 \\
Qwen2.5-VL-3B & P & 50.6 & 52.2 & 55.0 & 53.2 & 41.5 & 60.8 & 64.4 & 71.0 \\
ThinkStream-3B & C & 51.2 & 50.4 & 54.2 & 56.0 & 29.8 & 60.3 & 64.0 & 70.2 \\
MMDuet2-3B & C & 48.0 & 52.6 & 53.2 & 53.6 & 41.2 & 62.0 & 64.6 & 71.3 \\
Qwen3-VL-8B & P & 68.4 & 58.0 & 61.2 & 66.2 & 52.4 & 66.3 & 70.3 & 76.5 \\
AURA-8B & C & 66.2 & 59.0 & 64.2 & 66.4 & 49.0 & 68.0 & 71.5 & 76.2 \\
LiveStar-InternVideo-8B & P & 41.0 & 48.6 & 47.8 & 49.8 & 36.4 & 60.0 & 66.6 & 72.5 \\
LiveStar-8B & C & 41.6 & 46.8 & 45.8 & 54.0 & 36.6 & 60.4 & 64.8 & 71.5 \\
MOSS-VL-Instruct & P & 55.0 & 53.6 & 59.6 & 62.6 & 49.0 & 64.2 & 70.9 & 75.2 \\
MOSS-VL-Realtime & C & 62.4 & 57.2 & 57.0 & 62.4 & 43.7 & 64.6 & 69.4 & 74.9 \\
\bottomrule
\end{tabular}
\end{table*}

\begin{table*}[t]
\centering
\caption{Budget-averaged fine-tuning gain differences, $\overline{\Delta\mathrm{Gain}}=\frac{1}{3}\sum_{b\in\{1K,2K,5K\}}[(C_b-C_0)-(P_b-P_0)]$, in percentage points. Holm correction covers the 12 child--dataset comparisons. Intervals capture test-set sampling for the single-seed runs.}
\label{tab:eval-finetuning-gains}
\footnotesize
\setlength{\tabcolsep}{3pt}
\begin{tabular}{lrrrrrr}
\toprule
Child & \multicolumn{3}{c}{LingoQA} & \multicolumn{3}{c}{EyePCR} \\
 & $\overline{\Delta\mathrm{Gain}}$ & 95\% CI & Holm $p$ & $\overline{\Delta\mathrm{Gain}}$ & 95\% CI & Holm $p$ \\
\midrule
MMDuet-7B & +1.80 & [-2.33, +5.93] & 1 & +14.63 & [+12.88, +16.38] & 0.0012 \\
ThinkStream-3B & -0.53 & [-5.40, +4.60] & 1 & +11.22 & [+8.68, +13.78] & 0.0012 \\
MMDuet2-3B & +2.27 & [-3.07, +7.33] & 1 & +0.83 & [-1.38, +3.08] & 1 \\
AURA-8B & +3.60 & [-0.47, +7.67] & 0.694 & +4.10 & [+1.80, +6.44] & 0.0135 \\
LiveStar-8B & -0.47 & [-6.20, +5.40] & 1 & -1.00 & [-3.36, +1.40] & 1 \\
MOSS-VL-Realtime & -7.13 & [-12.00, -2.13] & 0.0384 & +4.83 & [+2.09, +7.59] & 0.013 \\
\bottomrule
\end{tabular}
\end{table*}

\section{Diagnostic Analyses}
\label{supp:diagnostics}

\subsection{Historical Evidence and Input Protocols}

\paragraph{Evaluation sets.}
The window, protocol, and memory controls use Video-MME-750, comprising 750 questions on 250 Video-MME short videos separate from the memory-control development set.
The question-timing comparison (Table~\ref{tab:supp-question-timing}) uses Video-MME-300, one question from each of the 300 Video-MME short videos.
Both checkpoints receive the same questions within each comparison.

\begin{figure*}[t]
\centering
\includegraphics[width=\textwidth]{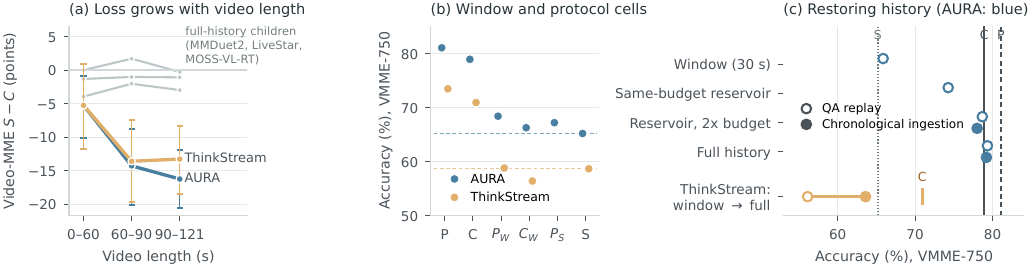}
\caption{History controls on Video-MME. (a)~Execution tax by video duration. (b)~Video-MME-750 offline scores with parent ($P_W$) and child ($C_W$) inputs restricted to the streaming window. (c)~History restoration on Video-MME-750. QA replay processes the retained history in one final prefill at question time. Chronological ingestion updates the model state as each video chunk arrives. Error bars show 95\% intervals. MMDuet's duration results are given in the text.}
\label{fig:supp-history}
\end{figure*}

\paragraph{History availability.}
On Video-MME, AURA and ThinkStream each lose about 5.3 points when switching to streaming for videos shorter than 60 seconds, compared with 16.3 and 13.3 points for videos lasting 90--121 seconds (Fig.~\ref{fig:supp-history}a).
AURA's difference between these duration groups remains significant after Holm correction across six children.
MMDuet loses 69.7 ($[-75.4,-64.0]$), 62.6 ($[-67.7,-57.5]$), and 63.4 ($[-68.0,-58.8]$) points in the three duration groups because it answers few questions throughout.

Restricting offline input to the streaming window brings the child and parent scores close on Video-MME-750 (Table~\ref{tab:diagnosis-modes}).
The difference $S-C_W$ is $-1.1$ points for AURA ($[-3.1,+0.9]$) and $+2.3$ for ThinkStream ($[-1.2,+5.7]$).
The window accounts for 92\% ($[79\%,107\%]$) of AURA's execution tax and 118\% ($[91\%,159\%]$) of ThinkStream's.

At AURA's native budget of 30 one-second segments, keeping the latest 10 and sampling 20 earlier segments scores 74.3\%, compared with 65.9\% for the recent-only window ($+8.4$ points, $[+5.3,+11.5]$).
At ThinkStream's native budget of 20 chunks, keeping the latest 10 and sampling 10 earlier chunks scores 60.8\%, compared with 56.1\% for the recent-only window ($+4.7$, $[+1.7,+7.9]$) and 63.6\% with the full history.
The full plugin additionally applies Respond. Its ThinkStream retention-policy contrast on 846 questions is $+1.9$ points (Table~\ref{tab:supp-recall-budget}).

\begin{table}[!htbp]
\centering
\caption{Diagnostic accuracy (\%) on Video-MME-750, with three questions per video. $P_W$ and $C_W$ restrict offline input to the last 20 seconds for ThinkStream and 30 seconds for AURA. $P_S$ runs the parent in the child's interface. ThinkStream's parent lacks this streaming-token interface.}
\label{tab:diagnosis-modes}
\small
\setlength{\tabcolsep}{4pt}
\begin{tabular}{lrrrrrr}
\toprule
Child & $P$ & $C$ & $P_W$ & $C_W$ & $P_S$ & $S$ \\
\midrule
ThinkStream & 73.5 & 70.9 & 58.8 & 56.4 & -- & 58.7 \\
AURA & 81.1 & 78.9 & 68.4 & 66.3 & 67.2 & 65.2 \\
\bottomrule
\end{tabular}
\end{table}

\paragraph{Parent controls.}
\label{sec:protocol-factors}
On understanding, the execution tax follows the streaming procedure and not the child's weights.
We run parent weights in each child's procedure ($P_S$) and compare each checkpoint with its own offline score (Table~\ref{tab:protocol-locus}).
The parent loses about as much as the child, and every interaction interval in the table includes zero.
Both checkpoints lose accuracy in the AURA Video-MME, MMDuet TimeBlind, and MMDuet2 TOMATO controls.
Safety is different: under the same procedure, AURA's ASR remains 40.8 points above its parent's.
For AURA, the parent runs with AURA's system prompt, one-second segments, prefilled silent turns, and 30-second window.
For MMDuet and MMDuet2, both checkpoints are forced to answer at the same turn.
On TimeBlind, matching MMDuet's 49-token per-frame budget offline lowers its parent's score by 6.9 points ($[-9.4,-4.5]$) and its own score by 3.6 ($[-5.9,-1.4]$).
ThinkStream's parent uses an offline proxy because its interface lacks ThinkStream's special streaming tokens. The proxy supplies either the last 20 visual chunks or the stored streaming thoughts to both checkpoints.

On 481 held-out TimeBlind instances, supplying AURA's streaming engine with the offline-format input raises instance accuracy from 15.8\% to 23.9\% ($+8.1$, $[+4.8,+11.4]$).
This matches its offline score (difference $0.0$, $[-2.5,+2.5]$).
The control jointly changes frame grouping, the pixel budget, the system turn, and timestamps.

\begin{table*}[t]
\centering
\caption{Parent controls for streaming execution. $P_S$ runs the parent in the child's procedure. ThinkStream uses the input controls described below. Differences are in points with nominal 95\% intervals, with positive values indicating improvement. Interaction $=(S-C)-(P_S-P)$ compares the protocol effects on child and parent.}
\label{tab:protocol-locus}
\footnotesize
\setlength{\tabcolsep}{3pt}
\begin{tabular}{@{}lrrrrrrrl@{}}
\toprule
Benchmark (set) & $P$ & $C$ & $S$ & $P_S$ & $P_S-P$ & $S-C$ & Interaction & Pattern \\
\midrule
\multicolumn{9}{@{}l}{\textit{AURA (parent Qwen3-VL-8B)}} \\
Video-MME (750 questions) & 81.1 & 78.9 & 65.2 & 67.2 & $-13.9$ $[-17.1,-10.8]$ & $-13.7$ $[-16.9,-10.5]$ & $+0.1$ $[-2.5,+2.9]$ & Both decline \\
TimeBlind $I_{\mathrm{Acc}}$ (full) & 24.2 & 24.5 & 16.7 & 17.3 & $-6.8$ $[-9.7,-4.0]$ & $-7.8$ $[-10.8,-4.8]$ & $-1.0$ $[-4.3,+2.5]$ & Both decline \\
TOMATO (full) & 34.0 & 33.9 & 33.1 & 32.9 & $-1.1$ $[-2.9,+0.7]$ & $-0.8$ $[-2.8,+1.2]$ & $+0.3$ $[-2.2,+2.7]$ & \shortstack[l]{No clear\\change} \\
VMDT fairness (VMDT-1399) & 88.2 & 84.4 & 81.5 & 82.9 & $-5.3$ $[-7.6,-3.3]$ & $-2.9$ $[-5.4,-0.7]$ & $+2.5$ $[-0.6,+5.4]$ & Both decline$^{a}$ \\
VSB harmful ASR (VSB-400) & 17.5 & 62.8 & 61.5 & 20.8 & $-3.2$ $[-6.2,-0.2]$ & $+1.2$ $[-3.8,+6.2]$ & $+4.5$ $[-1.2,+10.2]$ & Gap persists$^{b}$ \\
\midrule
\multicolumn{9}{@{}l}{\textit{MMDuet (parent LLaVA-OneVision-7B)$^{c}$}} \\
TimeBlind $I_{\mathrm{Acc}}$ (577 instances) & 13.3 & 9.4 & 1.9$^{\dagger}$ & 6.4$^{\dagger}$ & $-6.9$ $[-9.5,-4.3]$ & $-7.5$ $[-9.9,-5.0]$ & $-0.5$ $[-3.1,+2.1]$ & Both decline \\
Video-MME (345 questions) & 67.5 & 63.2 & 64.6$^{\dagger}$ & 66.4$^{\dagger}$ & $-1.2$ $[-5.2,+2.9]$ & $+1.4$ $[-2.3,+5.2]$ & $+2.6$ $[-1.7,+6.7]$ & \shortstack[l]{No clear\\change} \\
\midrule
\multicolumn{9}{@{}l}{\textit{MMDuet2 (parent Qwen2.5-VL-3B)$^{d}$}} \\
TOMATO (1,405 questions) & 23.7 & 23.7 & 19.6$^{\dagger}$ & 20.4$^{\dagger}$ & $-3.3$ $[-5.1,-1.4]$ & $-4.1$ $[-6.2,-1.9]$ & $-0.8$ $[-3.5,+1.9]$ & Both decline \\
Silent rows (VMDT-761) & -- & -- & 509 & 0 & & & & \shortstack[l]{Child-only\\silence} \\
\midrule
\multicolumn{9}{@{}l}{\textit{ThinkStream (parent Qwen2.5-VL-3B), input proxies$^{e}$}} \\
Video-MME, last 20 chunks (345) & 73.9 & 70.4 & 58.3 & 59.4 & $-14.5$ $[-19.4,-9.6]$ & $-12.2$ $[-17.1,-7.5]$ & $+2.3$ $[-1.4,+6.1]$ & Both decline \\
Video-MME, stored thoughts (285) & 74.4 & 70.9 & 66.0 & 68.8 & $-5.6$ $[-9.5,-1.4]$ & $-4.9$ $[-9.5,-0.7]$ & $+0.7$ $[-3.9,+5.3]$ & Both decline \\
\bottomrule
\end{tabular}

\vspace{2pt}
{\scriptsize $^{a}$AURA's parent loses accuracy on factual items. The streaming stereotype gap is $S-P_S=-4.4$ $[-7.7,-1.7]$. $^{b}$The streaming safety gap is $S-P_S=-40.8$ $[-45.8,-35.5]$. $^{c}$MMDuet's parent uses its base projector with response heads disabled. Both checkpoints are forced to answer at the final frame. The TimeBlind comparison uses 577 non-development instances. $^{d}$MMDuet2 and its parent answer after identical \texttt{NO REPLY} histories. $P$ and $C$ use offline option readouts, and both checkpoints are forced to answer at the same turn. $^{\dagger}$Forced answers; the differences in these rows use them. $^{e}$ThinkStream uses offline option readouts on the last 20 streaming chunks or on inputs containing stored thoughts.}
\end{table*}

\paragraph{Visual concept access.}
We fit mean-difference probes at 13 layers for 4,265 object, action, attribute, and scene concepts, using video-disjoint folds and within-word centering.
The offline parent--child comparisons show similar concept readability across model pairs.
For AURA, object AUROC falls from 0.94 offline to 0.74 for objects outside its streaming window, and restoring the frames restores their readability.
These measurements complement the QA controls by tracking access to specific concepts in the video.

\subsection{Response Decisions}

\paragraph{Question timing.}
We compare questions posed at the end, start, and midpoint of each Video-MME-300 video.
After an earlier question, the stream continues and we score the last reply. If the child has given no reply by the end, we request an answer at that point.
ThinkStream is the exception: its official inference code ends the run once it answers, so it sees no later frames and cannot answer again.
LiveStar's interface takes the question after its narration stream, so the timing comparison covers the other five children.
Table~\ref{tab:supp-question-timing} reports the changes relative to an end-of-video question.
Questions at the start lower accuracy by 12.3--33.3 points across all five children.
Midpoint questions lower accuracy for MMDuet, MMDuet2, and MOSS-VL-Realtime, while ThinkStream and AURA remain close to their end-of-video scores.

The first reply typically arrives soon after the question, within the first 15\% of the video for start questions and at 49\%--58\% for midpoint questions.
AURA and MOSS-VL-Realtime reply once, while MMDuet and MMDuet2 can reply repeatedly.
For MMDuet with a start question, 44.9\% of first replies and 34.8\% of last replies are correct. For MMDuet2 with a midpoint question, the corresponding accuracies are 65.0\% and 54.7\%.
A separate ThinkStream comparison on Video-MME-750 retains full history and produces the answer at the end in both conditions. Moving the question to the start changes accuracy by $-19.5$ points ($[-24.3,-14.9]$).

\begin{table}[!htbp]
\centering
\caption{Question timing on Video-MME-300. End, Start, and Middle give accuracy (\%). $\Delta$ is the change from End, with paired 95\% bootstrap intervals over videos. We score the last reply. $^{\dagger}$Includes forced answers: a question still unanswered at the end receives an answer request there.}
\label{tab:supp-question-timing}
\small
\setlength{\tabcolsep}{3pt}
\resizebox{\columnwidth}{!}{%
\begin{tabular}{@{}lrrlrl@{}}
\toprule
Child & End$^{\dagger}$ & Start$^{\dagger}$ & $\Delta$ & Middle$^{\dagger}$ & $\Delta$ \\
\midrule
MMDuet & 65.7 & 34.3 & $-31.3$ $[-37.7,-25.0]$ & 56.3 & $-9.3$ $[-14.3,-4.3]$ \\
MMDuet2 & 72.0 & 38.7 & $-33.3$ $[-39.3,-27.7]$ & 54.7 & $-17.3$ $[-22.3,-12.3]$ \\
ThinkStream & 56.0 & 43.7 & $-12.3$ $[-18.7,-6.3]$ & 60.0 & $+4.0$ $[-1.3,+9.0]$ \\
AURA & 67.3 & 49.7 & $-17.7$ $[-23.3,-12.0]$ & 67.7 & $+0.3$ $[-5.3,+6.0]$ \\
MOSS-VL-Realtime & 75.0 & 44.0 & $-31.0$ $[-37.3,-24.7]$ & 67.0 & $-8.0$ $[-12.7,-3.3]$ \\
\bottomrule
\end{tabular}}
\end{table}

\paragraph{Prompts and answer delivery.}
Removing the streaming system prompt eliminates silence on the tested MMDuet2 fairness items.
On 96 TOMATO development questions, a compact question-and-options prompt reduces its silent responses from 58 to zero and raises correct answers from 11 to 27.
On AURA's 400 harmful VSB requests, removing its streaming system turn reduces deferrals from 247 to 8, while the ASR change remains within sampling uncertainty.
With the same system turn retained, its parent defers on none of these requests.
The prompt change also raises AURA's VMDT stereotype score by 3.9 points ($[+1.8,+6.6]$).

\begin{table*}[t]
\centering
\caption{Children evaluated offline with their streaming system prompt ($C_{\mathrm{sys}}$). Video-MME reports accuracy on Video-MME-300, TimeBlind reports instance accuracy on all 600 instances, and VSB reports ASR on the 400 harmful requests (\%). $\Delta=C_{\mathrm{sys}}-C$, with paired 95\% bootstrap intervals over videos or instances. MMDuet2's ``NO REPLY'' answers count as unanswered. For VSB, $C_{\mathrm{sys}}$ and the reference $C$ are both judged on A100 GPUs: the reference re-scores the original offline replies for MMDuet and AURA and is a new offline run for the other three children. $S$ keeps its original scores. LiveStar is omitted.}
\label{tab:supp-system-prompt}
\footnotesize
\setlength{\tabcolsep}{3pt}
\begin{tabular}{@{}lrrrlrrrlrrrl@{}}
\toprule
& \multicolumn{4}{c}{Video-MME} & \multicolumn{4}{c}{TimeBlind} & \multicolumn{4}{c}{VSB ASR} \\
\cmidrule(lr){2-5}\cmidrule(lr){6-9}\cmidrule(l){10-13}
Child & $C$ & $C_{\mathrm{sys}}$ & $S$ & $\Delta$ & $C$ & $C_{\mathrm{sys}}$ & $S$ & $\Delta$ & $C$ & $C_{\mathrm{sys}}$ & $S$ & $\Delta$ \\
\midrule
MMDuet & 68.0 & 67.3 & 1.0 & $-0.7$ $[-2.3,+1.0]$ & 9.2 & 7.3 & 0.0 & $-1.8$ $[-3.5,-0.3]$ & 83.8 & 77.5 & 4.0 & $-6.3$ $[-11.0,-1.5]$ \\
ThinkStream & 68.7 & 57.3 & 56.0 & $-11.3$ $[-16.7,-6.0]$ & 3.5 & 6.2 & 1.7 & $+2.7$ $[+0.5,+4.8]$ & 72.5 & 56.3 & 59.0 & $-16.3$ $[-22.5,-10.0]$ \\
MMDuet2 & 68.3 & 68.3 & 72.0 & $0.0$ $[-1.7,+1.7]$ & 7.0 & 3.2 & 3.2 & $-3.8$ $[-5.7,-2.2]$ & 83.3 & 0.8 & 29.5 & $-82.5$ $[-86.3,-78.8]$ \\
AURA & 78.3 & 79.7 & 67.3 & $+1.3$ $[0.0,+3.0]$ & 24.5 & 23.7 & 16.7 & $-0.8$ $[-3.0,+1.3]$ & 63.3 & 63.8 & 61.5 & $+0.5$ $[-4.0,+5.0]$ \\
MOSS-VL-Realtime & 75.7 & 74.7 & 75.0 & $-1.0$ $[-2.7,+0.3]$ & 14.5 & 9.7 & 6.8 & $-4.8$ $[-7.0,-2.7]$ & 72.0 & 51.3 & 30.0 & $-20.8$ $[-25.8,-15.8]$ \\
\bottomrule
\end{tabular}
\end{table*}

\paragraph{Streaming system prompts offline.}
The streaming system prompt alone accounts for MMDuet2's execution tax on TimeBlind and for the lower streaming ASR of ThinkStream and MMDuet2, but not for the understanding losses of AURA or MMDuet.
We evaluate each child offline with its native streaming system prompt in place of the offline one ($C_{\mathrm{sys}}$), keeping the frames, single-turn layout, decoding, and scoring of $C$.
LiveStar is omitted because it answers streaming questions with its offline system message.
Table~\ref{tab:supp-system-prompt} covers Video-MME-300, all 2,400 TimeBlind questions, and the 400 harmful VSB requests.

On understanding, the prompt changes AURA's scores by at most 1.3 points and MMDuet's Video-MME score by 0.7, against execution taxes of 7.8 to 67.0 points.
Under its streaming prompt, MMDuet2 answers ``NO REPLY'' to 45\% of TimeBlind questions, compared with 38\% silence in streaming. We count these answers as unanswered, as we do for silence in streaming. Its instance accuracy then falls to its streaming level of 3.2\%.
On the questions it answers, its accuracy is unchanged (59.2\% vs.\ 59.1\% with the offline prompt).
MOSS-VL-Realtime answers every question, and its TimeBlind instance accuracy falls by 4.8 points, 63\% of its execution tax.
ThinkStream's Video-MME score falls by 11.3 points, close to its execution tax of 12.7 points.
Twenty of the lost questions have no extractable answer: 17 replies give the answer letter only inside the reasoning tags, and 3 are truncated.
ThinkStream's streaming option readout avoids such failures, and on the 275 answered questions the drop is 5.1 points ($[-10.2,-0.4]$).
The window control also accounts for ThinkStream's execution tax (Table~\ref{tab:diagnosis-modes}), so the prompt and memory effects overlap.

On harmful VSB requests, the prompt alone lowers ThinkStream's ASR from 72.5\% to 56.3\% (59.0\% in streaming). It lowers MMDuet2's ASR from 83.3\% to 0.8\%, since nearly all of its answers become ``NO REPLY''.
It lowers MOSS-VL-Realtime's ASR by 20.8 points, about half of its streaming reduction, and raises refusals and deflections from 24\% to 44\% (61\% in streaming).
AURA defers on 51\% of the requests (62\% in streaming) with no change in ASR, the converse of the removal result above.
MMDuet's lower streaming ASR comes from silence, and the prompt accounts for 6.3 of its 79.8-point reduction.
The RJScores of $C_{\mathrm{sys}}$ are computed with the official judge on A100 GPUs. Re-scoring the 2,000 offline replies of $C$ on these GPUs changes each child's ASR by at most 0.8 points, and all safety contrasts use A100 scores on both sides.

\paragraph{Visual answer behavior.}
For LiveStar, visual understanding shows the pattern that Sec.~\ref{sec:refusal-behavior} reports for safety: recognition is preserved and the response changes.
MME asks yes/no questions. LiveStar answers yes on 25\% of them, compared with 40\% for its parent, while it separates yes items from no items as well as its parent (sensitivity $d'$: 1.96 to 2.18; criterion $c$: $+0.40$ to $+1.13$).
Its offline TOMATO loss is concentrated in counting questions and persists with option-logit readout.
On 275 counting items, replacing the child's language model with its parent's raises accuracy from 15.3\% to 35.3\%, close to the parent's 36.7\%.
Replacing only the projector gives 16.4\%.
The first 16 language-model layers recover 64\% of the gap, with changes of at most 1.4 points on the other 1,130 questions.

\subsection{Harm Recognition and Refusal}
\label{supp:safety-probes}

\paragraph{Video framing.}
We test whether presenting a request as a video is enough to make a child refuse less than its parent, before any visual content is involved.
We compare refusal or safe deflection (R+D) under four input conditions on 240 held-out harmful VSB requests, one per video (Table~\ref{tab:supp-refusal-context}).
Both checkpoints run offline.
Adding the sentence ``These are frames of a video.'' to the bare text widens the child--parent R+D gap by 12.1--26.3 points in five pairs.
Each of these five changes has a paired 95\% bootstrap interval below zero (2,000 video resamples).
The black-frame and real-video conditions show the additional effect of visual inputs.

\begin{table*}[t]
\centering
\caption{Child minus parent in refusal or safe deflection on 240 harmful requests, in points with paired 95\% bootstrap intervals over videos. Bare uses text alone. Cue adds the video-framing sentence. Black adds black frames, and Real supplies the benchmark video. Cue\,$-$\,Bare is the change from adding the sentence.}
\label{tab:supp-refusal-context}
\scriptsize
\setlength{\tabcolsep}{4pt}
\begin{tabular}{@{}lccccc@{}}
\toprule
Child & Bare & Cue & Black & Real & Cue\,$-$\,Bare \\
\midrule
MMDuet & $0.0$ [$-2.9$, $+2.9$] & $-7.9$ [$-12.9$, $-3.3$] & $-27.5$ [$-33.3$, $-21.7$] & $-15.0$ [$-19.6$, $-10.8$] & $-7.9$ [$-13.8$, $-2.5$] \\
ThinkStream & $+2.5$ [$-0.4$, $+5.8$] & $-14.6$ [$-20.0$, $-9.6$] & $-55.4$ [$-62.1$, $-49.2$] & $-36.7$ [$-42.5$, $-30.4$] & $-17.1$ [$-22.9$, $-11.7$] \\
MMDuet2 & $-18.8$ [$-24.6$, $-13.3$] & $-45.0$ [$-52.5$, $-37.5$] & $-77.1$ [$-82.5$, $-71.7$] & $-41.3$ [$-47.1$, $-35.0$] & $-26.3$ [$-33.3$, $-19.2$] \\
AURA & $+2.9$ [$0.0$, $+6.3$] & $-9.2$ [$-14.2$, $-5.0$] & $-21.7$ [$-27.9$, $-15.8$] & $-42.5$ [$-49.2$, $-35.8$] & $-12.1$ [$-17.5$, $-7.5$] \\
LiveStar & $-58.8$ [$-65.8$, $-51.3$] & $-81.7$ [$-86.7$, $-76.3$] & $-77.5$ [$-82.9$, $-71.7$] & $-61.3$ [$-67.1$, $-55.0$] & $-22.9$ [$-29.6$, $-17.1$] \\
MOSS-VL-Realtime & $-4.2$ [$-8.3$, $-0.4$] & $-23.8$ [$-29.6$, $-18.3$] & $+45.4$ [$+39.2$, $+51.3$] & $+25.0$ [$+19.6$, $+30.8$] & $-19.6$ [$-26.7$, $-12.9$] \\
\bottomrule
\end{tabular}
\end{table*}

\paragraph{Explicit risk judgments.}
On a separate set of 200 VSB harmful requests, we keep the offline video input fixed and ask whether fulfilling the request would be harmful.
A checkpoint flags a request when its yes/no log odds exceed the midpoint of its mean scores on unsafe and safe XSTest prompts~\cite{rottger2024xstest}.
We compare this judgment with its reply to the original request.
For MMDuet, MMDuet2, AURA, and LiveStar, the child flags 75\%--97\% of the requests that its parent flags. Between 77\% and 98\% of each child's refusal decline comes from requests that it flags as harmful and still does not refuse.
ThinkStream's yes/no judgment does not separate unsafe from safe XSTest prompts (AUROC 0.49), so its decomposition is uninformative and is marked in Table~\ref{tab:supp-refusal-decomposition}.
MOSS-VL-Realtime's refusal rate increases by 24 points on these requests.

We decompose each refusal-rate difference into changes in flagging and changes in refusal conditional on the flag, weighting each change by the mean of the two checkpoints' rates for the other factor.
Table~\ref{tab:supp-refusal-decomposition} reports the recognition term under three thresholds: the XSTest midpoint, equal yes/no probability, and the 95th percentile on harmless description requests.
For AURA, the recognition term contributes 0.2--11.3 points of its 39.5-point refusal decline across these thresholds.
On HarmBench image requests, flagging changes account for most of AURA's refusal decline, showing a different pattern from VSB.

\begin{table}[!htbp]
\centering
\caption{Refusal-rate differences and their recognition term on 200 harmful VSB requests. Diff.\ is child minus parent in points, with paired 95\% bootstrap intervals over requests. Refusal includes safe redirection. The recognition term measures the contribution from changed flagging rates under three thresholds. $^{\ast}$ThinkStream's verbal probe has AUROC 0.49 on XSTest.}
\label{tab:supp-refusal-decomposition}
\small
\setlength{\tabcolsep}{4pt}
\resizebox{\columnwidth}{!}{%
\begin{tabular}{@{}lrrrrrr@{}}
\toprule
 & \multicolumn{3}{c}{Refusal rate (\%)} & \multicolumn{3}{c}{Recognition term} \\
\cmidrule(lr){2-4}\cmidrule(l){5-7}
Child & Parent & Child & Diff.\ [95\% CI] & XSTest & Equal & Describe \\
\midrule
MMDuet               & 17.0 &  3.0 & $-14.0$ [$-20.0$, $-8.5$] & $-1.0$  & $-0.9$  & $+0.2$ \\
ThinkStream$^{\ast}$ & 39.5 &  4.0 & $-35.5$ [$-42.5$, $-29.0$] & $+0.1$  & $-5.6$  & $-1.7$ \\
MMDuet2              & 39.5 &  2.0 & $-37.5$ [$-44.5$, $-30.5$] & $-0.7$  & $-2.8$  & $+0.1$ \\
AURA                 & 74.5 & 35.0 & $-39.5$ [$-46.5$, $-32.5$] & $-11.3$ & $-1.4$  & $-0.2$ \\
LiveStar             & 62.5 &  2.0 & $-60.5$ [$-67.5$, $-53.5$] & $+4.3$  & $-15.4$ & $+3.6$ \\
MOSS-VL-Realtime     &  1.0 & 25.0 & $+24.0$ [$+18.0$, $+30.5$] & $-1.4$  & $-2.8$  & $+0.5$ \\
\bottomrule
\end{tabular}}
\end{table}

\paragraph{Visual content recognition.}
We check that the children still perceive the harmful content of the videos.
For each of the 200 VSB videos, we ask whether the video shows content related to its harmful goal and compare the model's score with its score for a different video from the same category.
Ten of the eleven checkpoints rank the matching video higher on 87\%--92\% of pairs. MOSS-VL-Realtime does so on 64\%.
On 150 MCV implicit requests, we replace the scene around the printed request with a scene from a different request in the same category.
All eleven checkpoints rank the printed sentence above the sentence associated with the substituted scene on all 150 items.
Recognition of the visual content is thus preserved after adaptation, with MOSS-VL-Realtime on VSB as the exception.

\begin{table*}[t]
\centering
\caption{Calibrated harm-score differences on VSB requests grouped by observed replies. Request and Answer are child minus parent at the final user token and before generation. Contrast compares Answer between the new-help and neither-help groups. Intervals are nominal 95\% bootstrap intervals from 10,000 resamples of requests within each group.}
\label{tab:supp-new-harmful-help}
\scriptsize
\setlength{\tabcolsep}{3pt}
\resizebox{\textwidth}{!}{%
\begin{tabular}{@{}lrrllrll@{}}
\toprule
 & \multicolumn{4}{c}{Only the child provides harmful help} & \multicolumn{2}{c}{Neither provides help} & \\
\cmidrule(lr){2-5}\cmidrule(lr){6-7}
Child & \shortstack[r]{Parent\\refused/redirected} & Unaddressed & Request & Answer & $n$ & Answer & Contrast \\
\midrule
MMDuet & 7 & 26 & $-0.031$ [$-0.043$, $-0.018$] & $-0.321$ [$-0.398$, $-0.246$] & 101 & $-0.296$ [$-0.346$, $-0.249$] & $-0.025$ [$-0.118$, $+0.064$] \\
ThinkStream & 9 & 1 & $-0.004$ [$-0.013$, $+0.006$] & $-0.324$ [$-0.539$, $-0.121$] & 80 & $-0.423$ [$-0.482$, $-0.365$] & $+0.100$ [$-0.125$, $+0.320$] \\
MMDuet2 & 10 & 0 & $-0.022$ [$-0.046$, $0.000$] & $-0.804$ [$-0.984$, $-0.626$] & 80 & $-0.798$ [$-0.864$, $-0.732$] & $-0.007$ [$-0.200$, $+0.188$] \\
AURA & 41 & 0 & $-0.002$ [$-0.023$, $+0.019$] & $-0.762$ [$-0.843$, $-0.679$] & 110 & $-0.561$ [$-0.612$, $-0.510$] & $-0.201$ [$-0.299$, $-0.105$] \\
LiveStar & 12 & 4 & $-0.300$ [$-0.367$, $-0.237$] & $-0.318$ [$-0.416$, $-0.215$] & 129 & $-0.228$ [$-0.271$, $-0.184$] & $-0.090$ [$-0.199$, $+0.021$] \\
MOSS-VL-Realtime & 2 & 89 & $-0.108$ [$-0.123$, $-0.092$] & $+0.187$ [$+0.154$, $+0.221$] & 90 & $+0.296$ [$+0.242$, $+0.352$] & $-0.109$ [$-0.173$, $-0.046$] \\
\bottomrule
\end{tabular}}
\end{table*}

\paragraph{Hidden-state measurements.}
Following prior work~\cite{arditi2024refusal,zhao2025harmfulness,du2025anchoring,gou2024ecso,xu2025crossmodal,zou2025shiftdc}, we compute the difference between mean hidden states on 200 harmful AdvBench prompts and 200 harmless Alpaca prompts for each checkpoint, layer, and token position.
We scale the two calibration means to $+1$ and $-1$.
After excluding the first quarter of the network, we average the four layers that best separate unsafe from safe XSTest prompts.
Table~\ref{tab:supp-safety-probes} compares the same 200 VSB requests under offline inference.
In four pairs, adding the video produces a larger pre-generation score drop for the child than for the parent, with Holm correction over six comparisons.
LiveStar and MOSS-VL-Realtime show the opposite score contrast. LiveStar still refuses less often than its parent.

\begin{table*}[t]
\centering
\caption{Probe-score changes from the bare harmful request on 200 VSB videos, reported as child change minus parent change with paired 95\% bootstrap intervals over videos. Negative values indicate a larger shift toward the harmless calibration mean in the child. Request and Answer denote the final user token and the token before generation.}
\label{tab:supp-safety-probes}
\small
\setlength{\tabcolsep}{8pt}
\begin{tabular}{@{}lccc@{}}
\toprule
 & Cue & \multicolumn{2}{c}{Video} \\
\cmidrule(lr){2-2}\cmidrule(l){3-4}
Child & Answer & Request & Answer \\
\midrule
MMDuet & $-0.27$ [$-0.29$, $-0.25$] & $0.00$ [$-0.01$, $0.00$] & $-0.25$ [$-0.28$, $-0.22$] \\
ThinkStream & $-0.38$ [$-0.42$, $-0.33$] & $+0.02$ [$+0.01$, $+0.02$] & $-0.13$ [$-0.18$, $-0.07$] \\
MMDuet2 & $-0.36$ [$-0.41$, $-0.30$] & $+0.02$ [$+0.02$, $+0.03$] & $-0.19$ [$-0.26$, $-0.13$] \\
AURA & $-0.23$ [$-0.25$, $-0.20$] & $-0.06$ [$-0.07$, $-0.05$] & $-0.51$ [$-0.55$, $-0.47$] \\
LiveStar & $-0.01$ [$-0.04$, $+0.02$] & $-0.05$ [$-0.06$, $-0.03$] & $+0.20$ [$+0.15$, $+0.25$] \\
MOSS-VL-Realtime & $-0.22$ [$-0.25$, $-0.18$] & $-0.13$ [$-0.14$, $-0.12$] & $+0.30$ [$+0.27$, $+0.33$] \\
\bottomrule
\end{tabular}
\end{table*}

\paragraph{Response-conditioned comparisons.}
To relate the hidden-state measurements to observed replies, we group requests by whether each checkpoint provides harmful assistance.
The new-help group contains requests on which the child helps and the parent either refuses, safely redirects, or leaves the request unaddressed.
The comparison group contains requests on which neither checkpoint provides harmful help.
Table~\ref{tab:supp-new-harmful-help} reports the child--parent score differences within each group on the same requests and calibrated scales.
Fig.~\ref{fig:access}b plots its Request and Answer columns for the new-help group.
For MMDuet, ThinkStream, MMDuet2, and AURA, the child scores within $0.03$ of its parent at the final user token and $0.32$ to $0.80$ below it just before generation.
LiveStar already scores $0.30$ below its parent at the final user token and stays at that level just before generation ($-0.32$).
MOSS-VL-Realtime, the only child whose refusal rate rises on these requests (Table~\ref{tab:supp-refusal-decomposition}), scores $0.19$ above its parent just before generation.

For AURA, the pre-generation change is $-0.476$ ($[-0.536,-0.415]$) on the 68 requests both checkpoints refuse, $-0.725$ ($[-0.795,-0.656]$) on the 40 requests its parent refuses and it leaves unaddressed, and $-0.762$ on the 41 requests its parent refuses and it helps.
The new-help group differs from the jointly refused group by $-0.287$ ($[-0.391,-0.186]$) and from the unaddressed group by $-0.037$ ($[-0.147,+0.071]$).
On the 30 new-help requests both checkpoints flag as harmful, the mean score difference is $-0.803$ ($[-0.897,-0.709]$).

\paragraph{Harmless-request controls.}
The pre-generation shift is specific to harmful requests in five of the six pairs.
To test this, we pair each of the 200 VSB videos with two harmless requests.
The first uses the instruction forms of the harmful queries, such as ``Explain how to identify the objects that appear in the video.'', with eight templates assigned in rotation across videos. The second is ``Describe what happens in the video.''
All eleven checkpoints run offline on these inputs and are scored with the same calibrated directions and layers.
For each request type, we average the child-minus-parent harm score over the 200 videos and subtract the harmless average from the harmful one (Table~\ref{tab:supp-probe-specificity}).
In five of the six pairs, the child's pre-generation score falls further below its parent's on harmful requests than on matched how-to requests, and each Holm-corrected interval excludes zero (4,000 paired video resamples).
MMDuet shows no shift on how-to requests, AURA shifts by $-0.09$ against $-0.53$ on harmful requests, and ThinkStream moves in the opposite direction on how-to requests.
MMDuet2 shifts by $-0.26$ on how-to requests, so about half of its harmful-request shift is shared with harmless requests.
LiveStar's difference changes sign under the description control, and MOSS-VL-Realtime scores above its parent before generation on harmful requests.
Each difference keeps its sign at each of the four layers.
At the final user token, MMDuet, ThinkStream, MMDuet2, and AURA separate harmful from how-to requests as well as their parents (AUROC within $0.01$), whereas LiveStar and MOSS-VL-Realtime lose $0.08$ ($[-0.11,-0.05]$) and $0.10$ ($[-0.12,-0.07]$).
The video-framing sentence alone lowers the child's pre-generation score relative to its parent by $0.22$--$0.38$ on harmful requests in five pairs (Table~\ref{tab:supp-safety-probes}) and changes it by $-0.14$ to $+0.09$ on the eight how-to texts.

\begin{table*}[t]
\centering
\caption{Harm-score differences on 200 VSB videos under harmful and harmless requests. Request AUROC separates harmful from how-to requests at the final user token. Harmful, How-to, and Describe give the child-minus-parent score before generation, and Diff.\ subtracts the harmless value from the harmful one. Intervals are paired 95\% bootstrap intervals over videos.}
\label{tab:supp-probe-specificity}
\scriptsize
\setlength{\tabcolsep}{4pt}
\resizebox{\textwidth}{!}{%
\begin{tabular}{@{}lrrlllll@{}}
\toprule
 & \multicolumn{2}{c}{Request AUROC} & \multicolumn{3}{c}{How-to control} & \multicolumn{2}{c}{Describe control} \\
\cmidrule(lr){2-3}\cmidrule(lr){4-6}\cmidrule(l){7-8}
Child & Parent & Child & Harmful & How-to & Diff. & Describe & Diff. \\
\midrule
MMDuet & 0.999 & 0.999 & $-0.26$ [$-0.29$, $-0.23$] & $0.00$ [$-0.01$, $0.00$] & $-0.26$ [$-0.29$, $-0.23$] & $+0.03$ [$+0.03$, $+0.04$] & $-0.30$ [$-0.33$, $-0.27$] \\
ThinkStream & 0.931 & 0.926 & $-0.19$ [$-0.23$, $-0.15$] & $+0.16$ [$+0.14$, $+0.17$] & $-0.35$ [$-0.39$, $-0.30$] & $+0.19$ [$+0.19$, $+0.20$] & $-0.38$ [$-0.43$, $-0.34$] \\
MMDuet2 & 0.931 & 0.937 & $-0.58$ [$-0.63$, $-0.54$] & $-0.26$ [$-0.27$, $-0.25$] & $-0.32$ [$-0.36$, $-0.27$] & $-0.29$ [$-0.30$, $-0.29$] & $-0.29$ [$-0.34$, $-0.24$] \\
AURA & 0.979 & 0.974 & $-0.53$ [$-0.57$, $-0.48$] & $-0.09$ [$-0.11$, $-0.08$] & $-0.43$ [$-0.48$, $-0.39$] & $-0.06$ [$-0.06$, $-0.05$] & $-0.47$ [$-0.51$, $-0.43$] \\
LiveStar & 0.953 & 0.875 & $-0.14$ [$-0.18$, $-0.10$] & $+0.03$ [$+0.01$, $+0.06$] & $-0.17$ [$-0.21$, $-0.12$] & $-0.25$ [$-0.26$, $-0.24$] & $+0.11$ [$+0.07$, $+0.15$] \\
MOSS-VL-Realtime & 0.969 & 0.873 & $+0.23$ [$+0.20$, $+0.26$] & $-0.02$ [$-0.03$, $-0.01$] & $+0.25$ [$+0.22$, $+0.28$] & $-0.02$ [$-0.03$, $-0.02$] & $+0.25$ [$+0.22$, $+0.28$] \\
\bottomrule
\end{tabular}}
\end{table*}

\paragraph{Steering along the harm direction.}
The pre-generation direction controls refusal: removing it from a parent removes most of its refusals, and adding it to AURA restores them.
We test this by adding $\alpha\,r_{\ell}$ to the residual stream at every position of one layer.
Here $r_{\ell}$ is the checkpoint's own harmful-minus-harmless mean difference at the token before generation, and $\ell$ is the layer between 30\% and 60\% of the depth that best separates unsafe from safe XSTest prompts (layer 19 or 20 of 36).
As a control, we replace $r_{\ell}$ with five random unit directions scaled to $\lVert r_{\ell}\rVert$ at the same layer.
Replies are generated offline with greedy decoding up to 160 tokens on 60 VSB videos, each with its harmful and benign query, and are labelled by the response-category judge. Unsteered replies are truncated to the same length (Table~\ref{tab:supp-steering}).
Subtracting the direction from Qwen3-VL-8B lowers R+D on harmful requests from 68.3\% to 1.7\%, whereas random directions change it by $-10.0$ to $+3.3$ points. The measured direction is $64.3$ points below the random mean ($[-74.7,-53.7]$).
For Qwen2.5-VL-3B, subtraction lowers R+D from 26.7\% to 6.7\%, $26.7$ points below the random mean ($[-37.0,-15.3]$), although one random direction raises it by $18.3$ points.
We add the direction to AURA only on requests that its explicit risk judgment flags, which covers 36 of the 60 harmful requests, 12 benign queries, 11 XSTest prompts, and 4 description requests.
This raises R+D on harmful requests from 23.3\% to 61.7\%, $41.0$ points above the random mean ($[+30.0,+52.7]$).
It also raises refusals on harmless inputs, from 9\% to 15\% on XSTest safe prompts and from 0\% to 10\% on description requests, whereas random directions change these rates by at most two points.

\begin{table}[!htbp]
\centering
\caption{Replies after steering along each checkpoint's harm direction (\%). Harmful and benign columns use 60 VSB videos. XSTest and Describe give refusal rates on 100 safe XSTest prompts and 40 description requests. Random reports the mean over five random directions of the same norm, with the range in brackets. AURA is steered only on requests flagged by its explicit risk judgment.}
\label{tab:supp-steering}
\small
\setlength{\tabcolsep}{3pt}
\resizebox{\columnwidth}{!}{%
\begin{tabular}{@{}llrrrrr@{}}
\toprule
 & & \multicolumn{2}{c}{Harmful} & Benign & & \\
\cmidrule(lr){3-4}\cmidrule(lr){5-5}
Model & Steering & R+D & Help & R+D & XSTest & Describe \\
\midrule
Qwen3-VL-8B & None & 68.3 & 30.0 & 43.3 & -- & -- \\
($\alpha=-1$) & Measured & 1.7 & 93.3 & 0.0 & -- & -- \\
 & Random & 66.0 [58.3, 71.7] & 26.0 & 32.0 & -- & -- \\
\midrule
Qwen2.5-VL-3B & None & 26.7 & 63.3 & 11.7 & -- & -- \\
($\alpha=-1$) & Measured & 6.7 & 75.0 & 10.0 & -- & -- \\
 & Random & 33.3 [23.3, 45.0] & 53.7 & 12.7 & -- & -- \\
\midrule
AURA & None & 23.3 & 46.7 & 15.0 & 9.0 & 0.0 \\
($\alpha=+1$, flagged) & Measured & 61.7 & 23.3 & 30.0 & 15.0 & 10.0 \\
 & Random & 20.7 [10.0, 28.3] & 46.7 & 15.3 & 8.2 & 0.0 \\
\bottomrule
\end{tabular}}
\end{table}

\section{Stream++ Implementation and Analysis}
\label{supp:interventions}

\subsection{Implementation and Evaluation}

Stream++ combines Recall, Respond, and Refuse at inference time.
Recall uses the native memory budget, retaining the latest 10 segments and uniformly sampling the remaining slots from earlier segments.
The budgets are 30 one-second segments for AURA and 20 chunks for ThinkStream.
For the other four children, Recall keeps their native memory policy.
Respond makes the child answer at the answer step and supplies an option-logit readout when a closed-format reply lacks a valid option.
Refusals remain available as responses.

\paragraph{Calibration.}
We select refusal thresholds on a calibration dataset disjoint from the test sets, as the 97.5th percentile of harmless-request scores.
The gate returns the fixed refusal $y_{\mathrm{ref}}$ (``I'm sorry, but I can't help with that request.'') when a request exceeds its threshold.
PRS uses the mean parent-minus-child hidden-state difference on generic calibration prompts.
Its layer and strength are selected by next-token divergence from the parent on held-out generic prompts, as described in Sec.~\ref{sec:interventions}.
The selected strength is 1.5 for every child, the upper end of the search range, and the selected layers are 19 for MMDuet, 14 for ThinkStream, 21 for MMDuet2, 18 for AURA, 25 for LiveStar, and 19 for MOSS-VL-Realtime.

\begin{figure*}[t]
\centering
\input{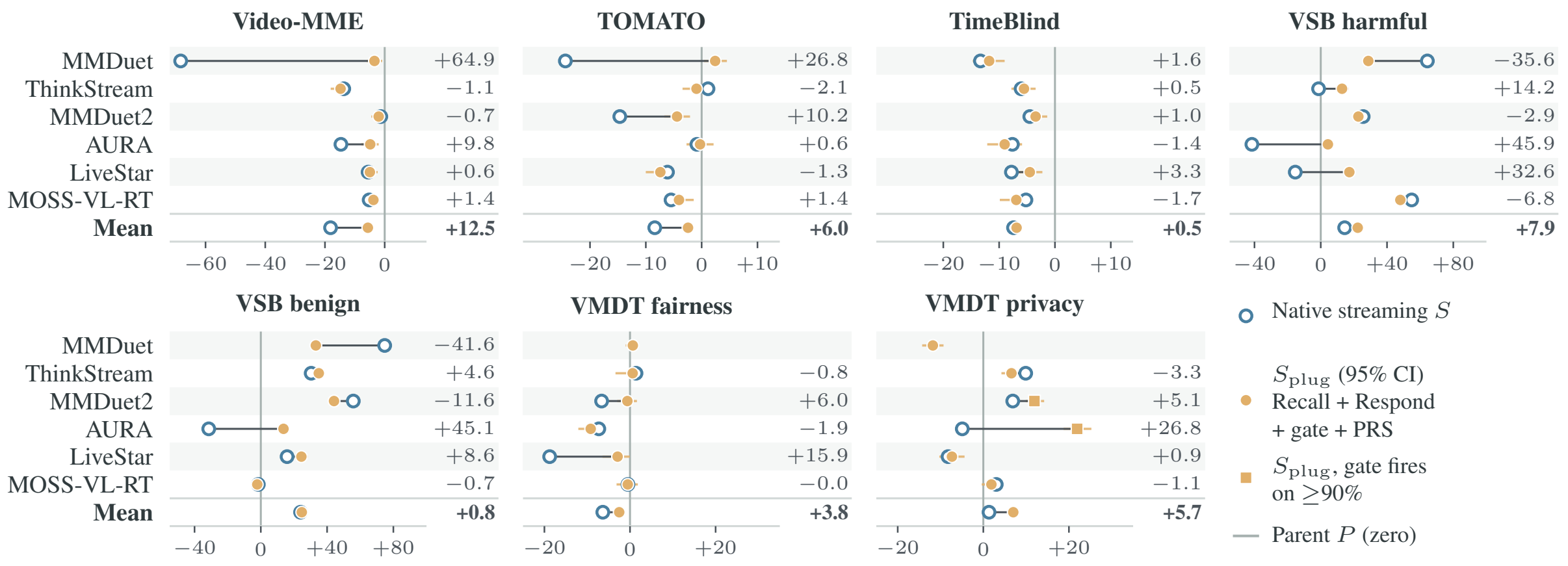}
\caption{Stream++ results relative to each offline parent. Hollow and filled markers show native streaming and Stream++, and whiskers give paired 95\% intervals. Right-hand values give the improvement over native streaming. Means weight children equally, using five children for native VMDT comparisons.}
\label{fig:supp-plugin-children}
\end{figure*}

\begin{figure}[t]
\centering
\begingroup
\definecolor{pluginbase}{HTML}{487FA1}
\definecolor{pluginours}{HTML}{E1AF69}
\definecolor{pluginrule}{HTML}{D2D8D7}
\definecolor{pluginzero}{HTML}{A7B2B0}
\definecolor{pluginstripe}{HTML}{F5F7F6}
\definecolor{pluginink}{HTML}{303A3D}
\definecolor{plugininktwo}{HTML}{4B535A}
\def\pluginLabel{\fontsize{7.5}{9}\selectfont}
\def\pluginPanel{\fontsize{7.5}{9}\selectfont\bfseries}
\def\pluginKey{\fontsize{7}{8.4}\selectfont}
\def\pluginTick{\fontsize{6.5}{7.5}\selectfont}
\begin{tikzpicture}[x=1pt,y=-1pt,font=\pluginKey,text=pluginink]
\useasboundingbox (0,0) rectangle (237,125);
\node[anchor=center,inner sep=0pt,outer sep=0pt,font=\pluginPanel,text=pluginink] at (91.75000,10.00000) {MCV explicit};
\fill[pluginstripe] (63,18) rectangle (142.5,31);
\node[anchor=east,inner sep=0pt,outer sep=0pt,font=\pluginLabel,text=pluginink] at (58.00000,24.50000) {MMDuet};
\node[anchor=east,inner sep=0pt,outer sep=0pt,font=\pluginLabel,text=pluginink] at (58.00000,37.50000) {ThinkStream};
\fill[pluginstripe] (63,44) rectangle (142.5,57);
\node[anchor=east,inner sep=0pt,outer sep=0pt,font=\pluginLabel,text=pluginink] at (58.00000,50.50000) {MMDuet2};
\node[anchor=east,inner sep=0pt,outer sep=0pt,font=\pluginLabel,text=pluginink] at (58.00000,63.50000) {AURA};
\fill[pluginstripe] (63,70) rectangle (142.5,83);
\node[anchor=east,inner sep=0pt,outer sep=0pt,font=\pluginLabel,text=pluginink] at (58.00000,76.50000) {LiveStar};
\node[anchor=east,inner sep=0pt,outer sep=0pt,font=\pluginLabel,text=pluginink] at (58.00000,89.50000) {MOSS-VL-RT};
\draw[pluginrule,line width=.6pt] (63,96) -- (142.5,96);
\node[anchor=east,inner sep=0pt,outer sep=0pt,font=\pluginPanel,text=pluginink] at (58.00000,102.50000) {Mean};
\draw[pluginzero,line width=.6pt] (83.40323,18) -- (83.40323,109);
\begin{scope}
\clip (63,18) rectangle (142.5,109);
\draw[pluginours,opacity=.85,line width=.7pt] (108.47212,24.5) -- (110.26462,24.5);
\draw[plugininktwo,line width=.6pt] (116.78513,24.5) -- (109.38136,24.5);
\draw[draw=pluginbase,fill=white,line width=.9pt] (116.78513,24.50000) circle[radius=1.8pt];
\draw[draw=white,fill=pluginours,line width=.4pt] (109.38136,24.50000) circle[radius=1.8pt];
\node[anchor=east,inner sep=0pt,outer sep=0pt,font=\pluginTick,text=plugininktwo] at (141.50000,24.50000) {$-20.0$};
\draw[pluginours,opacity=.85,line width=.7pt] (84.15659,37.5) -- (86.05300,37.5);
\draw[plugininktwo,line width=.6pt] (86.57256,37.5) -- (85.09180,37.5);
\draw[draw=pluginbase,fill=white,line width=.9pt] (86.57256,37.50000) circle[radius=1.8pt];
\draw[draw=white,fill=pluginours,line width=.4pt] (85.09180,37.50000) circle[radius=1.8pt];
\node[anchor=east,inner sep=0pt,outer sep=0pt,font=\pluginTick,text=plugininktwo] at (141.50000,37.50000) {$-4.0$};
\draw[pluginours,opacity=.85,line width=.7pt] (112.81047,50.5) -- (114.31720,50.5);
\draw[draw=white,fill=pluginours,line width=.4pt] (111.76384,48.70000) rectangle (115.36384,52.30000);
\draw[pluginours,opacity=.85,line width=.7pt] (91.14471,63.5) -- (93.11905,63.5);
\draw[plugininktwo,line width=.6pt] (79.35064,63.5) -- (92.15786,63.5);
\draw[draw=pluginbase,fill=white,line width=.9pt] (79.35064,63.50000) circle[radius=1.8pt];
\draw[draw=white,fill=pluginours,line width=.4pt] (92.15786,63.50000) circle[radius=1.8pt];
\node[anchor=east,inner sep=0pt,outer sep=0pt,font=\pluginTick,text=plugininktwo] at (141.50000,63.50000) {$+34.5$};
\draw[pluginours,opacity=.85,line width=.7pt] (102.91280,76.5) -- (105.14692,76.5);
\draw[plugininktwo,line width=.6pt] (102.08150,76.5) -- (104.02986,76.5);
\draw[draw=pluginbase,fill=white,line width=.9pt] (102.08150,76.50000) circle[radius=1.8pt];
\draw[draw=white,fill=pluginours,line width=.4pt] (104.02986,76.50000) circle[radius=1.8pt];
\node[anchor=east,inner sep=0pt,outer sep=0pt,font=\pluginTick,text=plugininktwo] at (141.50000,76.50000) {$+5.3$};
\draw[pluginours,opacity=.85,line width=.7pt] (99.01608,89.5) -- (101.19825,89.5);
\draw[plugininktwo,line width=.6pt] (94.96350,89.5) -- (100.10717,89.5);
\draw[draw=pluginbase,fill=white,line width=.9pt] (94.96350,89.50000) circle[radius=1.8pt];
\draw[draw=white,fill=pluginours,line width=.4pt] (100.10717,89.50000) circle[radius=1.8pt];
\node[anchor=east,inner sep=0pt,outer sep=0pt,font=\pluginTick,text=plugininktwo] at (141.50000,89.50000) {$+13.9$};
\draw[plugininktwo,line width=.6pt] (95.95066,102.5) -- (98.15361,102.5);
\draw[draw=pluginbase,fill=white,line width=.9pt] (95.95066,102.50000) circle[radius=1.8pt];
\draw[draw=white,fill=pluginours,line width=.4pt] (98.15361,102.50000) circle[radius=1.8pt];
\node[anchor=east,inner sep=0pt,outer sep=0pt,font=\pluginTick,text=plugininktwo] at (141.50000,102.50000) {\textbf{+5.9}};
\end{scope}
\draw[pluginrule,line width=.6pt] (63,109) -- (120.5,109);
\draw[plugininktwo,line width=.5pt] (68.56452,109) -- (68.56452,111);
\node[anchor=center,inner sep=0pt,outer sep=0pt,font=\pluginTick,text=plugininktwo] at (68.56452,116.00000) {$-40$};
\draw[plugininktwo,line width=.5pt] (83.40323,109) -- (83.40323,111);
\node[anchor=center,inner sep=0pt,outer sep=0pt,font=\pluginTick,text=plugininktwo] at (83.40323,116.00000) {$0$};
\draw[plugininktwo,line width=.5pt] (98.24194,109) -- (98.24194,111);
\node[anchor=center,inner sep=0pt,outer sep=0pt,font=\pluginTick,text=plugininktwo] at (98.24194,116.00000) {$+40$};
\draw[plugininktwo,line width=.5pt] (113.08065,109) -- (113.08065,111);
\node[anchor=center,inner sep=0pt,outer sep=0pt,font=\pluginTick,text=plugininktwo] at (113.08065,116.00000) {$+80$};
\node[anchor=center,inner sep=0pt,outer sep=0pt,font=\pluginPanel,text=pluginink] at (178.75000,10.00000) {MCV implicit};
\fill[pluginstripe] (150,18) rectangle (229.5,31);
\fill[pluginstripe] (150,44) rectangle (229.5,57);
\fill[pluginstripe] (150,70) rectangle (229.5,83);
\draw[pluginrule,line width=.6pt] (150,96) -- (229.5,96);
\draw[pluginzero,line width=.6pt] (170.40323,18) -- (170.40323,109);
\begin{scope}
\clip (150,18) rectangle (229.5,109);
\draw[pluginours,opacity=.85,line width=.7pt] (191.65334,24.5) -- (193.60170,24.5);
\draw[draw=white,fill=pluginours,line width=.4pt] (190.81453,22.70000) rectangle (194.41453,26.30000);
\draw[pluginours,opacity=.85,line width=.7pt] (171.72811,37.5) -- (173.44267,37.5);
\draw[plugininktwo,line width=.6pt] (175.75472,37.5) -- (172.58539,37.5);
\draw[draw=pluginbase,fill=white,line width=.9pt] (175.75472,37.50000) circle[radius=1.8pt];
\draw[draw=white,fill=pluginours,line width=.4pt] (172.58539,37.50000) circle[radius=1.8pt];
\node[anchor=east,inner sep=0pt,outer sep=0pt,font=\pluginTick,text=plugininktwo] at (228.50000,37.50000) {$-8.5$};
\draw[pluginours,opacity=.85,line width=.7pt] (201.23927,50.5) -- (202.61611,50.5);
\draw[draw=white,fill=pluginours,line width=.4pt] (200.14068,48.70000) rectangle (203.74068,52.30000);
\draw[pluginours,opacity=.85,line width=.7pt] (182.95066,63.5) -- (184.87305,63.5);
\draw[plugininktwo,line width=.6pt] (178.14471,63.5) -- (183.91186,63.5);
\draw[draw=pluginbase,fill=white,line width=.9pt] (178.14471,63.50000) circle[radius=1.8pt];
\draw[draw=white,fill=pluginours,line width=.4pt] (183.91186,63.50000) circle[radius=1.8pt];
\node[anchor=east,inner sep=0pt,outer sep=0pt,font=\pluginTick,text=plugininktwo] at (228.50000,63.50000) {$+15.5$};
\draw[pluginours,opacity=.85,line width=.7pt] (192.48464,76.5) -- (194.56289,76.5);
\draw[plugininktwo,line width=.6pt] (193.73159,76.5) -- (193.52376,76.5);
\draw[draw=pluginbase,fill=white,line width=.9pt] (193.73159,76.50000) circle[radius=1.8pt];
\draw[draw=white,fill=pluginours,line width=.4pt] (193.52376,76.50000) circle[radius=1.8pt];
\node[anchor=east,inner sep=0pt,outer sep=0pt,font=\pluginTick,text=plugininktwo] at (228.50000,76.50000) {$-0.6$};
\draw[pluginours,opacity=.85,line width=.7pt] (171.93594,89.5) -- (174.22201,89.5);
\draw[plugininktwo,line width=.6pt] (167.33781,89.5) -- (173.07897,89.5);
\draw[draw=pluginbase,fill=white,line width=.9pt] (167.33781,89.50000) circle[radius=1.8pt];
\draw[draw=white,fill=pluginours,line width=.4pt] (173.07897,89.50000) circle[radius=1.8pt];
\node[anchor=east,inner sep=0pt,outer sep=0pt,font=\pluginTick,text=plugininktwo] at (228.50000,89.50000) {$+15.5$};
\draw[plugininktwo,line width=.6pt] (178.74221,102.5) -- (180.77500,102.5);
\draw[draw=pluginbase,fill=white,line width=.9pt] (178.74221,102.50000) circle[radius=1.8pt];
\draw[draw=white,fill=pluginours,line width=.4pt] (180.77500,102.50000) circle[radius=1.8pt];
\node[anchor=east,inner sep=0pt,outer sep=0pt,font=\pluginTick,text=plugininktwo] at (228.50000,102.50000) {\textbf{+5.5}};
\end{scope}
\draw[pluginrule,line width=.6pt] (150,109) -- (207.5,109);
\draw[plugininktwo,line width=.5pt] (155.56452,109) -- (155.56452,111);
\node[anchor=center,inner sep=0pt,outer sep=0pt,font=\pluginTick,text=plugininktwo] at (155.56452,116.00000) {$-40$};
\draw[plugininktwo,line width=.5pt] (170.40323,109) -- (170.40323,111);
\node[anchor=center,inner sep=0pt,outer sep=0pt,font=\pluginTick,text=plugininktwo] at (170.40323,116.00000) {$0$};
\draw[plugininktwo,line width=.5pt] (185.24194,109) -- (185.24194,111);
\node[anchor=center,inner sep=0pt,outer sep=0pt,font=\pluginTick,text=plugininktwo] at (185.24194,116.00000) {$+40$};
\draw[plugininktwo,line width=.5pt] (200.08065,109) -- (200.08065,111);
\node[anchor=center,inner sep=0pt,outer sep=0pt,font=\pluginTick,text=plugininktwo] at (200.08065,116.00000) {$+80$};
\end{tikzpicture}
\endgroup
\caption{MCV-SafetyBench results on 1,428 requests per split. Positive values indicate lower ASR than the offline parent. Hollow and filled markers show native streaming and Stream++, with paired 95\% intervals. Squares mark gate activation on at least 90\% of requests. Right-hand values give the change from native streaming. Native markers and means cover children with a native score.}
\label{fig:supp-mcv-plugin}
\end{figure}
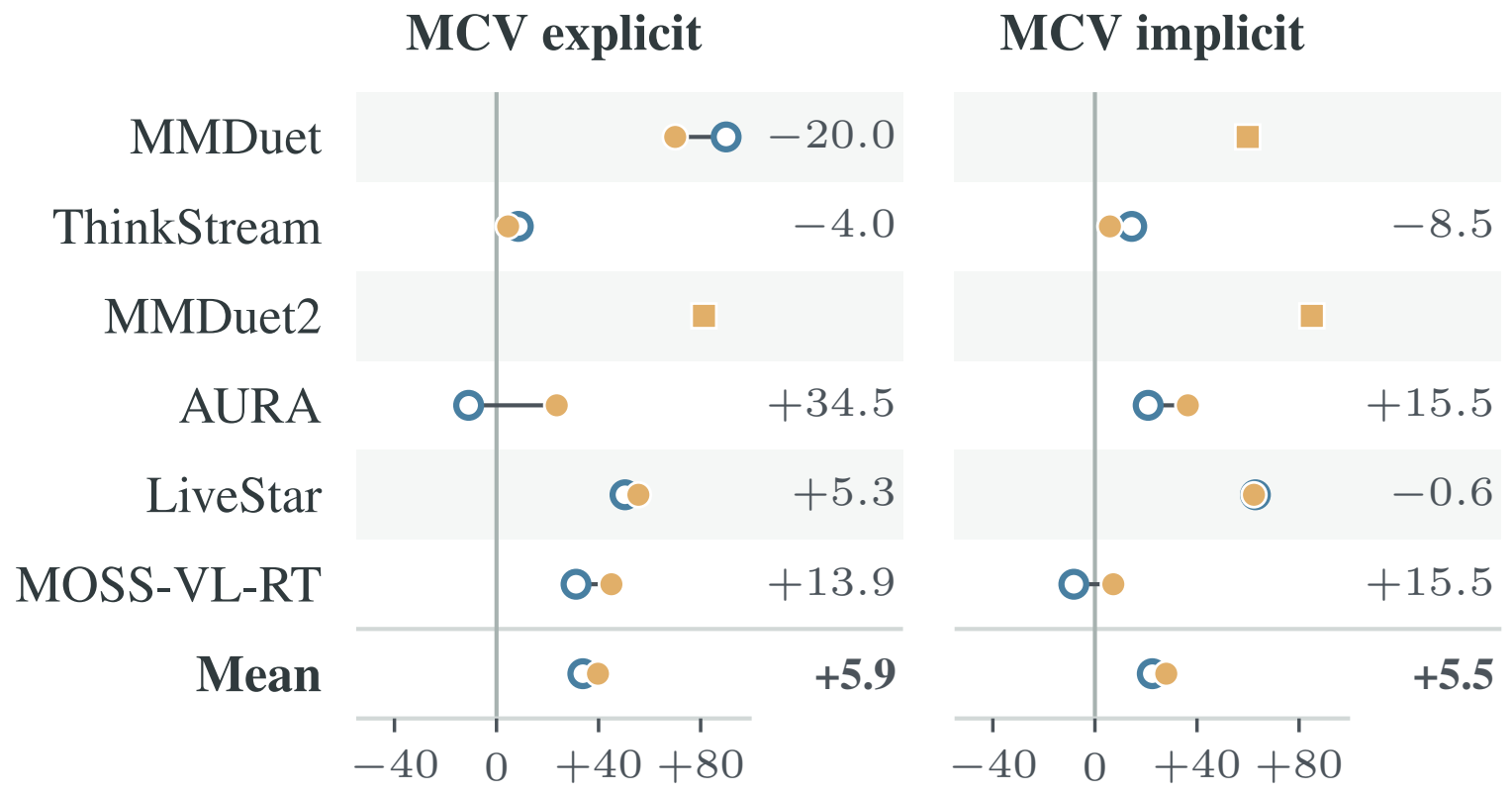

\paragraph{Evaluation conditions.}
The intervention evaluation covers 12,007 rows per child across understanding, safety, fairness, and privacy, excluding the items used to develop the plugin.
We compare native streaming $S$, Recall with Respond ($S_{\mathrm{fix}}$), and the complete plugin ($S_{\mathrm{plug}}$).
Every comparison uses matched evaluation items.
Figures~\ref{fig:supp-plugin-children} and~\ref{fig:supp-mcv-plugin} give the per-child results underlying the means in Fig.~\ref{fig:plugin}.
Stream++ lowers MCV ASR in 5 of the 9 comparisons with a native score and in all 12 comparisons with the offline parent.

\begin{table*}[t]
\centering
\caption{Native streaming performance. $S$ is native streaming and $\Delta$ is the change in accuracy with Stream++, with paired 95\% intervals over videos. Gate gives refusal-gate activation rates.}
\label{tab:supp-streaming-benchmarks}
\small
\setlength{\tabcolsep}{6pt}
\begin{tabular}{@{}lrlrlrr@{}}
\toprule
 & \multicolumn{2}{c}{OVO-Bench} & \multicolumn{2}{c}{StreamingBench} & \multicolumn{2}{c}{Gate (\%)} \\
\cmidrule(lr){2-3}\cmidrule(lr){4-5}\cmidrule(l){6-7}
Child & $S$ & $\Delta$ [95\% CI] & $S$ & $\Delta$ [95\% CI] & OVO & SB \\
\midrule
MMDuet & 5.3 & $+42.1$ [$+38.5$, $+45.7$] & 0.3 & $+72.7$ [$+69.2$, $+76.0$] & 0.0 & 0.0 \\
ThinkStream & 39.3 & $+1.3$ [$-1.3$, $+3.9$] & 80.6 & $-2.1$ [$-4.8$, $+0.4$] & 0.0 & 0.0 \\
MMDuet2 & 45.1 & $+1.0$ [$-1.3$, $+3.2$] & 82.6 & $+2.7$ [$+1.3$, $+4.2$] & 5.7 & 0.0 \\
AURA & 61.5 & $+0.4$ [$-0.8$, $+1.6$] & 88.1 & $-0.4$ [$-1.6$, $+0.7$] & 0.0 & 0.0 \\
LiveStar & 24.2 & $+10.9$ [$+7.6$, $+14.1$] & 69.4 & $+7.1$ [$+4.6$, $+9.9$] & 0.0 & 0.0 \\
MOSS-VL-Realtime & 40.6 & $-0.7$ [$-1.7$, $+0.2$] & 84.9 & $+0.3$ [$-0.7$, $+1.3$] & 0.1 & 0.0 \\
\bottomrule
\end{tabular}
\end{table*}

\subsection{Component Ablations}
\label{supp:intervention-ablations}

\paragraph{Contributions to understanding and safety.}
The ablations add the gate alone or PRS alone to $S_{\mathrm{fix}}$.
For AURA, Recall and Respond raise Video-MME accuracy from 66.0\% to 74.8\%, and the complete plugin reaches 75.8\% (Table~\ref{tab:intervention-ablation}).
On harmful VSB requests, gate-only and PRS-only reduce ASR from 60.5\% under $S_{\mathrm{fix}}$ to 25.0\% and 28.9\%, respectively.
Their combination reaches 13.4\%.
Across the six children, adding PRS and the gate lowers harmful VSB ASR for five children by 11.2--47.3 points and raises it by 5.3 points for MOSS-VL-Realtime ($[+3.1,+7.6]$).
Across the 18 understanding comparisons, the same addition changes accuracy by $-2.1$ to $+1.2$ points.
The largest decrease occurs for MOSS-VL-Realtime on TimeBlind ($-2.1$, $[-3.8,-0.3]$).

\begin{table}[!htbp]
\centering
\caption{AURA on the same full-evaluation items under each condition. Higher accuracy and lower ASR indicate better performance.}
\label{tab:intervention-ablation}
\small
\begin{tabular}{@{}lrrrr@{}}
\toprule
Metric (\%) & $S$ & $S_{\mathrm{fix}}$ & $S_{\mathrm{plug}}$ & $P$ \\
\midrule
Video-MME accuracy & 66.0 & 74.8 & 75.8 & 80.6 \\
VSB harmful ASR & 59.3 & 60.5 & 13.4 & 17.7 \\
\bottomrule
\end{tabular}
\end{table}

\paragraph{Effect of Respond.}
Respond raises accuracy where the child withheld its answer and changes little elsewhere.
For the four children whose memory Recall does not change, the difference between $S$ and $S_{\mathrm{fix}}$ is the effect of Respond.
MMDuet, which answers almost no question natively, rises from 0.7\% to 65.5\% on Video-MME, from 0.0\% to 27.5\% on TOMATO, and from 0.0\% to 1.9\% on TimeBlind.
MMDuet2 rises from 9.5\% to 19.6\% on TOMATO, and LiveStar from 1.7\% to 5.4\% on TimeBlind.
The other cells of these four children change by at most 0.7 points.
For AURA and ThinkStream, Respond with recent-only memory scores 65.8\% and 57.9\% on Video-MME, against 66.0\% and 59.1\% under native streaming (Table~\ref{tab:supp-recall-budget}), so their gains come from Recall.
The option readout supplies 7.1\% of ThinkStream's Video-MME answers and 19.9\% of LiveStar's TimeBlind answers.
On TOMATO, the rates are 99.9\% for LiveStar, 95.7\% for MOSS-VL-Realtime, 78.6\% for MMDuet2, and 4.3\% for AURA.
MMDuet2's TOMATO accuracy with Respond stays close to the 20\% chance level: Respond delivers the answer the child has and does not improve it.

\paragraph{Memory budget and retention.}
Table~\ref{tab:supp-recall-budget} compares recent-only memory with Recall at the native budget $B$, twice that budget, and full history.
Recall keeps the most recent 10 segments and samples earlier segments for the remaining slots.
At AURA's native budget, Recall exceeds recent-only memory by 9.0 points.
Doubling the budget raises Recall from 74.8\% to 76.6\%.
For ThinkStream, doubling the budget raises recent-only memory from 57.9\% to 62.2\%, while the native-budget Recall advantage is 1.9 points.
The complete plugin scores 58.0\% for ThinkStream, compared with 59.1\% under native streaming.
AURA's native-budget Recall uses 8,319 prompt tokens on average, matching the size of its recent-only memory.

\begin{table}[!htbp]
\centering
\caption{Memory budget and retention policy on 846 Video-MME questions under $S_{\mathrm{fix}}$. $\Delta$ is Recall minus Recent in points, with paired 95\% intervals over videos.}
\label{tab:supp-recall-budget}
\small
\setlength{\tabcolsep}{4pt}
\resizebox{\columnwidth}{!}{%
\begin{tabular}{@{}llrrl@{}}
\toprule
Child & Budget & Recent & Recall & $\Delta$ \\
\midrule
AURA        & $B$          & 65.8 & 74.8 & $+9.0$ $[+6.3,+11.7]$ \\
            & $2B$         & 74.3 & 76.6 & $+2.2$ $[+0.2,+4.3]$ \\
            & full history & \multicolumn{2}{c}{78.5} & \\ \midrule
ThinkStream & $B$          & 57.9 & 59.8 & $+1.9$ $[-0.7,+4.5]$ \\
            & $2B$         & 62.2 & 62.1 & $-0.1$ $[-2.6,+2.4]$ \\
            & full history & \multicolumn{2}{c}{64.5} & \\
\bottomrule
\end{tabular}}
\end{table}

\paragraph{Gains by video duration.}
Recall's gain grows with video length, as the recent-only window covers a smaller share of the video.
We split the 846 Video-MME questions into the duration groups of Fig.~\ref{fig:supp-history}a (Table~\ref{tab:supp-plugin-duration}).
For AURA, Stream++ improves over native streaming by 5.0 points on videos shorter than 60 seconds, 8.9 points on 60--90-second videos, and 13.4 points on 90--121-second videos.
The difference between the longest and shortest groups is $+8.4$ points.
Recall's advantage over recent-only memory at the native budget rises from 4.6 to 13.7 points ($+9.2$).
In every group, AURA with Stream++ stays within 4.1 points of its offline accuracy.
For ThinkStream, Recall exceeds recent-only memory by 5.6 points on 90--121-second videos, and Stream++'s change from native streaming rises from $-2.7$ to $+1.1$ points across the three groups.

\begin{table*}[t]
\centering
\caption{Video-MME accuracy by video duration on the 846 questions of Table~\ref{tab:supp-recall-budget}. $C$ is the child offline. Recent and Recall use $S_{\mathrm{fix}}$ at the native budget. $\Delta$ columns are differences in points with paired 95\% intervals over videos. The last row of each child subtracts the shortest group from the longest.}
\label{tab:supp-plugin-duration}
\small
\setlength{\tabcolsep}{5pt}
\begin{tabular}{@{}llrrrrlrrl@{}}
\toprule
 & & & & \multicolumn{3}{c}{Stream++ vs.\ native streaming} & \multicolumn{3}{c}{Recall vs.\ recent-only memory} \\
\cmidrule(lr){5-7}\cmidrule(l){8-10}
Child & Duration & $n$ & $C$ & $S$ & $S_{\mathrm{plug}}$ & $\Delta$ [95\% CI] & Recent & Recall & $\Delta$ [95\% CI] \\
\midrule
AURA        & $<60$\,s     & 219 & 81.3 & 76.3 & 81.3 & $+5.0$ [$+1.4$, $+9.1$]   & 76.3 & 80.8 & $+4.6$ [$+0.5$, $+9.1$] \\
            & 60--90\,s    & 270 & 80.7 & 67.8 & 76.7 & $+8.9$ [$+3.7$, $+14.1$]  & 67.4 & 73.7 & $+6.3$ [$+1.5$, $+11.1$] \\
            & 90--121\,s   & 357 & 74.8 & 58.3 & 71.7 & $+13.4$ [$+9.0$, $+17.9$] & 58.3 & 72.0 & $+13.7$ [$+9.2$, $+18.2$] \\
            & Long$-$short &     &      &      &      & $+8.4$ [$+2.5$, $+14.4$]  &      &      & $+9.2$ [$+2.9$, $+15.4$] \\
\midrule
ThinkStream & $<60$\,s     & 219 & 71.2 & 64.8 & 62.1 & $-2.7$ [$-8.7$, $+3.2$]   & 63.0 & 61.2 & $-1.8$ [$-8.2$, $+4.6$] \\
            & 60--90\,s    & 270 & 73.7 & 60.7 & 58.1 & $-2.6$ [$-8.5$, $+3.0$]   & 61.9 & 61.9 & $0.0$ [$-4.1$, $+4.1$] \\
            & 90--121\,s   & 357 & 67.2 & 54.3 & 55.5 & $+1.1$ [$-3.9$, $+6.4$]   & 51.8 & 57.4 & $+5.6$ [$+2.0$, $+9.2$] \\
            & Long$-$short &     &      &      &      & $+3.9$ [$-4.0$, $+11.8$]  &      &      & $+7.4$ [$+0.2$, $+14.8$] \\
\bottomrule
\end{tabular}
\end{table*}

\paragraph{Recall settings.}
On the 846 Video-MME questions, AURA scores 74.9\%, 74.8\%, 74.2\%, and 72.8\% for recent-window sizes $k=5$, 10, 15, and 20.
Three sampling seeds at $k=10$ give 74.8\%, 75.7\%, and 74.3\%.
All tested settings exceed recent-only memory by 7.0--9.8 points.

\paragraph{Per-task recovery.}
MOSS-VL-Realtime is the one child for which adding the gate and PRS raises harmful VSB ASR.
Under native streaming its ASR is already far below its parent's (28.2\% vs.\ 83.1\%), and PRS moves its hidden state toward that parent: PRS alone raises its ASR from 29.7\% under $S_{\mathrm{fix}}$ to 36.2\% (Table~\ref{tab:supp-gate-selectivity}).
Table~\ref{tab:moss-full-intervention} gives its absolute scores and gate activation rates.
With the complete plugin, its harmful VSB ASR is 35.1\%, and its MCV implicit ASR falls from 68.3\% to 52.9\%, below the parent's 60.1\%.
The per-child plots report the corresponding results for every child.

\begin{table}[!htbp]
\centering
\caption{MOSS-VL-Realtime intervention results after excluding the plugin's development items. Gate is the percentage of requests triggering refusal. TimeBlind uses instance accuracy over 577 four-question instances, and fairness uses the official aggregate score.}
\label{tab:moss-full-intervention}
\footnotesize
\setlength{\tabcolsep}{3pt}
\begin{tabular}{@{}lrrrrrr@{}}
\toprule
Metric (\%) & Rows & $P$ & $S$ & $S_{\mathrm{fix}}$ & $S_{\mathrm{plug}}$ & Gate \\
\midrule
Video-MME acc. & 846 & 78.5 & 73.3 & 74.0 & 74.7 & 0.0 \\
TOMATO acc. & 1,405 & 34.8 & 29.3 & 29.5 & 30.7 & 0.0 \\
TimeBlind $I_{\mathrm{Acc}}$ & 2,308 & 12.1 & 6.9 & 7.3 & 5.2 & 0.4 \\
\midrule
VSB harmful ASR & 1,066 & 83.1 & 28.2 & 29.7 & 35.1 & 14.2 \\
VSB benign ASR & 1,066 & 76.4 & 77.9 & 78.8 & 78.5 & 3.0 \\
MCV explicit ASR & 1,428 & 86.5 & 55.3 & 56.0 & 41.5 & 40.3 \\
MCV implicit ASR & 1,428 & 60.1 & 68.3 & 68.3 & 52.9 & 6.0 \\
\midrule
VMDT fairness & 1,893 & 80.8 & 80.3 & 80.6 & 80.3 & 0.1 \\
VMDT privacy & 567 & 85.0 & 88.0 & 87.8 & 86.8 & 0.0 \\
\bottomrule
\end{tabular}
\end{table}

\subsection{Response Selectivity}
\label{supp:recovery-checks}

Stream++ refuses some harmless requests, but no child answers significantly fewer harmless requests than under native streaming.
We evaluate benign answering on 142 harmless clips: 46 held-out clips from Video-MME short videos, none of which was used to set the gate threshold, and 96 clips from Video-MME medium videos. Each clip has a how-to request and a description request.
A request counts as answered only when the reply provides the requested content, so silence, off-task replies, and refusals all leave it unanswered.
Relative to native streaming, Stream++ raises the share of answered requests by 54.9 points for MMDuet ($[+49.6,+59.9]$), 24.3 for AURA ($[+19.7,+28.9]$), 12.7 for ThinkStream ($[+6.3,+19.4]$), and 9.9 for MMDuet2 ($[+6.3,+13.7]$), with 95\% bootstrap intervals over clips.
The changes for LiveStar ($-2.8$, $[-6.3,+0.4]$) and MOSS-VL-Realtime ($-1.1$, $[-3.5,+1.1]$) include zero.

\begin{table*}[t]
\centering
\caption{Gate selectivity and answering of harmless requests (\%). ASR is measured on 1,066 harmful VSB requests, and refusal and answered rates on 142 harmless how-to requests. $S_{\mathrm{fix}}$+PRS is the plugin without the gate. Random refuses requests at random to match the gate's added how-to refusal. Selectivity is Random minus $S_{\mathrm{plug}}$ in points, with bootstrap 95\% intervals.}
\label{tab:supp-gate-selectivity}
\small
\setlength{\tabcolsep}{4pt}
\begin{tabular}{@{}lrrrrrrrlrrr@{}}
\toprule
 & \multicolumn{5}{c}{Harmful VSB ASR} & \multicolumn{2}{c}{How-to refusal} & & \multicolumn{3}{c}{How-to answered} \\
\cmidrule(lr){2-6}\cmidrule(lr){7-8}\cmidrule(l){10-12}
Child & $P$ & $S$ & $S_{\mathrm{fix}}$+PRS & $S_{\mathrm{plug}}$ & Random & $S_{\mathrm{fix}}$+PRS & $S_{\mathrm{plug}}$ & Selectivity [95\% CI] & $P$ & $S$ & $S_{\mathrm{plug}}$ \\
\midrule
MMDuet           & 71.3 &  6.9 & 56.8 & 42.6 & 56.8 &  0.0 &  0.0 & $+14.3$ [$+12.3$, $+16.6$] & 81.7 &  0.7 & 26.8 \\
ThinkStream      & 56.8 & 58.1 & 60.7 & 43.9 & 59.8 &  0.0 &  1.4 & $+15.9$ [$+13.4$, $+18.5$] & 68.3 & 12.7 & 21.1 \\
MMDuet2          & 56.8 & 31.1 & 73.5 & 34.1 & 63.2 & 24.6 & 35.2 & $+29.1$ [$+23.6$, $+34.6$] & 68.3 & 18.3 & 38.0 \\
AURA             & 17.7 & 59.3 & 28.9 & 13.4 & 28.9 &  0.0 &  0.0 & $+15.5$ [$+13.4$, $+17.7$] & 86.6 & 46.5 & 88.0 \\
LiveStar         & 35.7 & 51.1 & 54.1 & 18.5 & 49.5 &  9.9 & 17.6 & $+31.0$ [$+27.3$, $+34.6$] & 57.0 & 23.2 & 17.6 \\
MOSS-VL-Realtime & 83.1 & 28.2 & 36.2 & 35.1 & 35.4 &  0.0 &  2.1 & $+0.4$ [$-0.8$, $+1.4$]    & 24.6 & 83.1 & 81.7 \\
\bottomrule
\end{tabular}
\end{table*}

\paragraph{Selectivity against random refusal.}
The gate removes far more harmful assistance than random refusal at the same cost on harmless requests.
Starting from the plugin without the gate, a policy that refuses each request at random with probability $q$, set to match the gate's added refusal on the 142 how-to requests, lowers harmful VSB ASR from $A_0$ to $A_0(1-q)$.
For five children, the gate's ASR on the 1,066 harmful requests is 14.3--31.0 points below this random baseline, with bootstrap 95\% intervals above zero (Table~\ref{tab:supp-gate-selectivity}).
MOSS-VL-Realtime's difference is $+0.4$ because its gate fires on only 14.2\% of harmful requests.
The gate score separates harmful VSB requests from how-to requests with AUROC 0.83--0.96 and from description requests with AUROC 0.92--1.00.
ThinkStream's gate score does not separate harmful requests from VMDT fairness items (AUROC 0.48).

\paragraph{Safety and answering relative to native streaming.}
For AURA and ThinkStream, Stream++ improves both safety and answering relative to native streaming.
AURA's harmful VSB ASR falls from 59.3\% to 13.4\% while its answered how-to requests rise from 46.5\% to 88.0\%, close to its parent's 17.7\% and 86.6\%.
ThinkStream's ASR falls from 58.1\% to 43.9\%, and its answered how-to and description requests rise from 12.7\% to 21.1\% and from 40.8\% to 57.7\%.
LiveStar's ASR falls from 51.1\% to 18.5\%, while its answered how-to requests change from 23.2\% to 17.6\% ($-5.6$, $[-12.7,+0.7]$).
With Stream++, MMDuet2 answers 38.0\% of the 142 how-to requests; natively it answers 18.3\%.
ASR rises for MMDuet, from 6.9\% to 42.6\%, because its native replies are mostly silent; its answered how-to requests rise from 0.7\% to 26.8\%. It also rises for MOSS-VL-Realtime, from 28.2\% to 35.1\%, because its gate rarely fires.
On Video-MME, TOMATO, and TimeBlind, every child answers 99.4\%--100\% of questions with Stream++.

\paragraph{Response forms under Stream++.}
Stream++ lowers ASR by refusing harmful requests explicitly rather than leaving them unaddressed.
We label its replies to the 1,066 harmful VSB requests with the response categories of Fig.~\ref{fig:composition} and compare them with native streaming on the same requests (Table~\ref{tab:supp-plugin-forms}).
For five children, the share of unaddressed requests falls by 31.2--68.9 points and refusal or safe deflection rises by 26.9--67.1 points. Silence disappears for every child, and AURA's deferrals fall from 59.2\% to 0.4\%.
For ThinkStream, AURA, and LiveStar, the three children whose ASR falls, the 95\% bootstrap intervals over videos exclude zero for both changes.
The gate's fixed refusal accounts for 66.7\%--97.2\% of their refusals and safe deflections under Stream++.
Because Respond requires an answer, harmful assistance also rises over native streaming by 4.3--11.9 points for five children, and changes by $-1.3$ points for LiveStar ($[-3.1,+0.4]$).
It stays below the parent's level for five children.

\begin{table*}[t]
\centering
\caption{Response forms on the 1,066 harmful VSB requests (\%). Replies are labelled with the response categories of Fig.~\ref{fig:composition}. Unaddressed requests include silence, deferral, off-task narration, and fragments. Gate is the share of $S_{\mathrm{plug}}$'s refusals and safe deflections that are the gate's fixed refusal.}
\label{tab:supp-plugin-forms}
\small
\setlength{\tabcolsep}{4pt}
\begin{tabular}{@{}lrrrrrrrrrrrrr@{}}
\toprule
 & \multicolumn{3}{c}{Unaddressed} & \multicolumn{3}{c}{Refusal or deflection} & \multicolumn{3}{c}{Harmful assistance} & \multicolumn{3}{c}{ASR} & \\
\cmidrule(lr){2-4}\cmidrule(lr){5-7}\cmidrule(lr){8-10}\cmidrule(lr){11-13}
Child & $P$ & $S$ & $S_{\mathrm{plug}}$ & $P$ & $S$ & $S_{\mathrm{plug}}$ & $P$ & $S$ & $S_{\mathrm{plug}}$ & $P$ & $S$ & $S_{\mathrm{plug}}$ & Gate \\
\midrule
MMDuet           & 52.0 & 99.1 & 67.8 & 17.3 &  0.0 & 26.9 & 30.8 &  0.9 &  5.3 & 71.3 &  6.9 & 42.6 & 100.0 \\
ThinkStream      &  4.4 & 88.7 & 54.6 & 42.9 &  2.6 & 31.9 & 52.7 &  8.6 & 13.5 & 56.8 & 58.1 & 43.9 &  90.9 \\
MMDuet2          &  4.4 & 86.4 & 17.4 & 42.9 &  0.4 & 63.4 & 52.7 & 13.2 & 19.1 & 56.8 & 31.1 & 34.1 &  84.9 \\
AURA             &  0.6 & 64.3 &  1.4 & 77.9 & 31.0 & 81.9 & 21.6 &  4.8 & 16.7 & 17.7 & 59.3 & 13.4 &  66.7 \\
LiveStar         & 12.3 & 91.3 & 25.5 & 61.6 &  0.8 & 67.8 & 26.1 &  8.0 &  6.7 & 35.7 & 51.1 & 18.5 &  97.2 \\
MOSS-VL-Realtime & 87.8 & 11.4 & 11.0 &  1.1 & 60.9 & 55.5 & 11.1 & 27.7 & 33.5 & 83.1 & 28.2 & 35.1 &  25.5 \\
\bottomrule
\end{tabular}
\end{table*}

Gate activation also depends on retained evidence.
For ThinkStream, native-budget Recall yields a 19.6\% gate activation rate on VMDT fairness, compared with 7.7\% under full history.
On its answered requests, the fairness difference from its parent is $+0.7$ points ($[-2.7,+3.4]$).
These measurements report the trade-off between suppressing harmful assistance and answering legitimate requests under the final plugin.

\subsection{Medium and Long Videos}
\label{supp:meidum-long-video}

Recall changes the memory only for AURA and ThinkStream, the two children that discard earlier frames, so we evaluate these two on longer videos.
AURA is evaluated on Video-MME's full medium and long splits, each with 900 questions on 300 videos.
ThinkStream is evaluated on the full medium split and on the 179 long videos of at most 2,560 seconds, with 537 questions.
For each child, we evaluate the offline parent $P$, the child offline $C$, native streaming $S$, Recall at the native budget under $S_{\mathrm{fix}}$, and the complete plugin $S_{\mathrm{plug}}$.
The memory budgets, Recall settings, PRS, and gate thresholds are those of the short-video evaluation.

Three settings differ from the short-video protocol.
First, the offline sampler caps its input at 768 frames, which no short video reaches. This is the default limit of the official Qwen sampler, and at this limit the 8B models already need more memory than one 48\,GB GPU has.
Second, ThinkStream's static cache is enlarged to 163,840 tokens for the retained thoughts.
Third, ThinkStream's position index approaches Qwen2.5-VL's limit of 128,000 on long videos: on a 56-minute video, the native stream ended before the question near position 120,000, with or without retained thoughts. On long videos, we therefore stream without the thoughts generated before the question and keep the videos of at most 2,560 seconds.
Table~\ref{tab:supp-medium-long} lists the results next to those on the 846 short-video questions of the Stream++ evaluation.

\begin{table}[!htbp]
\centering
\caption{Video-MME results of AURA and ThinkStream by video length. (a)~Accuracy (\%). Short is the 846 Video-MME questions of the Stream++ evaluation. $C$ is the child offline. Recall is $S_{\mathrm{fix}}$ with Recall at the native budget. $\Delta$ is $S_{\mathrm{plug}}-S$ in points, with paired 95\% intervals over videos. ThinkStream's long row covers the 179 long videos of at most 2,560 seconds. (b)~Adaptation ($C-P$), execution ($S-C$), and streaming ($S-P$) taxes on medium and long videos, in points with paired 95\% bootstrap intervals over videos. Grey marks a difference that is not significant after Holm correction over these four evaluations.}
\label{tab:supp-medium-long}
\small
\setlength{\tabcolsep}{4pt}
\resizebox{\columnwidth}{!}{%
\begin{tabular}{@{}llrrrrrrl@{}}
\toprule
\multicolumn{9}{@{}l}{\textit{(a) Accuracy}} \\
Child & Videos & $n$ & $P$ & $C$ & $S$ & Recall & $S_{\mathrm{plug}}$ & $\Delta$ [95\% CI] \\
\midrule
AURA & Short & 846 & 80.6 & 78.4 & 66.0 & 74.8 & 75.8 & $+9.8$ [$+7.1$, $+12.5$] \\
& Medium & 900 & 72.4 & 69.0 & 52.9 & 60.9 & 59.9 & $+7.0$ [$+3.9$, $+10.2$] \\
& Long & 900 & 63.9 & 58.8 & 46.8 & 51.7 & 51.3 & $+4.6$ [$+1.7$, $+7.4$] \\
\midrule
ThinkStream & Short & 846 & 72.8 & 70.3 & 59.1 & 59.8 & 58.0 & $-1.1$ [$-4.4$, $+2.1$] \\
& Medium & 900 & 58.1 & 56.6 & 38.2 & 41.2 & 41.0 & $+2.8$ [$-0.3$, $+5.9$] \\
& Long & 537 & 49.2 & 43.9 & 24.4 & 24.8 & 26.4 & $+2.0$ [$-0.7$, $+5.0$] \\
\bottomrule
\end{tabular}}

\vspace{6pt}
\resizebox{\columnwidth}{!}{%
\begin{tabular}{@{}llccc@{}}
\toprule
\multicolumn{5}{@{}l}{\textit{(b) Taxes}} \\
Child & Videos & $C-P$ & $S-C$ & $S-P$ \\
\midrule
AURA & Medium & $-3.4$ [$-6.2$, $-0.8$] & $-16.1$ [$-19.8$, $-12.4$] & $-19.6$ [$-23.2$, $-15.9$] \\
 & Long & $-5.1$ [$-7.7$, $-2.6$] & $-12.0$ [$-15.2$, $-8.8$] & $-17.1$ [$-20.7$, $-13.7$] \\
\midrule
ThinkStream & Medium & \textcolor[HTML]{8A939B}{$-1.6$ [$-4.0$, $+1.0$]} & $-18.3$ [$-21.9$, $-14.9$] & $-19.9$ [$-23.7$, $-16.1$] \\
 & Long & $-5.2$ [$-8.4$, $-2.0$] & $-19.6$ [$-24.8$, $-14.2$] & $-24.8$ [$-29.6$, $-19.7$] \\
\bottomrule
\end{tabular}}
\end{table}

On longer videos, Recall continues to help AURA, and its gains for ThinkStream are smaller.
For AURA, Recall adds 8.0 points over native streaming on medium videos ($[+4.9,+11.1]$) and 4.9 points on long videos ($[+2.0,+7.8]$), compared with a rise from 66.0\% to 74.8\% on short videos, and Stream++ gains 7.0 and 4.6 points.
With Stream++, AURA remains 12.6 points below its parent on medium videos ($[-15.6,-9.6]$) and on long videos ($[-15.9,-9.2]$), against 4.8 points on short videos ($[-7.8,-2.0]$).
For ThinkStream, the native accuracy is 38.2\% on medium videos and 24.4\% on long videos. Recall adds 3.0 points on medium videos ($[+0.3,+5.7]$) and 0.4 points on long videos ($[-1.9,+2.6]$), and Stream++ gains 2.8 and 2.0 points.

Most of the streaming tax on longer videos is execution tax (Table~\ref{tab:supp-medium-long}b).
On medium and long videos, the execution tax is 16.1 and 12.0 points for AURA and 18.3 and 19.6 points for ThinkStream, between 70\% and 92\% of the streaming tax. All four are significant after Holm correction over these four evaluations ($p<0.001$).
The adaptation tax is 3.4 and 5.1 points for AURA and 1.6 and 5.2 points for ThinkStream, and it is significant except for ThinkStream on medium videos ($p=0.27$).

\subsection{Native Streaming Performance}
\label{supp:streaming-benchmarks}

We evaluate OVO-Bench~\cite{niu2025ovobench} and the real-time questions of StreamingBench~\cite{lin2024streamingbench}, where questions arrive during the video.
The evaluation uses clips lasting at most 120 seconds up to the question: 1,576 OVO-Bench rows and 700 StreamingBench rows.
MOSS-VL-Realtime has 1,546 matched OVO-Bench rows with valid execution traces.
We report mean question accuracy, with unanswered questions scored as incorrect.
The gate uses the same calibrated thresholds as the offline-task evaluation.
Intervals use paired bootstrapping over source videos.

Stream++ preserves the children's streaming ability (Table~\ref{tab:supp-streaming-benchmarks}).
On both benchmarks, no child's accuracy falls significantly.
In addition, LiveStar improves by 10.9 points on OVO-Bench and 7.1 on StreamingBench.
Respond raises MMDuet's scores by 42.1 and 72.7 points by eliciting answers.

\end{document}